\documentclass[twoside,11pt]{article}

\usepackage{jmlr2e}

\usepackage{graphicx}
\usepackage{amssymb}
\usepackage{amsmath}
\usepackage{amsbsy}
\usepackage{latexsym}
\usepackage{subfigure}
\usepackage{url}
\usepackage{multirow}

\newcommand{\QED}{\rule{7pt}{7pt}}

\newcommand{\numparams}{n}
\newcommand{\J}{J}

\newcommand{\argmin}{\mathop{\mathrm{argmin\,}}}

\newcommand{\unorm}[1]{\|#1\|}
\newcommand{\unorms}[1]{\unorm{#1}^2}

\newcommand{\iid}{\stackrel{\mathrm{i.i.d.}}{\sim}}

\newcommand{\mathbbE}{\mathbb{E}}

\newcommand{\mathbbR}{\mathbb{R}}
\newcommand{\mathbbV}{\mathbb{V}}

\newcommand{\boldone}{{\boldsymbol{1}}}

\newcommand{\boldH}{{\boldsymbol{H}}}
\newcommand{\boldI}{{\boldsymbol{I}}}

\newcommand{\boldU}{{\boldsymbol{U}}}

\newcommand{\boldh}{{\boldsymbol{h}}}

\newcommand{\boldx}{{\boldsymbol{x}}}

\newcommand{\boldbeta}{{\boldsymbol{\beta}}}

\newcommand{\boldtheta}{{\boldsymbol{\theta}}}

\newcommand{\calF}{{\mathcal{F}}}
\newcommand{\calG}{{\mathcal{G}}}
\newcommand{\calH}{{\mathcal{H}}}

\newcommand{\calO}{{\mathcal{O}}}

\newcommand{\calX}{{\mathcal{X}}}

\newcommand{\thetah}{{\widehat{\theta}}}

\newcommand{\boldthetah}{{\widehat{\boldtheta}}}

\newcommand{\inputdim}{d}

\newcommand{\ratiosymbol}{r}
\newcommand{\ratio}{\ratiosymbol}
\newcommand{\ratioh}{\widehat{\ratiosymbol}}
\newcommand{\ratiomodel}{g}

\newcommand{\boldxnu}{\boldx}
\newcommand{\boldxde}{\boldx'}

\newcommand{\pnu}{p}
\newcommand{\pde}{p'}
\newcommand{\Pnu}{P}
\newcommand{\Pde}{P'}
\newcommand{\pmix}{q_\alpha}

\newcommand{\nnu}{\nsample}
\newcommand{\nde}{\nsample'}
\newcommand{\nsample}{n}

\newcommand{\nparam}{b}

\newcommand{\xde}{x^{\mathrm{q}}}

\newcommand{\xnu}{x^{\mathrm{p}}}

\newcommand{\calXde}{\calX'}

\newcommand{\calXnu}{\calX}

\newcommand{\calXdet}{\widetilde{\calX}'}
\newcommand{\calXnut}{\widetilde{\calX}}

\newcommand{\boldHh}{{\widehat{\boldH}}}
\newcommand{\boldhh}{{\widehat{\boldh}}}

\newcommand{\Hh}{{\widehat{H}}}
\newcommand{\hh}{{\widehat{h}}}

\newcommand{\calDX}{{\mathcal{D}_{\mathrm{X}}}}

\newcommand{\EE}{{\mathrm{E}}}

\newcommand{\hw}{\widehat{g}}
\newcommand{\tw}{{g^*}}

\newcommand{\np}{n}
\newcommand{\nq}{n'}
\newcommand{\boldxq}{\boldx'}
\newcommand{\boldxp}{\boldx}

\newcommand{\QQ}{P'}
\newcommand{\PP}{P}
\newcommand{\Qn}{P'_{\nq}}
\newcommand{\Pn}{P_{\np}}
\newcommand{\pp}{p}
\newcommand{\qq}{p'}

\newcommand{\dd}{\mathrm{d}}

 \def\Rbb{\mathbb{R}}

\def\pnu{p}
\def\pde{p'}
\def\xnu{x}
\def\xde{x'}
\def\mnu{n}
\def\mde{n'}
\def\Gnu{G}
\def\Gde{G'}

\def\Vbb{\mathbb{V}} 
\def\Enu{{\mathbb{E}_{\pnu(\boldx)}}}   
\def\Ede{{\mathbb{E}_{\pde(\boldx)}}}  
\def\Vnu{{\mathbb{V}_{\pnu(\boldx)}}}   
\def\Vde{{\mathbb{V}_{\pde(\boldx)}}}  
\def\relPE{\mathrm{PE}_{\alpha}}  
\def\hatPEest{\widehat{\mathrm{PE}}_{\alpha}}     
\def\tildePEest{\widetilde{\mathrm{PE}}_{\alpha}} 

\def\relratioModel{\mathcal{G}}   
\def\ratio{r}                     
\def\relratio{{\ratio}_{\alpha}}  

\def\x{\mbox{\boldmath $x$}}
\def\param{\mbox{\boldmath $\theta$}}
\def\Param{\Theta}

\jmlrheading{1}{2000}{1-48}{4/00}{10/00}{Makoto Yamada, Taiji Suzuki, Takafumi Kanamori, Hirotaka Hachiya, and Masashi Sugiyama}

\ShortHeadings{Relative Density-Ratio Estimation}{Yamada, Suzuki, Kanamori, Hachiya, and Sugiyama}
\firstpageno{1}

\begin{document}

\title{Relative Density-Ratio Estimation\\
for Robust Distribution Comparison
}

\author{\name Makoto Yamada \email yamada@sg.cs.titech.ac.jp\\
\addr Tokyo Institute of Technology\\
2-12-1 O-okayama, Meguro-ku, Tokyo 152-8552, Japan.\\
\name Taiji Suzuki \email s-taiji@stat.t.u-tokyo.ac.jp\\
\addr The University of Tokyo\\
7-3-1 Hongo, Bunkyo-ku, Tokyo 113-8656, Japan.\\
\name Takafumi Kanamori \email kanamori@is.nagoya-u.ac.jp\\
\addr Nagoya University\\
Furocho, Chikusaku, Nagoya 464-8603, Japan.\\
\name Hirotaka Hachiya \email hachiya@sg.cs.titech.ac.jp\\
\addr Tokyo Institute of Technology\\
2-12-1 O-okayama, Meguro-ku, Tokyo 152-8552, Japan.\\
\name Masashi Sugiyama \email sugi@cs.titech.ac.jp\\
\addr Tokyo Institute of Technology\\
2-12-1 O-okayama, Meguro-ku, Tokyo 152-8552, Japan.
}
\editor{???}

\maketitle

\begin{abstract}
Divergence estimators based on direct approximation of density-ratios
without going through separate approximation of numerator and denominator densities
have been successfully applied to machine learning tasks
that involve distribution comparison
such as outlier detection, transfer learning, and two-sample homogeneity test.
However, since density-ratio functions often possess high fluctuation,
divergence estimation is still a challenging task in practice.
In this paper, we propose to use \emph{relative divergences}
for distribution comparison,
which involves approximation of \emph{relative density-ratios}.
Since relative density-ratios are always smoother than corresponding ordinary density-ratios,
our proposed method is favorable in terms of the non-parametric convergence speed.
Furthermore, we show that the proposed divergence estimator has asymptotic variance
\emph{independent} of the model complexity under a parametric setup,
implying that the proposed estimator hardly overfits
even with complex models.
Through experiments, we demonstrate the usefulness of the proposed approach.
\end{abstract}


\section{Introduction}\label{sec:introduction}
Comparing probability distributions is a fundamental task in statistical data processing.
It can be used for,
e.g., \emph{outlier detection} \citep{AISTATS:Smola+etal:2009,KAIS:Hido+etal:2011},
\emph{two-sample homogeneity test} \citep{NIPS2006_583,NN:Sugiyama+etal:2011b},
and \emph{transfer learning} \citep{JSPI:Shimodaira:2000,JMLR:Sugiyama+etal:2007}.

A standard approach to comparing probability densities $\pnu(\boldx)$ and $\pde(\boldx)$
would be to estimate a divergence from $\pnu(\boldx)$ to $\pde(\boldx)$,
such as the \emph{Kullback-Leibler (KL) divergence} \citep{Annals-Math-Stat:Kullback+Leibler:1951}:
\begin{align*}
  \mathrm{KL}[\pnu(\boldx),\pde(\boldx)]
  &:=\int\log\left(\frac{\pnu(\boldx)}{\pde(\boldx)}\right)\pnu(\boldx)\mathrm{d}\boldx.
\end{align*}
A naive way to estimate the KL divergence is to separately approximate
the densities $\pnu(\boldx)$ and $\pde(\boldx)$ from data
and plug the estimated densities in the above definition.
However, since density estimation is known to be a hard task \citep{book:Vapnik:1998},
this approach does not work well unless a good parametric model is available.
Recently, a divergence estimation approach
which directly approximates the \emph{density ratio},
\begin{align*}
  \ratio(\boldx):=\frac{\pnu(\boldx)}{\pde(\boldx)},
\end{align*}
without going through separate approximation of densities
$\pnu(\boldx)$ and $\pde(\boldx)$
has been proposed \citep{AISM:Sugiyama+etal:2008,IEEE-IT:Nguyen+etal:2010}.
Such density-ratio approximation methods
were proved to achieve the optimal non-parametric convergence rate
in the mini-max sense.

However, the KL divergence estimation via density-ratio approximation is
computationally rather expensive due to the non-linearity introduced
by the `log' term.
To cope with this problem,
another divergence called the \emph{Pearson (PE) divergence} 
\citep{PhMag:Pearson:1900} is useful.
The PE divergence from $\pnu(\boldx)$ to $\pde(\boldx)$ is defined as
\begin{align*}
  \mathrm{PE}[\pnu(\boldx),\pde(\boldx)]
  &:=\frac{1}{2}\int\left(\frac{\pnu(\boldx)}{\pde(\boldx)}-1\right)^2\pde(\boldx)\mathrm{d}\boldx.
\end{align*}
The PE divergence is a squared-loss variant of the KL divergence,
and they both belong to the class of the \emph{Ali-Silvey-Csisz\'ar divergences}
\citep[which is also known as the \emph{$f$-divergences},
see][]{JRSS-B:Ali+Silvey:1966,SSM-Hungary:Csiszar:1967}.
Thus, the PE and KL divergences share similar properties,
e.g., they are non-negative and vanish if and only if $\pnu(\boldx)=\pde(\boldx)$.

Similarly to the KL divergence estimation,
the PE divergence can also be accurately estimated based on density-ratio approximation
\citep{JMLR:Kanamori+etal:2009}:
the density-ratio approximator called \emph{unconstrained least-squares importance fitting} (uLSIF)
gives the PE divergence estimator \emph{analytically},
which can be computed just by solving a system of linear equations.
The practical usefulness of the uLSIF-based PE divergence estimator
was demonstrated in various applications
such as 
outlier detection \citep{KAIS:Hido+etal:2011},
two-sample homogeneity test \citep{NN:Sugiyama+etal:2011b},
and
dimensionality reduction \citep{AISTATS:Suzuki+Sugiyama:2010}.

In this paper, we first establish the non-parametric convergence rate 
of the uLSIF-based PE divergence estimator, 
which elucidates its superior theoretical properties.
However, it also reveals that its convergence rate is actually governed
by the `sup'-norm of the true density-ratio function: $\max_\boldx \ratio(\boldx)$.
This implies that, in the region where the denominator density $\pde(\boldx)$ takes small values,
the density ratio $\ratio(\boldx)=\pnu(\boldx)/\pde(\boldx)$ tends to take large values
and therefore the overall convergence speed becomes slow.
More critically, density ratios can even diverge to infinity
under a rather simple setting, e.g., 
when the ratio of two Gaussian functions is considered \citep{NIPS2010_0731}.
This makes the paradigm of divergence estimation
based on density-ratio approximation unreliable.

In order to overcome this fundamental problem, we propose an alternative approach
to distribution comparison called \emph{$\alpha$-relative divergence estimation}.
In the proposed approach, we estimate
the quantity called the \emph{$\alpha$-relative divergence},
which is the divergence from $\pnu(\boldx)$ to
the \emph{$\alpha$-mixture density} $\alpha \pnu(\boldx)+(1-\alpha)\pde(\boldx)$
for $0\le\alpha<1$.
For example, the $\alpha$-relative PE divergence is given by
\begin{align*}
  \mathrm{PE}_\alpha[\pnu(\boldx),\pde(\boldx)]
  &:=\mathrm{PE}[\pnu(\boldx),\alpha \pnu(\boldx)+(1-\alpha)\pde(\boldx)]\nonumber\\
  &\phantom{:}=\frac{1}{2}\int\left(
    \frac{\pnu(\boldx)}{\alpha \pnu(\boldx)+(1-\alpha)\pde(\boldx)}-1\right)^2
  \left(\alpha \pnu(\boldx)+(1-\alpha)\pde(\boldx)\right)\mathrm{d}\boldx.
\end{align*}
We estimate the $\alpha$-relative divergence by direct approximation of 
the \emph{$\alpha$-relative density-ratio}:
\begin{align*}
\relratio(\boldx):=\frac{\pnu(\boldx)}{\alpha \pnu(\boldx)+(1-\alpha)\pde(\boldx)}.  
\end{align*}

A notable advantage of this approach is that
the $\alpha$-relative density-ratio is
always bounded above by $1/\alpha$ when $\alpha>0$,
even when the ordinary density-ratio is unbounded.
Based on this feature, we theoretically show that
the $\alpha$-relative PE divergence estimator
based on $\alpha$-relative density-ratio approximation
is more favorable than the ordinary density-ratio approach
in terms of the non-parametric convergence speed.

We further prove that, under a correctly-specified parametric setup, 
the asymptotic variance of our $\alpha$-relative PE divergence estimator
does not depend on the model complexity.
This means that the proposed $\alpha$-relative PE divergence estimator
hardly overfits even with complex models.

Through extensive experiments on outlier detection, two-sample homogeneity test,
and transfer learning,
we demonstrate that our proposed $\alpha$-relative PE divergence estimator
compares favorably with alternative approaches.

The rest of this paper is structured as follows.
In Section~\ref{sec:formulation},
our proposed relative PE divergence estimator is described.
In Section~\ref{sec:analysis},
we provide non-parametric analysis of the convergence rate
and parametric analysis of the variance of the proposed PE divergence estimator.
In Section~\ref{sec:experiments},
we experimentally evaluate the performance of
the proposed method on various tasks.
Finally, in Section~\ref{sec:conclusion},
we conclude the paper by summarizing our contributions
and describing future prospects.

\section{Estimation of Relative Pearson Divergence
  via Least-Squares Relative Density-Ratio Approximation}
\label{sec:formulation}
In this section, we propose an estimator of the relative Pearson (PE) divergence
based on least-squares relative density-ratio approximation.

\subsection{Problem Formulation}
Suppose we are given independent and identically distributed 
(i.i.d.) samples $\{\boldxnu_i\}_{i=1}^{\nnu}$
from a $\inputdim$-dimensional distribution $\Pnu$ with density $\pnu(\boldx)$
and i.i.d.~samples $\{\boldxde_j\}_{j=1}^{\nde}$
from another $\inputdim$-dimensional distribution $\Pde$ with density $\pde(\boldx)$:
\begin{align*}
\{\boldxnu_i\}_{i = 1}^{\nnu} &\iid \Pnu,\\
\{\boldxde_j\}_{j = 1}^{\nde} &\iid \Pde.
\end{align*}
The goal of this paper is to compare the two underlying distributions
$\Pnu$ and $\Pde$ only using the two sets of samples
$\{\boldxnu_i\}_{i=1}^{\nnu}$ and $\{\boldxde_j\}_{j=1}^{\nde}$.

For $0\le\alpha<1$,
let $\pmix(\boldx)$ be the \emph{$\alpha$-mixture density} of $\pnu(\boldx)$ and $\pde(\boldx)$:
\begin{align*}
  \pmix(\boldx):=\alpha \pnu(\boldx)+(1-\alpha)\pde(\boldx).
\end{align*}
Let $\relratio(\boldx)$ be the \emph{$\alpha$-relative density-ratio}
of $\pnu(\boldx)$ and $\pde(\boldx)$:
\begin{align}
\relratio(\boldx):=\frac{\pnu(\boldx)}{\alpha \pnu(\boldx)+(1-\alpha)\pde(\boldx)}
=\frac{\pnu(\boldx)}{\pmix(\boldx)}.
\label{alpha-ratio}
\end{align}
We define \emph{the $\alpha$-relative PE divergence} 
from $\pnu(\boldx)$ to $\pde(\boldx)$ as
\begin{align}
  \mathrm{PE}_\alpha
  &:=\frac{1}{2}\mathbbE_{\pmix(\boldx)}\left[(\relratio(\boldx)-1)^2\right],
   \label{alpha-PE}
\end{align}
where $\mathbbE_{p(\boldx)}[f(\boldx)]$ denotes the expectation of $f(\boldx)$ under $p(\boldx)$:
\begin{align*}
  \mathbbE_{p(\boldx)}[f(\boldx)]=\int f(\boldx)p(\boldx)\mathrm{d}\boldx.
\end{align*}
When $\alpha=0$, $\mathrm{PE}_\alpha$ is reduced to the ordinary PE divergence.
Thus, the $\alpha$-relative PE divergence can be regarded as a
`smoothed' extension of the ordinary PE divergence.

Below, we give a method for estimating the $\alpha$-relative PE divergence
based on the approximation of the $\alpha$-relative density-ratio.

\subsection{Direct Approximation of $\alpha$-Relative Density-Ratios}
\label{sec:uLSIF}
Here, we describe a method for approximating
the $\alpha$-relative density-ratio \eqref{alpha-ratio}.

Let us model the $\alpha$-relative density-ratio $\relratio(\boldx)$
by the following kernel model:
\begin{align*}
  \ratiomodel(\boldx;\boldtheta):=\sum_{\ell=1}^{\nnu} \theta_\ell K(\boldx,\boldxnu_\ell),
\end{align*}
where $\boldtheta:=(\theta_1,\ldots,\theta_{\nnu})^\top$
are parameters to be learned from data samples,
$^\top$ denotes the transpose of a matrix or a vector,
and
$K(\boldx,\boldx')$ is a kernel basis function.
In the experiments, we use the Gaussian kernel:
\begin{align*}
  K(\boldx,\boldx')=
  \exp\left(-\frac{\unorms{\boldx-\boldx'}}{2\sigma^2}\right),
\end{align*}
where $\sigma$ ($>0$) is the kernel width.

The parameters $\boldtheta$
in the model $\ratiomodel(\boldx;\boldtheta)$ are determined so that
the following expected squared-error $\J$ is minimized:
\begin{align*}
  \J(\boldtheta)
  &:=
  \frac{1}{2}\mathbbE_{\pmix(\boldx)}
  \left[\left(\ratiomodel(\boldx;\boldtheta)-\relratio(\boldx)\right)^2\right]\\
  &\phantom{:}=
  \frac{\alpha}{2}\mathbbE_{\pnu(\boldx)}\left[\ratiomodel(\boldx;\boldtheta)^2\right]
  +\frac{(1 - \alpha)}{2}\mathbbE_{\pde(\boldx)}\left[\ratiomodel(\boldx;\boldtheta)^2\right]
  -\mathbbE_{\pnu(\boldx)}\left[\ratiomodel(\boldx;\boldtheta)\right]
  +\mathrm{Const.},
\end{align*}
where we used $\relratio(\boldx)\pmix(\boldx)=\pnu(\boldx)$ in the third term.
Approximating the expectations by empirical averages,
we obtain the following optimization problem:
\begin{align}
  \boldthetah:=\argmin_{\boldtheta\in\mathbbR^{\numparams}}
  \left[\frac{1}{2}\boldtheta^\top\boldHh\boldtheta-\boldhh^\top\boldtheta
  +\frac{\lambda}{2}\boldtheta^\top\boldtheta\right],
  \label{uLSIF-optimization-empirical}
\end{align}
where a penalty term
$\lambda\boldtheta^\top\boldtheta/2$ is included for regularization purposes,
and $\lambda$ $(\ge0)$ denotes the regularization parameter.
$\boldHh$ is the $\nnu\times\nnu$ matrix with the $(\ell,\ell')$-th element
\begin{align}
  \Hh_{\ell,\ell'}:=
  \frac{\alpha}{\nnu}\sum_{i=1}^{\nnu} K(\boldxnu_i,\boldxnu_\ell)K(\boldxnu_i,\boldxnu_{\ell'})
  + \frac{(1-\alpha)}{\nde} \sum_{j=1}^{\nde}
  K(\boldxde_j,\boldxnu_\ell)K(\boldxde_j,\boldxnu_{\ell'}).
 \label{Hh}
\end{align}
$\boldhh$ is the $\nnu$-dimensional vector with the $\ell$-th element
\begin{align*}
  \hh_{\ell}:=\frac{1}{\nnu}\sum_{i=1}^{\nnu} K(\boldxnu_i,\boldxnu_\ell).
\end{align*}
It is easy to confirm that the solution of Eq.\eqref{uLSIF-optimization-empirical}
can be \emph{analytically} obtained as
\begin{align*}
  \boldthetah=(\boldHh+\lambda\boldI_{\nnu})^{-1}\boldhh,
\end{align*}
where $\boldI_{\nnu}$ denotes the $\nnu$-dimensional identity matrix.
Finally, a density-ratio estimator is given as
\begin{align}
  \ratioh_\alpha(\boldx):=\ratiomodel(\boldx;\boldthetah)
  =\sum_{\ell=1}^{\nnu} \thetah_\ell K(\boldx,\boldxnu_\ell).
  \label{alpha-uLSIF}
\end{align}

When $\alpha=0$, the above method is reduced to 
a direct density-ratio estimator called
\emph{unconstrained least-squares importance fitting} 
\citep[uLSIF;][]{JMLR:Kanamori+etal:2009}.
Thus, the above method can be regarded as an extension of uLSIF
to the $\alpha$-relative density-ratio.
For this reason, we refer to our method as \emph{relative uLSIF} (RuLSIF).

The performance of RuLSIF depends on 
the choice of the kernel function
(the kernel width $\sigma$ in the case of the Gaussian kernel)
and the regularization parameter $\lambda$.
Model selection of RuLSIF is possible based on cross-validation
with respect to the squared-error criterion $J$,
in the same way as the original uLSIF \citep{JMLR:Kanamori+etal:2009}.

A MATLAB$^\text{\textregistered}$ implementation of 
RuLSIF is available from 
\begin{center}
(made public after acceptance)
\end{center}

\subsection{$\alpha$-Relative PE Divergence Estimation Based on RuLSIF}
Using an estimator of the $\alpha$-relative density-ratio $\relratio(\boldx)$,
we can construct estimators of the $\alpha$-relative PE divergence \eqref{alpha-PE}.
After a few lines of calculation,
we can show that the $\alpha$-relative PE divergence \eqref{alpha-PE}
is equivalently expressed as
\begin{align*}
  \mathrm{PE}_\alpha
  &\phantom{:}=
  -\frac{\alpha}{2}\mathbbE_{\pnu(\boldx)}\left[\relratio(\boldx)^2\right]
  -\frac{(1 - \alpha)}{2}\mathbbE_{\pde(\boldx)}\left[\relratio(\boldx)^2\right]
  +\mathbbE_{\pnu(\boldx)}\left[\relratio(\boldx)\right]
  -\frac{1}{2}\\
  &\phantom{:}=
  \frac{1}{2}\mathbbE_{\pnu(\boldx)}\left[\relratio(\boldx)\right]
  -\frac{1}{2}.
\end{align*}
Note that the first line can also be obtained via \emph{Legendre-Fenchel convex duality}
of the divergence functional \citep{book:Rockafellar:1970}.

Based on these expressions, we consider the following two estimators:
\begin{align}
\widehat{\mathrm{PE}}_\alpha
&:=
-\frac{\alpha}{2\nnu}\sum_{i = 1}^{\nnu} \ratioh(\boldxnu_i)^2
- \frac{(1-\alpha)}{2\nde}\sum_{j = 1}^{\nde} \ratioh(\boldxde_j)^2
+\frac{1}{\nnu}\sum_{i=1}^{\nnu}\ratioh(\boldxnu_i) 
-\frac{1}{2},
\label{eq:PE1}\\
\widetilde{\mathrm{PE}}_\alpha
&:=\frac{1}{2\nnu}\sum_{i=1}^{\nnu}\ratioh(\boldxnu_i)-\frac{1}{2}.
\label{eq:PE2}
\end{align}
We note that the $\alpha$-relative PE divergence \eqref{alpha-PE}
can have further different expressions than the above ones,
and corresponding estimators can also be constructed similarly.
However, the above two expressions will be particularly useful:
the first estimator $\widehat{\mathrm{PE}}_\alpha$
has superior theoretical properties (see Section~\ref{sec:analysis})
and the second one $\widetilde{\mathrm{PE}}_\alpha$
is simple to compute.

\subsection{Illustrative Examples}
\label{subsec:illustration}
Here, we numerically illustrate the behavior of RuLSIF
\eqref{alpha-uLSIF} using toy datasets.
Let the numerator distribution be $\Pnu = N(0,1)$,
where $N(\mu, \sigma^2)$ denotes the normal distribution with mean $\mu$ and variance $\sigma^2$.
The denominator distribution $\Pde$ is set as follows:
\begin{description}
\item[(a)] $\Pde=N(0,1)$: $\Pnu$ and $\Pde$ are the same.
\item[(b)] $\Pde=N(0,0.6)$: $\Pde$ has smaller standard deviation than $\Pnu$.
\item[(c)] $\Pde=N(0,2)$: $\Pde$ has larger standard deviation than $\Pnu$.
\item[(d)] $\Pde=N(0.5,1)$: $\Pnu$ and $\Pde$ have different means.
\item[(e)] $\Pde=0.95N(0,1)+0.05N(3,1)$: $\Pde$ contains an additional component to $\Pnu$.
\end{description}
We draw $\nnu=\nde=300$ samples from the above densities,
and compute RuLSIF for $\alpha=0$, $0.5$, and $0.95$.

\begin{figure}[p]
  \centering
  \subfigure[$\Pde=N(0,1)$: $\Pnu$ and $\Pde$ are the same.]{
    \begin{tabular}{@{}c@{}c@{}c@{}c@{}}
 \includegraphics[width=.24\textwidth]{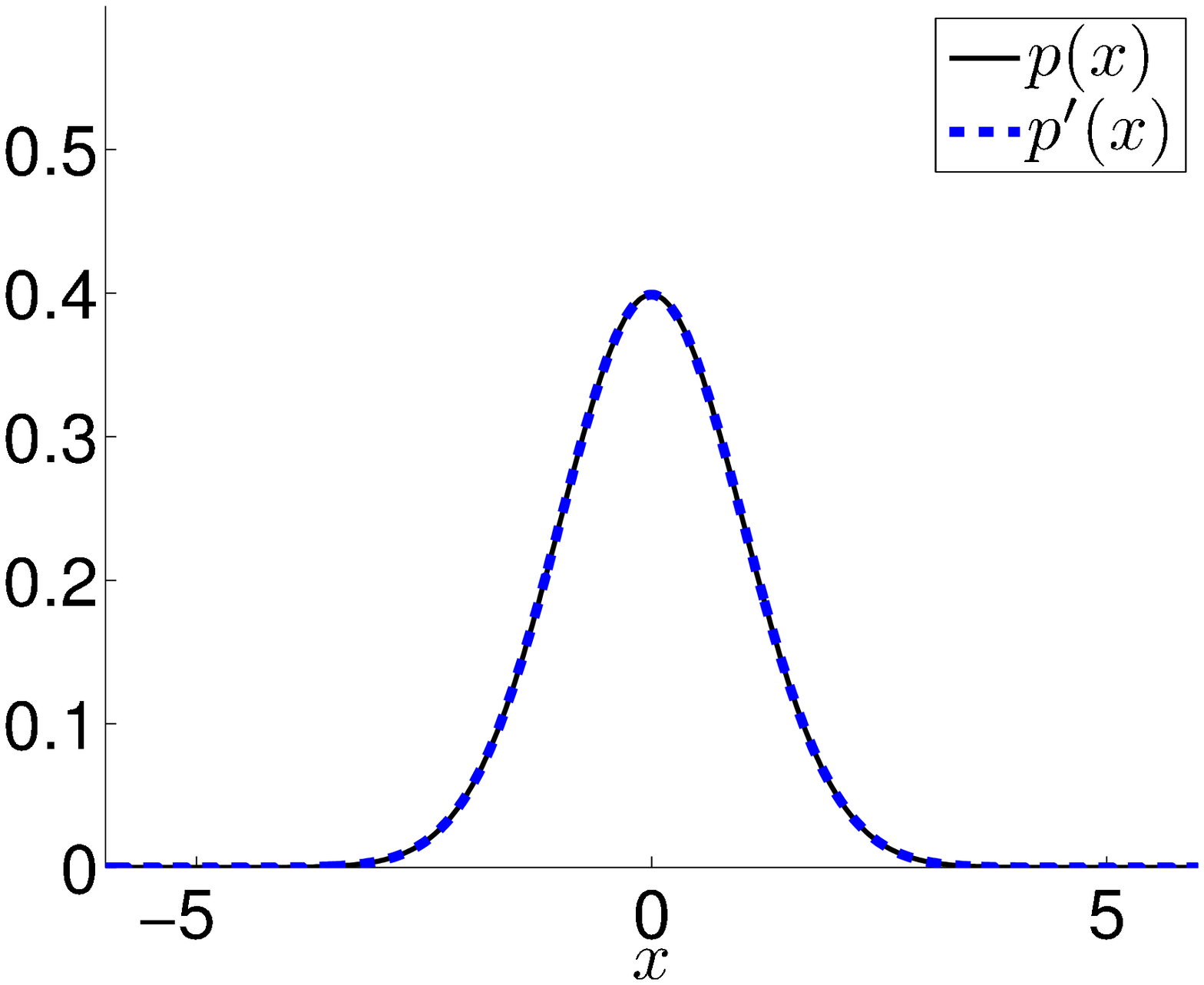}&
 \includegraphics[width=.24\textwidth]{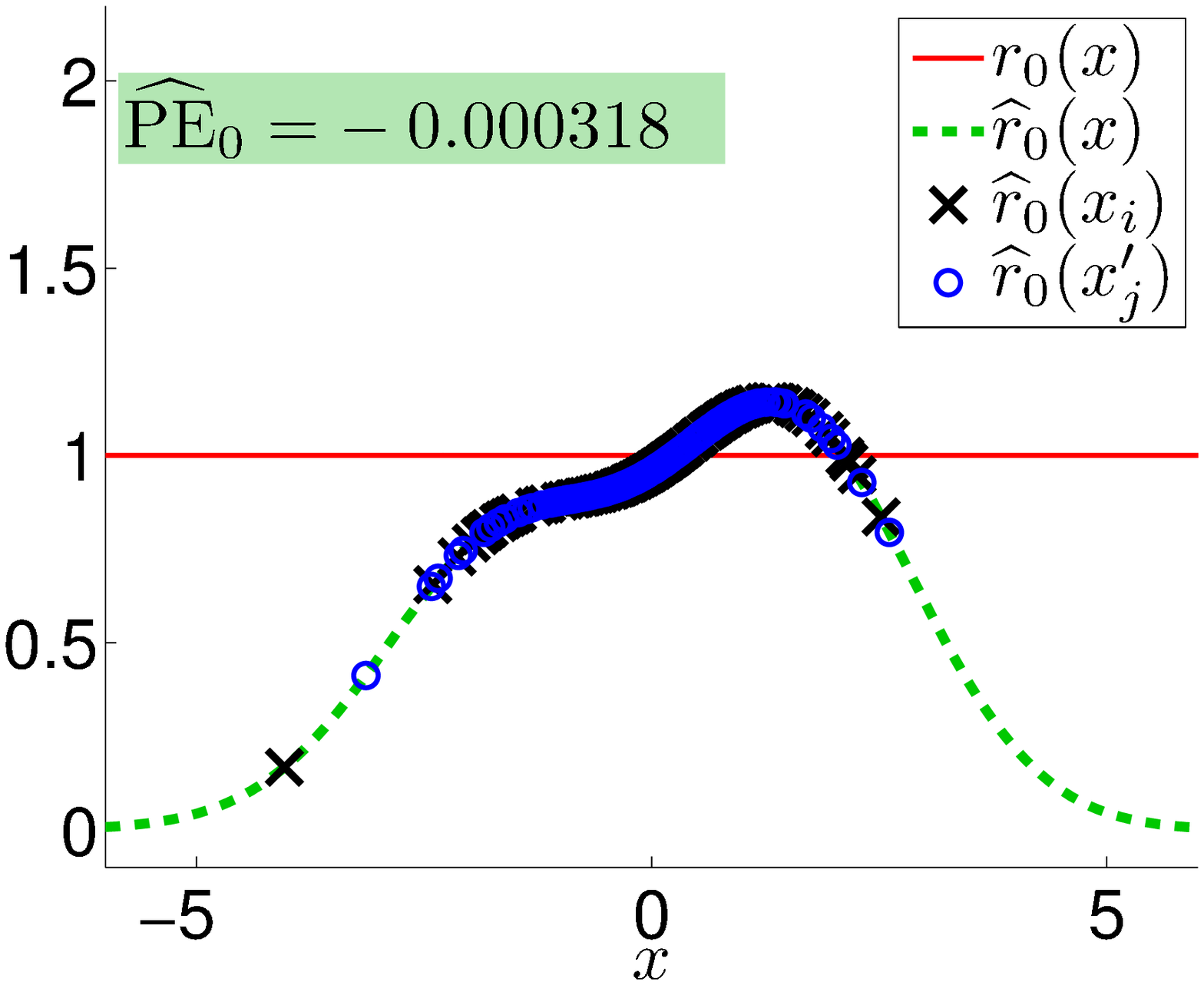}&
 \includegraphics[width=.24\textwidth]{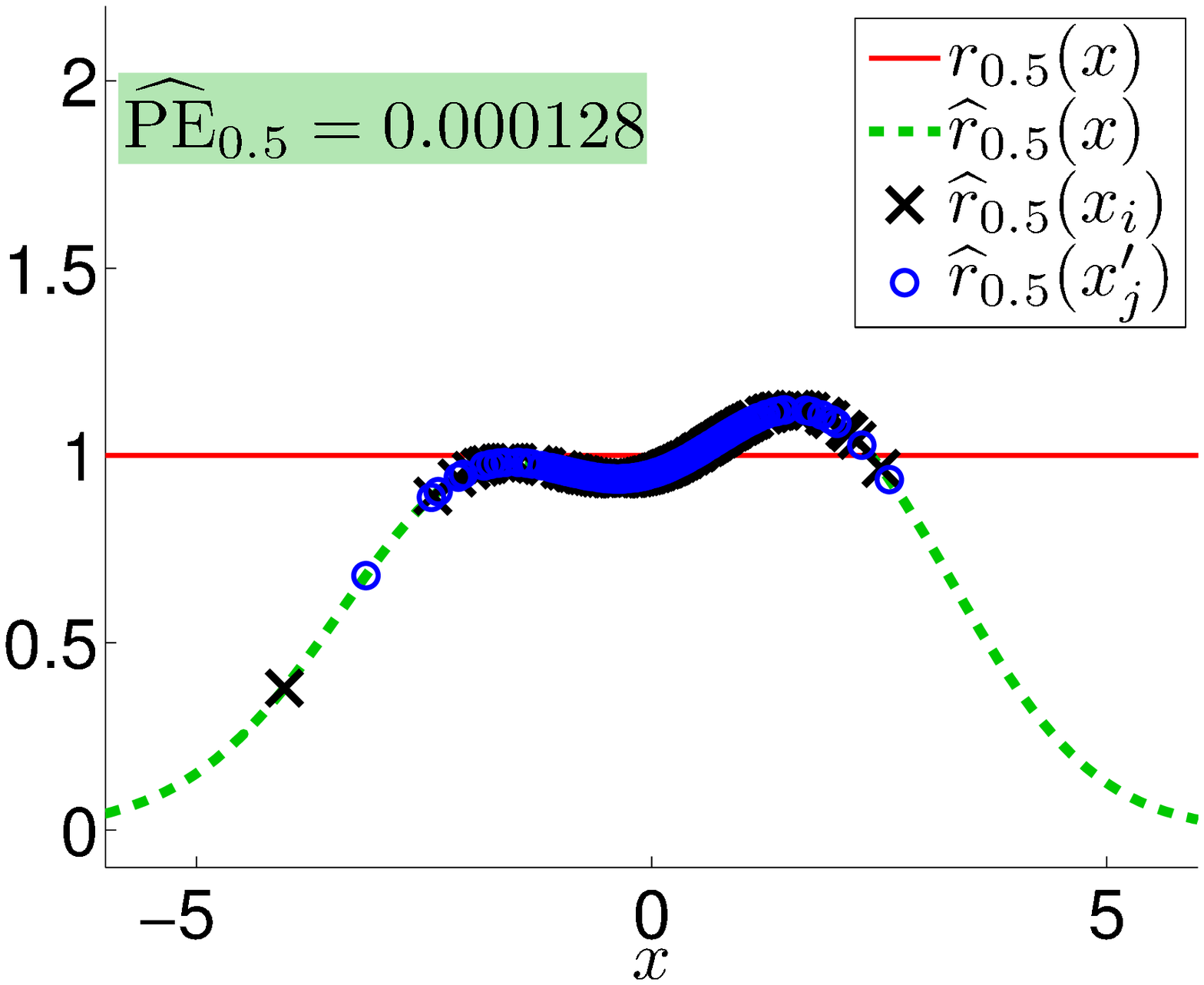}&
 \includegraphics[width=.24\textwidth]{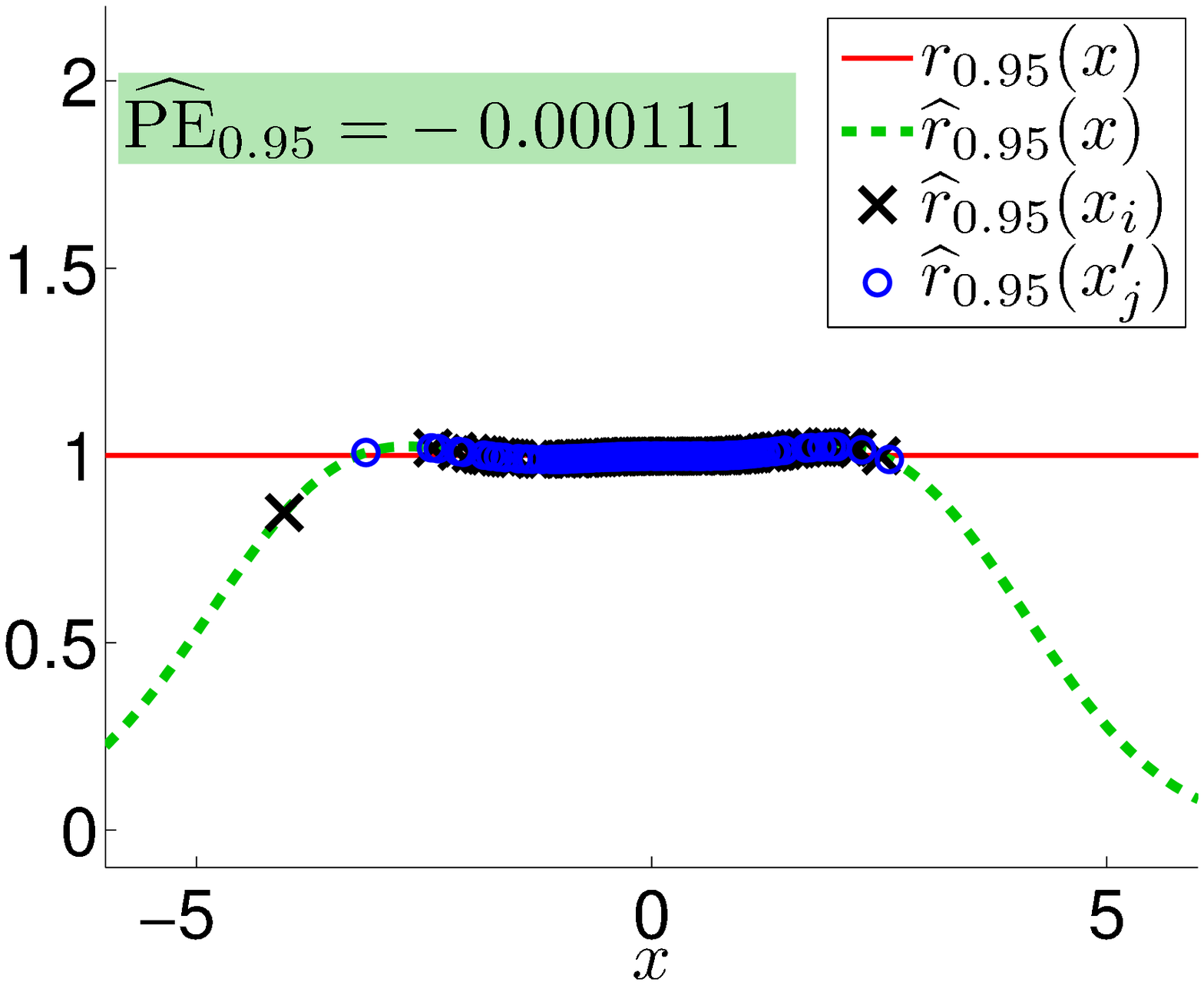}
    \end{tabular}
      }
 \subfigure[$\Pde=N(0,0.6)$: $\Pde$ has smaller standard deviation than $\Pnu$.]{
    \begin{tabular}{@{}c@{}c@{}c@{}c@{}}
 \includegraphics[width=.24\textwidth]{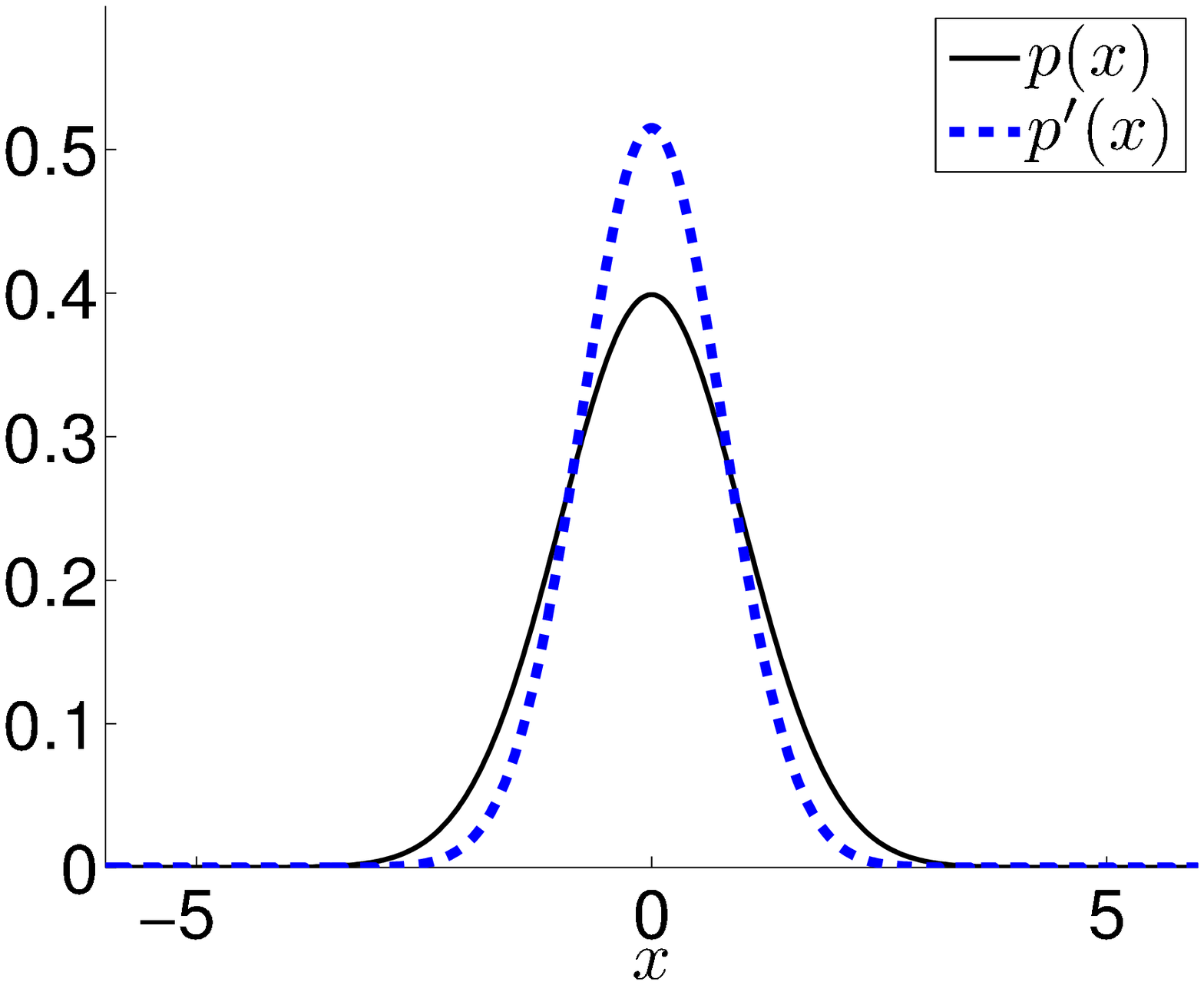}&
 \includegraphics[width=.24\textwidth]{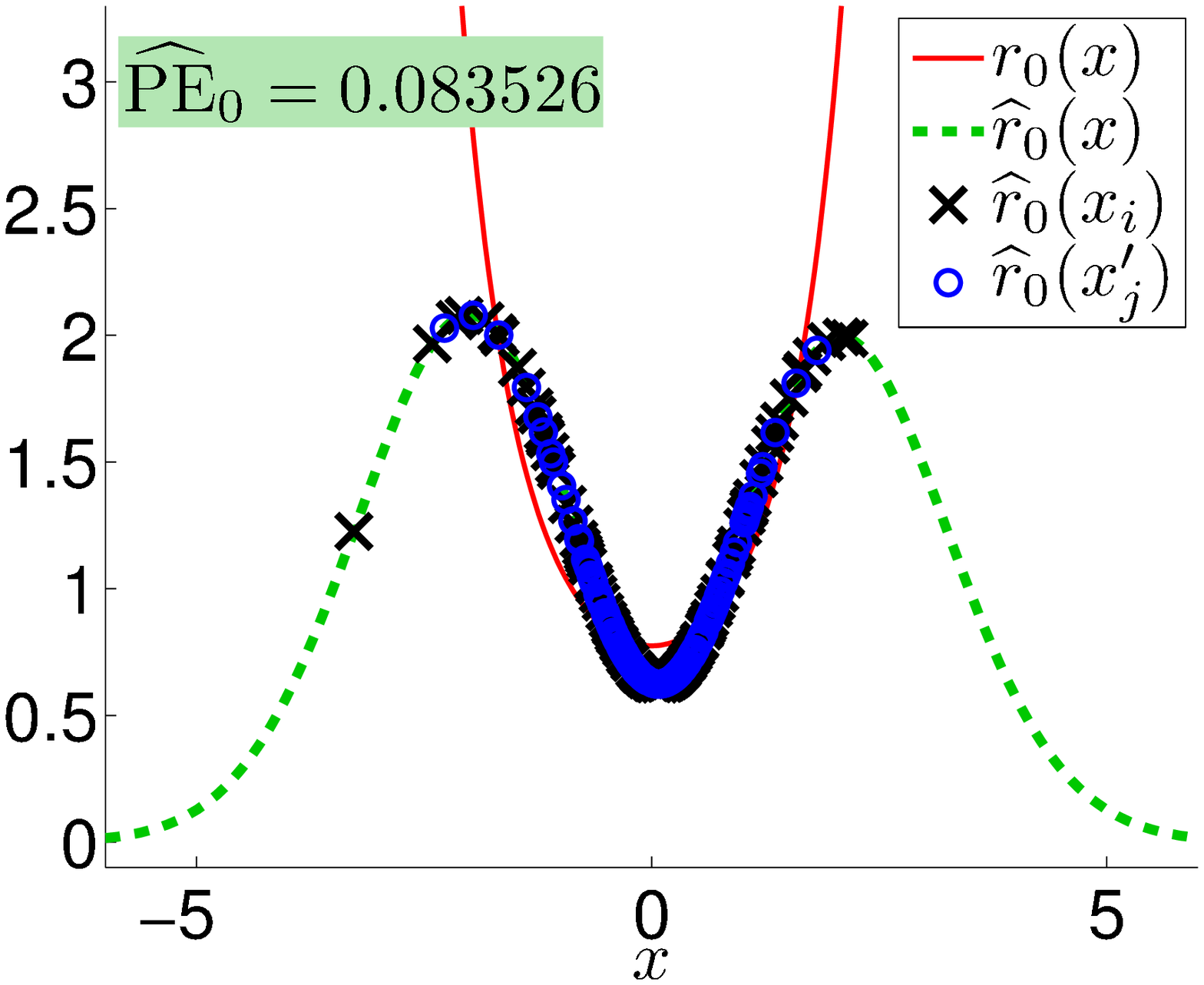}&
 \includegraphics[width=.24\textwidth]{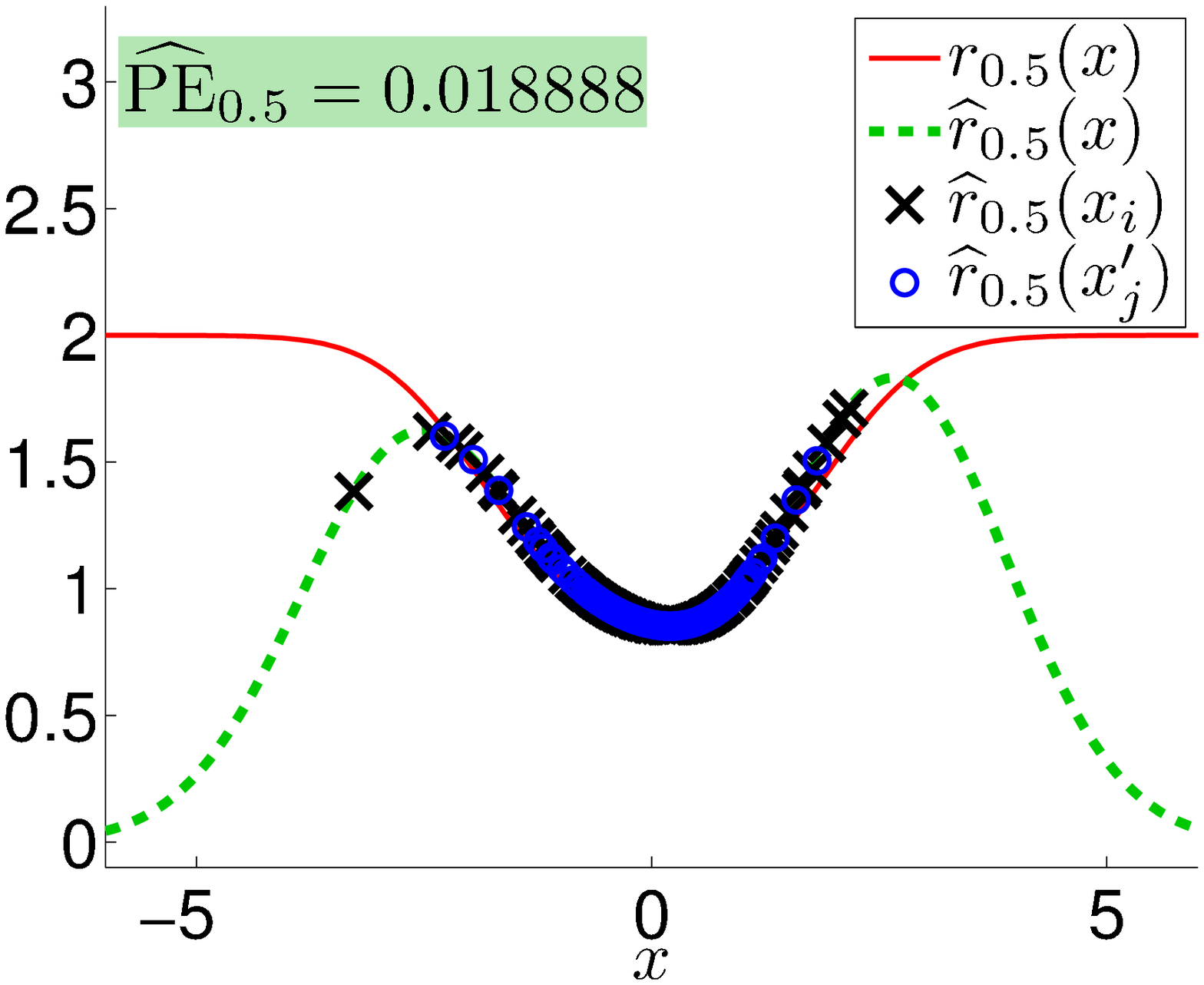}&
 \includegraphics[width=.24\textwidth]{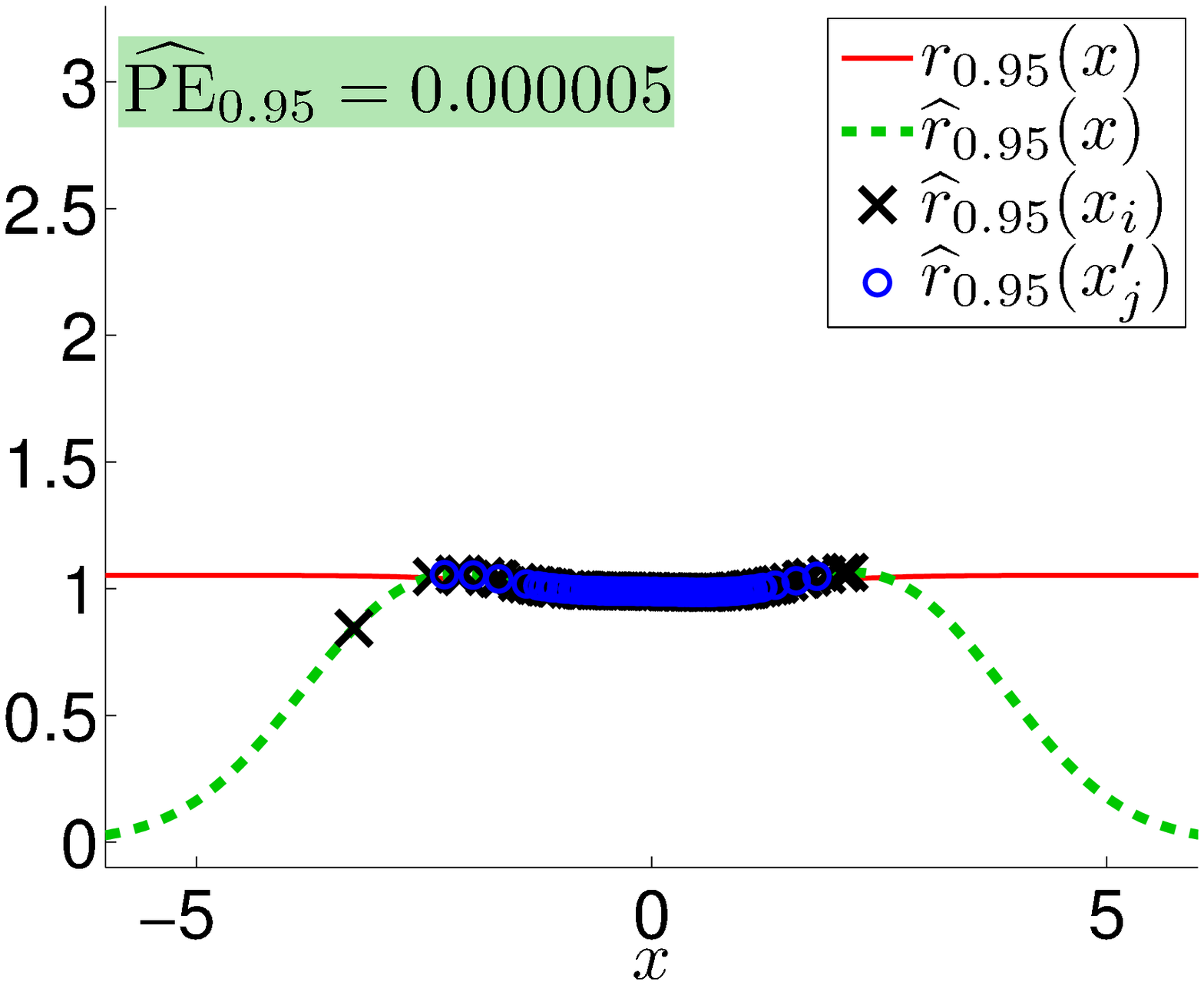}
    \end{tabular}
      }
 \subfigure[$\Pde=N(0,2)$: $\Pde$ has larger standard deviation than $\Pnu$.]{
    \begin{tabular}{@{}c@{}c@{}c@{}c@{}}
 \includegraphics[width=.24\textwidth]{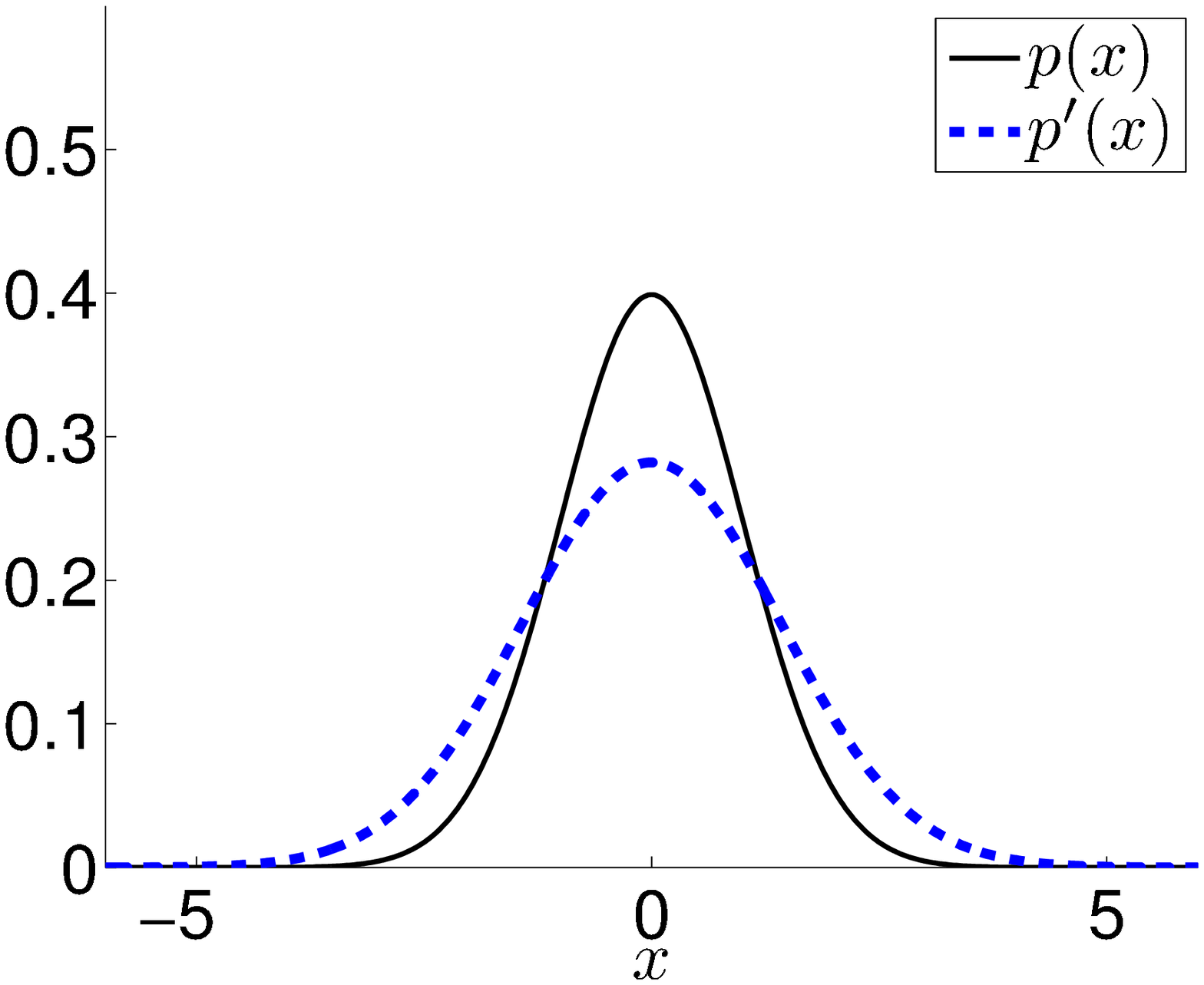}&
 \includegraphics[width=.24\textwidth]{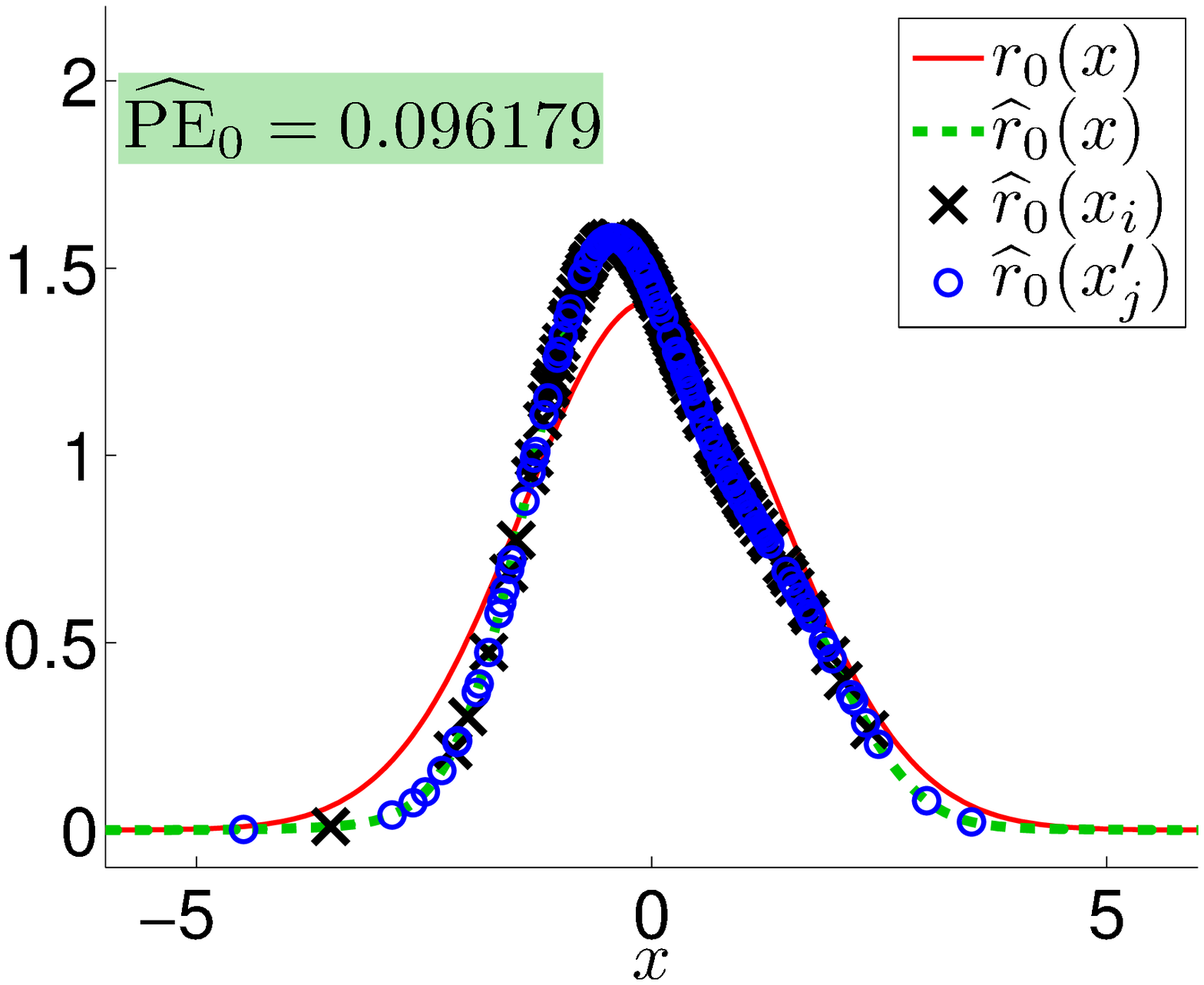}&
 \includegraphics[width=.24\textwidth]{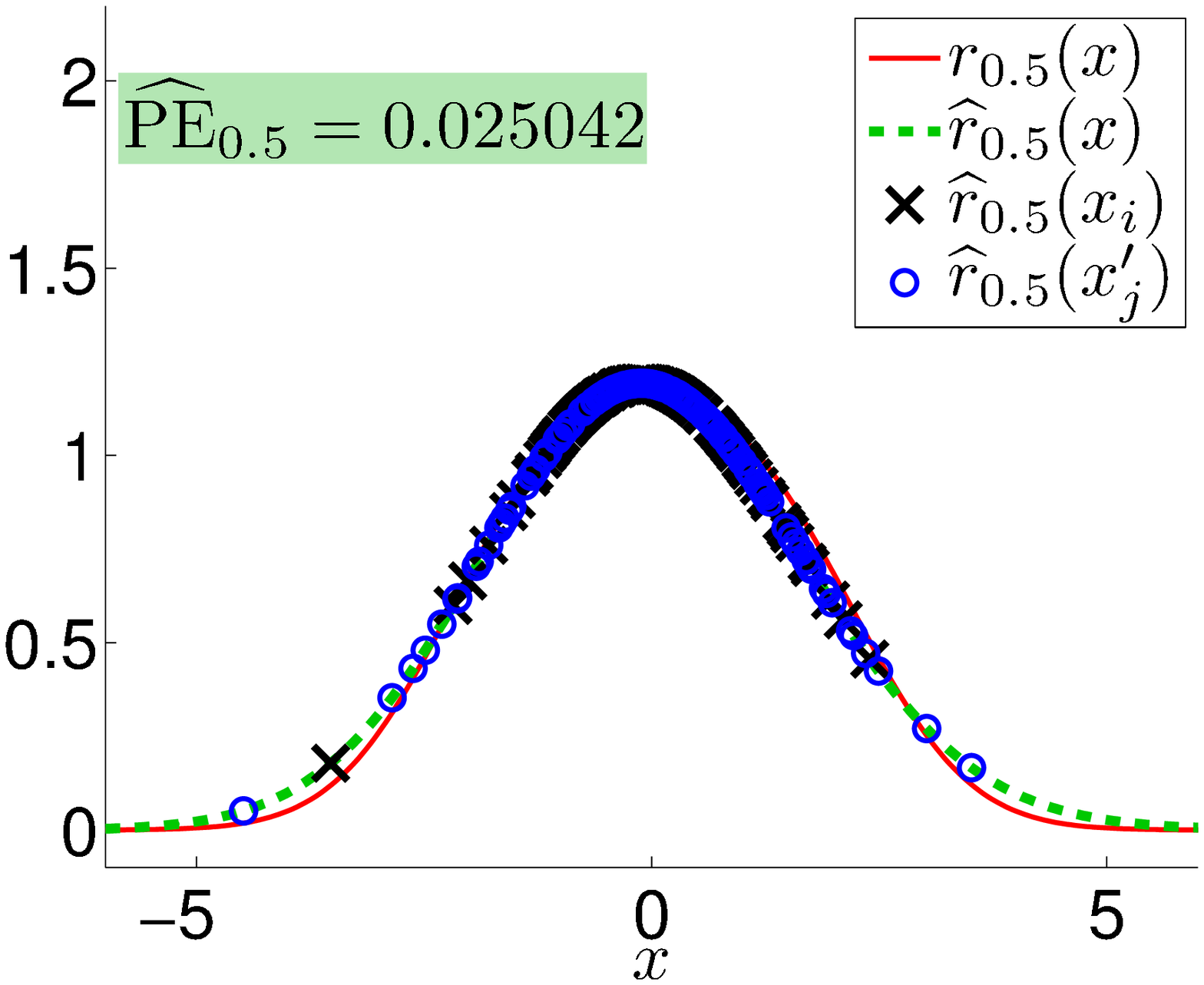}&
 \includegraphics[width=.24\textwidth]{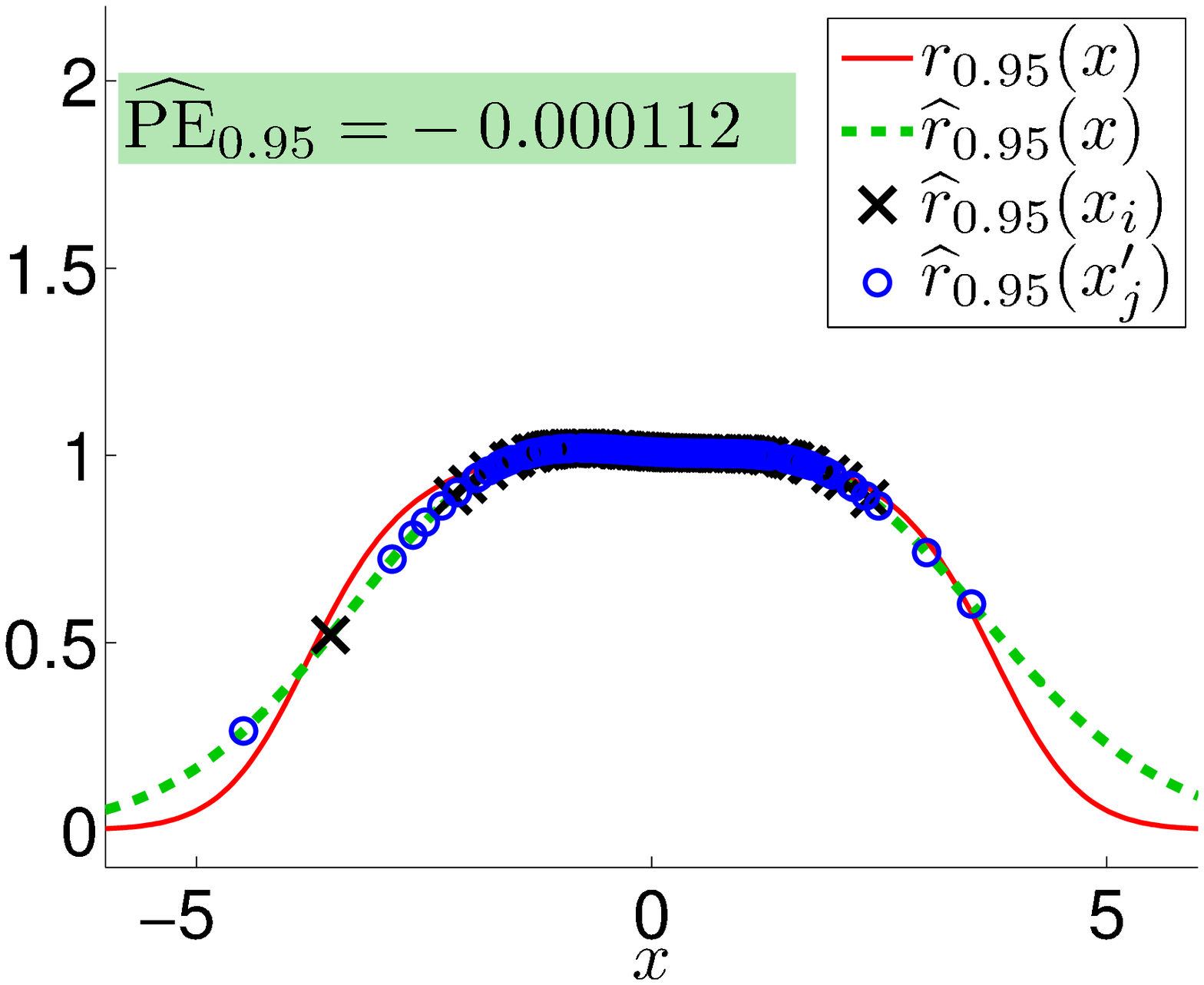}
    \end{tabular}
      }
 \subfigure[$\Pde=N(0.5,1)$: $\Pnu$ and $\Pde$ have different means.]{
    \begin{tabular}{@{}c@{}c@{}c@{}c@{}}
 \includegraphics[width=.24\textwidth]{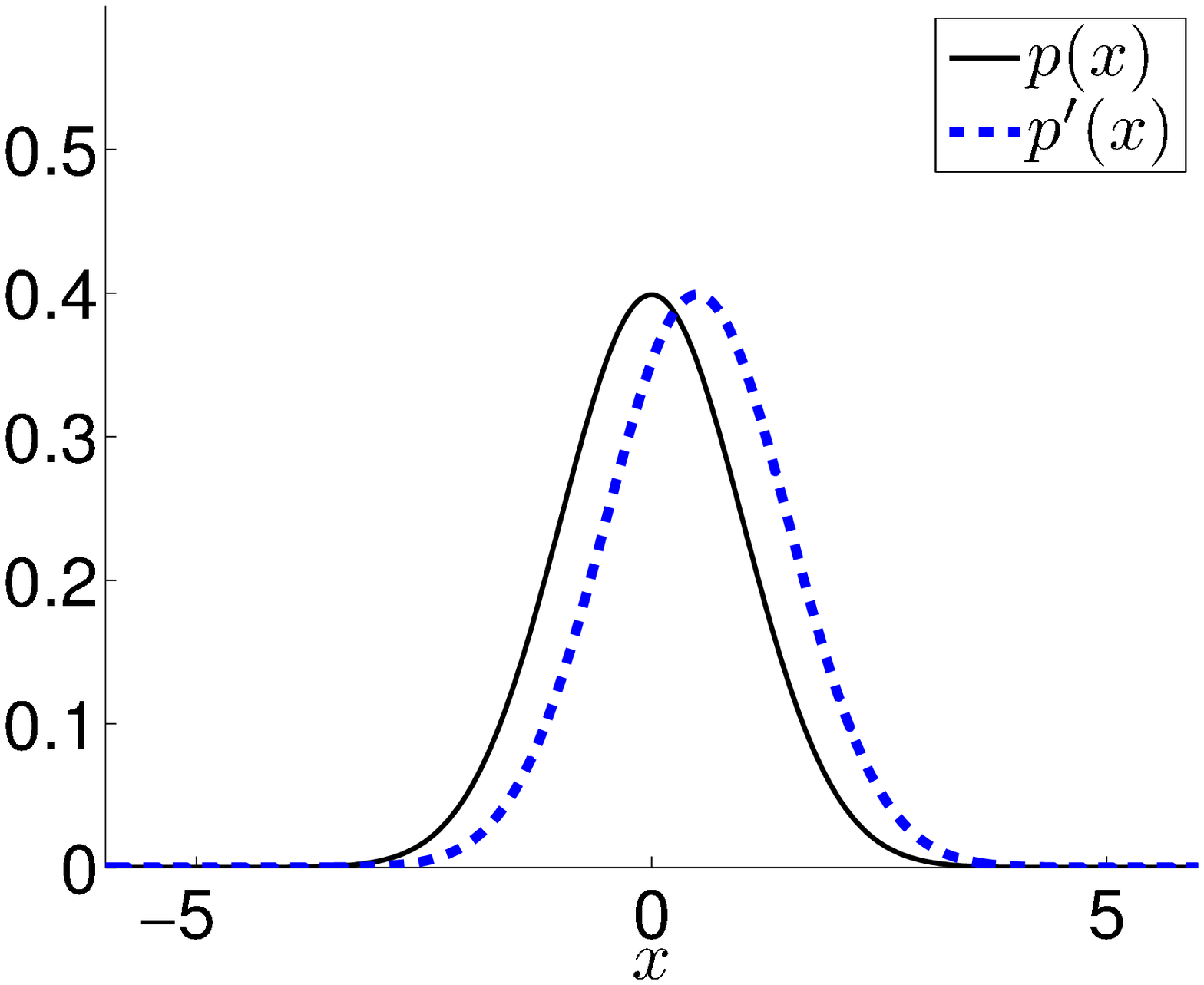}&
 \includegraphics[width=.24\textwidth]{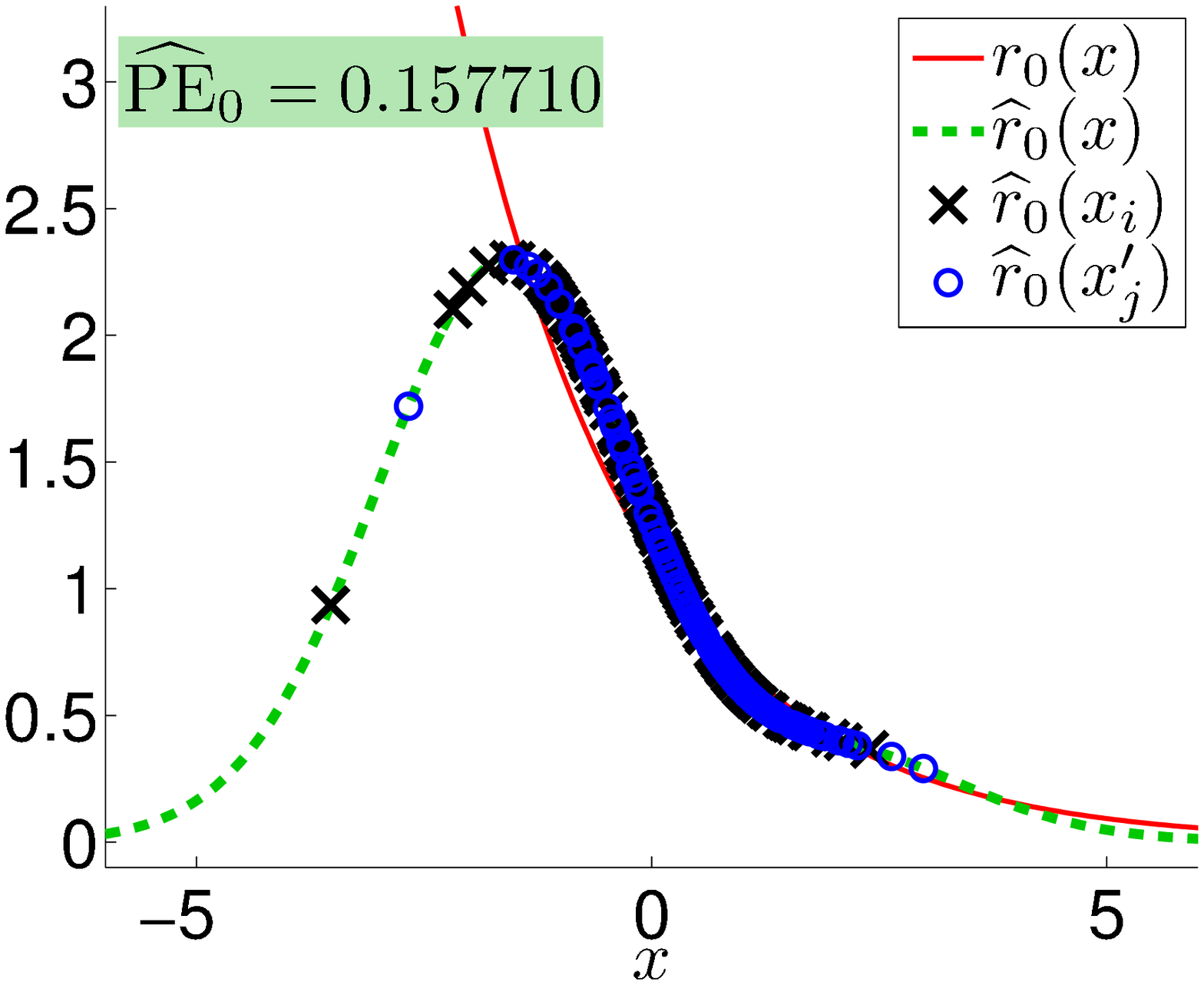}&
 \includegraphics[width=.24\textwidth]{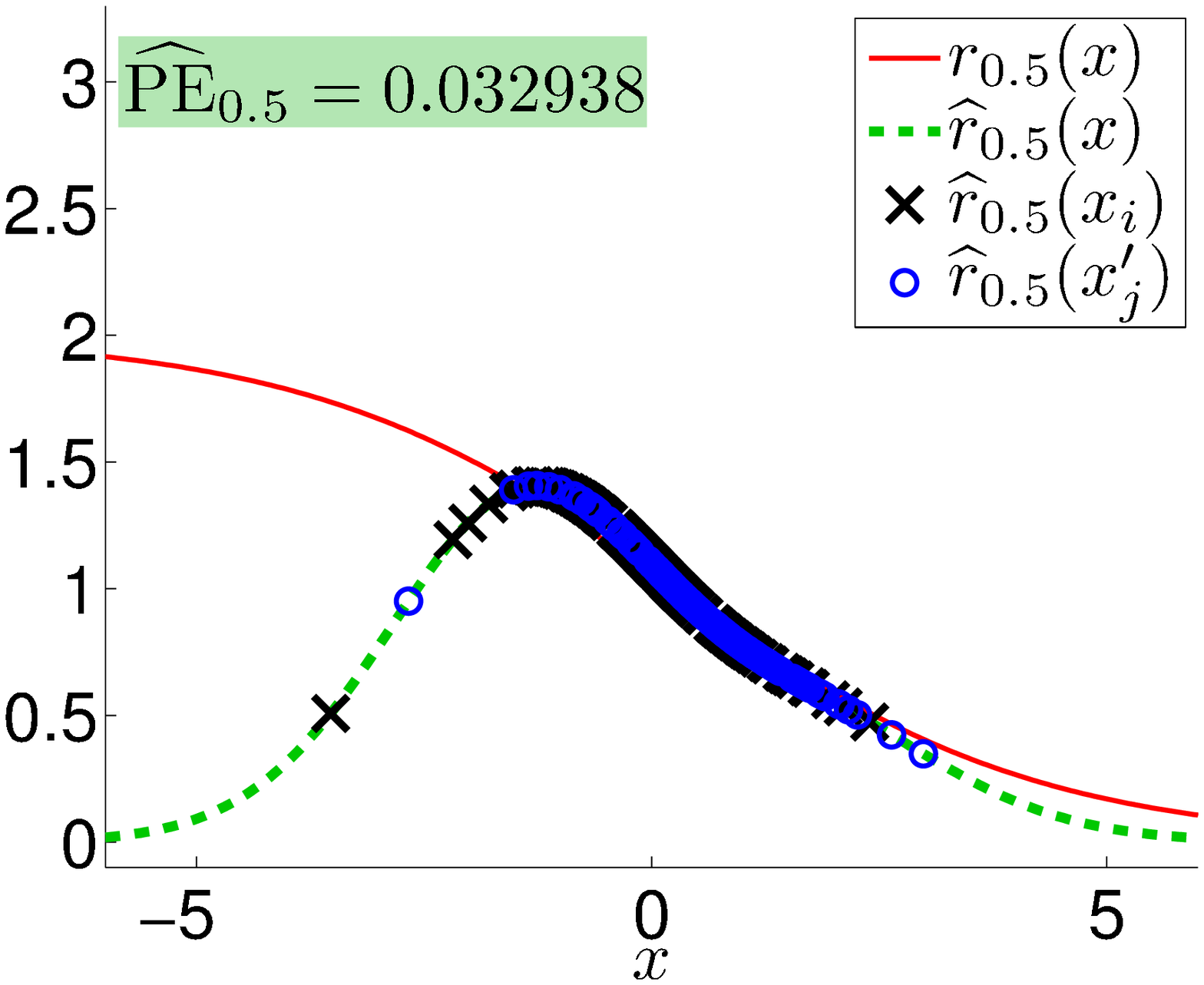}&
 \includegraphics[width=.24\textwidth]{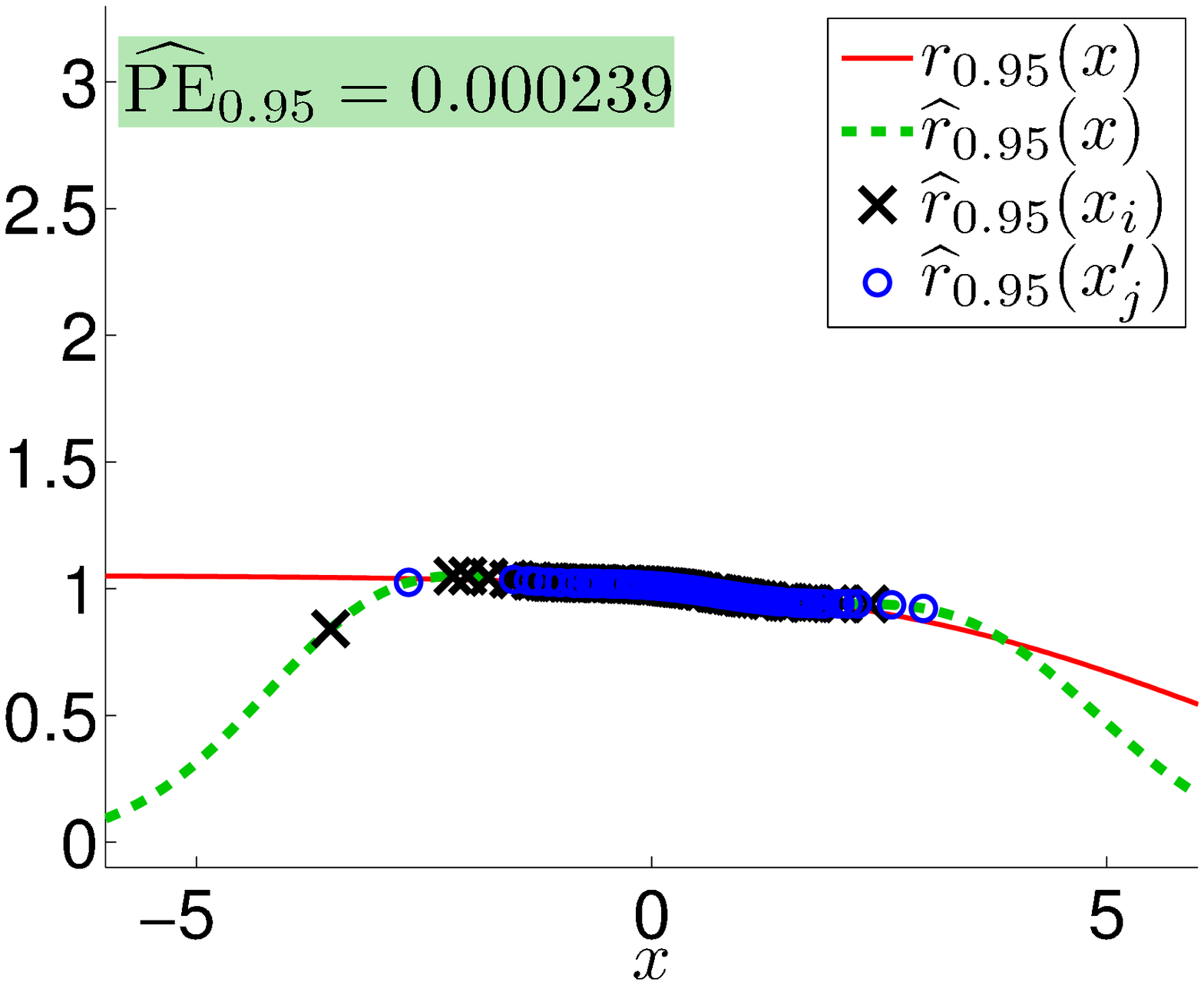}
    \end{tabular}
      }
 \subfigure[$\Pde=0.95N(0,1)+0.05N(3,1)$: $\Pde$ contains an additional component to $\Pnu$.]{
    \begin{tabular}{@{}c@{}c@{}c@{}c@{}}
  \includegraphics[width=.24\textwidth]{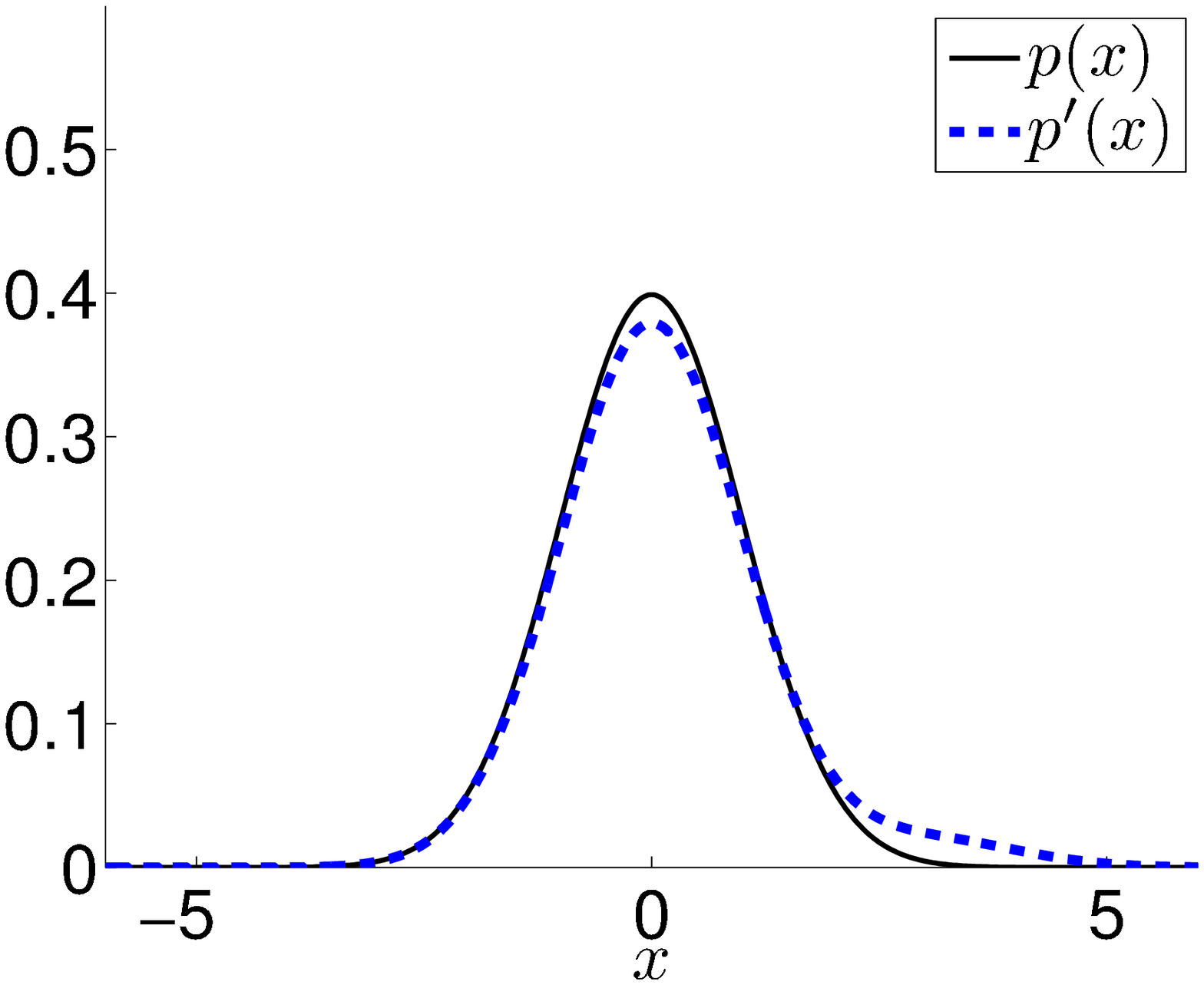}&
  \includegraphics[width=.24\textwidth]{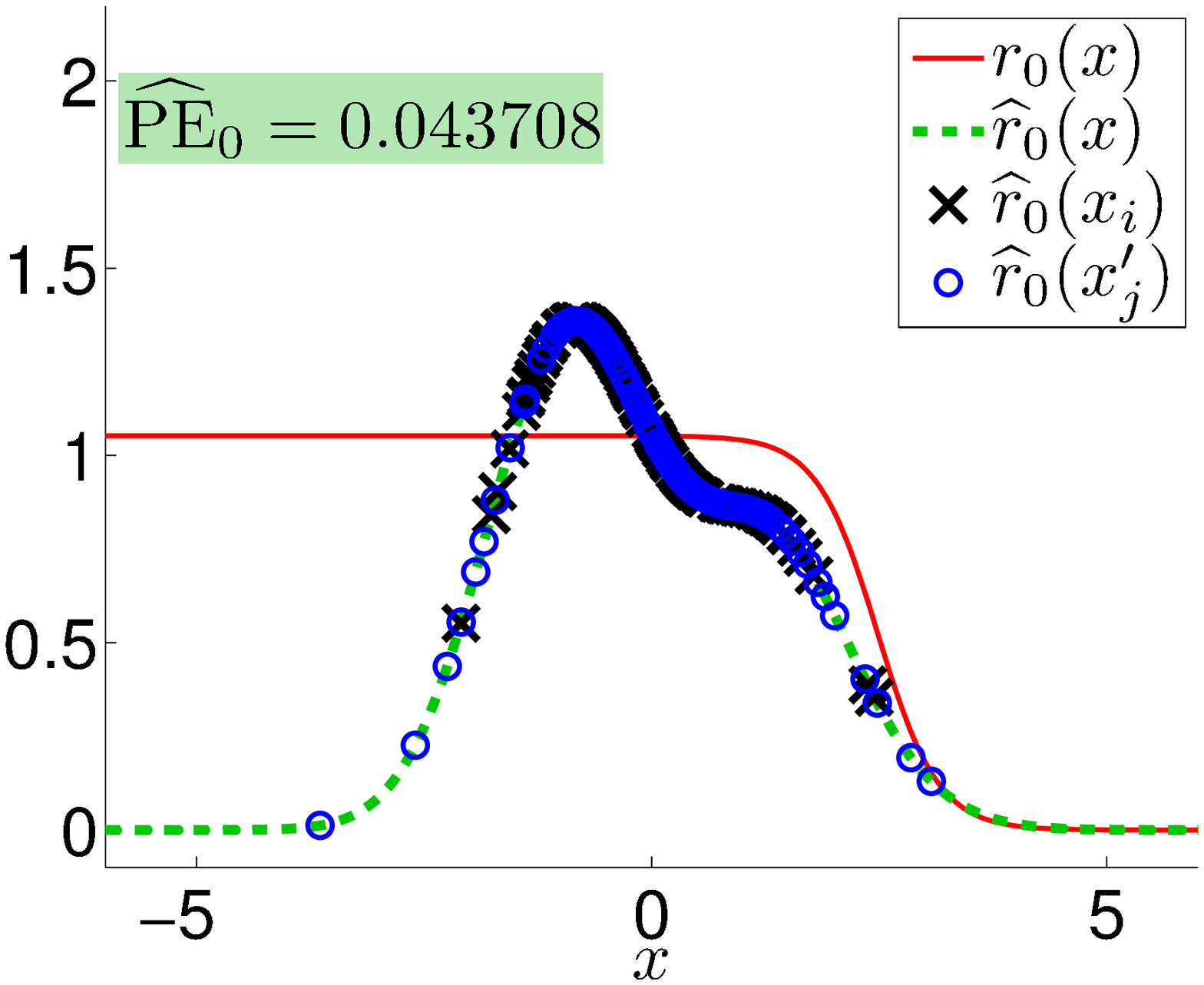}&
  \includegraphics[width=.24\textwidth]{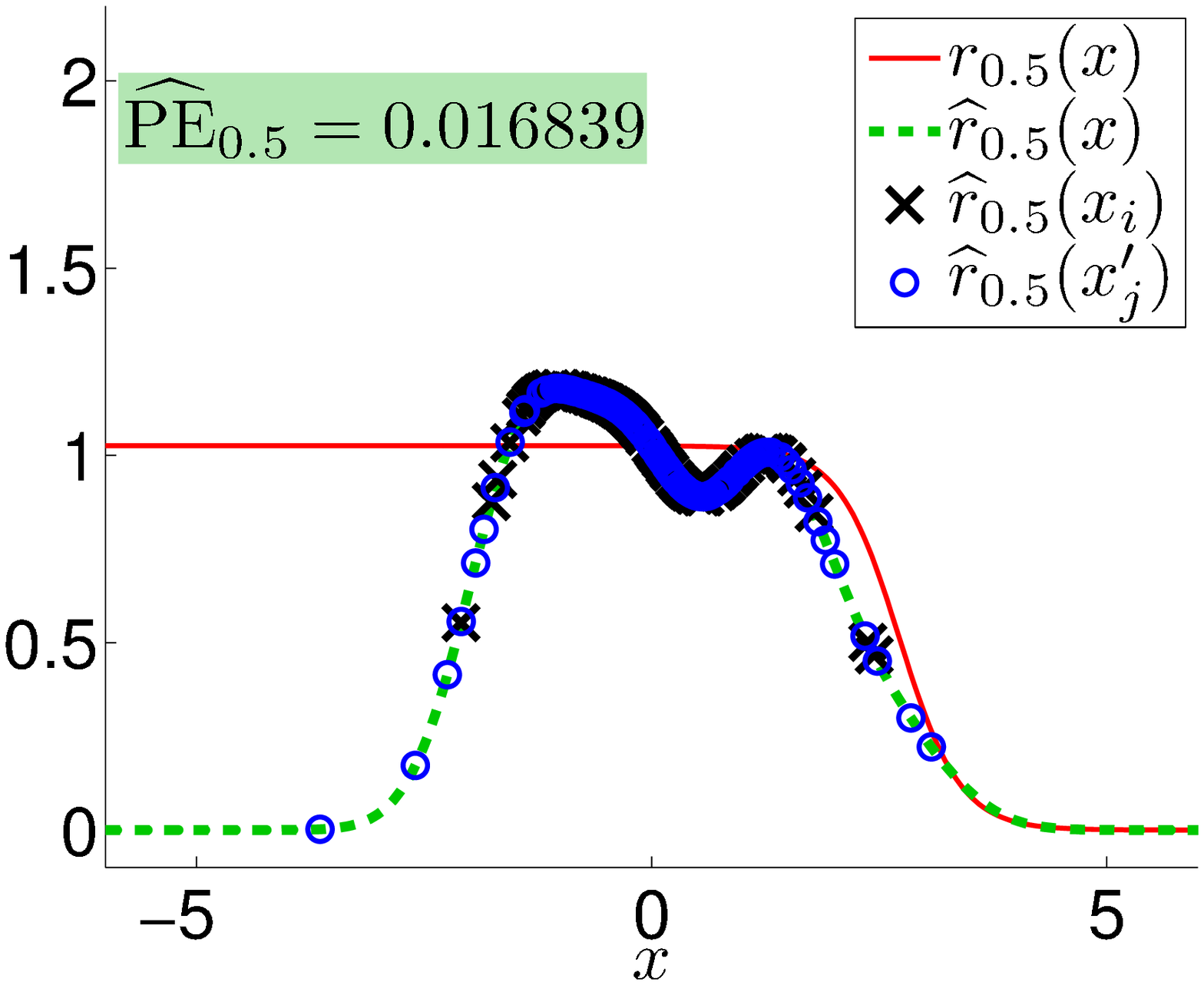}&
  \includegraphics[width=.24\textwidth]{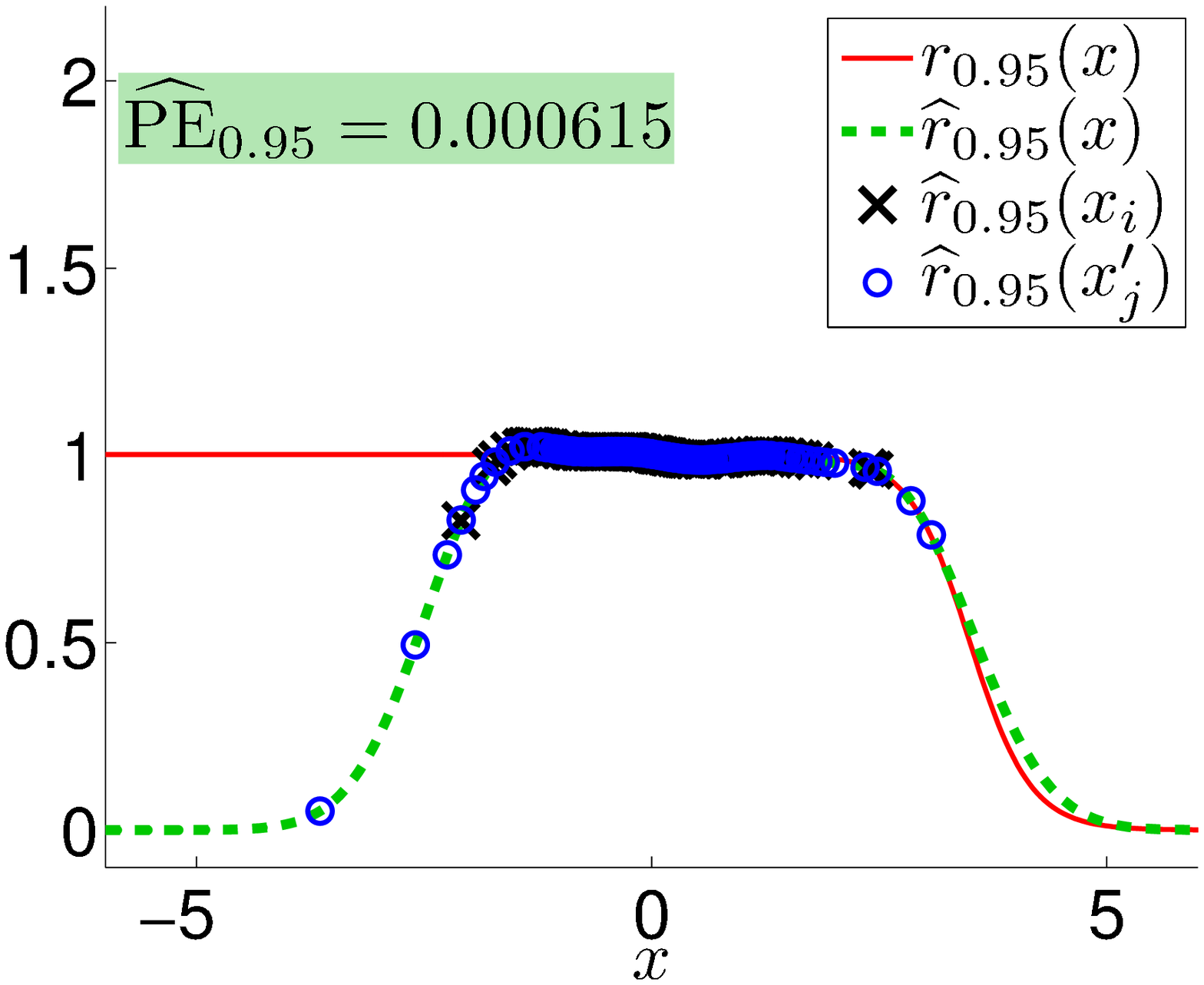}
    \end{tabular}
    \label{fig:illustrative-ratio-mixture}
      }
 \caption{Illustrative examples of density-ratio approximation by RuLSIF.
   From left to right: true densities ($\Pnu = N(0,1)$),
   true density-ratios, and their estimates for $\alpha=0$, $0.5$, and $0.95$.}
    \label{fig:illustrative-ratio}
  \end{figure}

Figure~\ref{fig:illustrative-ratio}
shows the true densities, true density-ratios, and their estimates
by RuLSIF.
As can be seen from the graphs,
the profiles of the true $\alpha$-relative density-ratios
get smoother as $\alpha$ increases.
In particular, in the datasets (b) and (d),
the true density-ratios for $\alpha=0$ diverge to infinity,
while those for $\alpha=0.5$ and $0.95$ are bounded
(by $1/\alpha$).
Overall, as $\alpha$ gets large,
the estimation quality of RuLSIF tends to be improved
since the complexity of true density-ratio functions is reduced.

Note that, in the dataset (a) where $\pnu(\boldx)=\pde(\boldx)$,
the true density-ratio $\relratio(\boldx)$ does not depend on $\alpha$
since $\relratio(\boldx)=1$ for any $\alpha$.
However, the estimated density-ratios still depend on $\alpha$ through
the matrix $\boldHh$ (see Eq.\eqref{Hh}).

\section{Theoretical Analysis}
\label{sec:analysis}
In this section, we analyze theoretical properties of
the proposed PE divergence estimators.
More specifically, we provide non-parametric analysis of the convergence rate
in Section~\ref{subsec:nonpara-analysis},
and parametric analysis of the estimation variance
in Section~\ref{subsec:para-analysis}.
Since our theoretical analysis is highly technical,
we focus on explaining practical insights we can gain from
the theoretical results here;
we describe all the mathematical details
of the non-parametric convergence-rate analysis
in Appendix~\ref{appendix:nonparametric}
and
the parametric variance analysis
in Appendix~\ref{appendix:parametric}.

For theoretical analysis,
let us consider a rather abstract form of our relative density-ratio estimator described as
\begin{align}
\mathop{\mathrm{argmin}}_{g\in \calG} 
\left[
\frac{\alpha}{2\nnu}\sum_{i=1}^{\nnu} \ratiomodel(\boldxnu_i)^2
  + \frac{(1-\alpha)}{2\nde} \sum_{j=1}^{\nde}\ratiomodel(\boldxde_j)^2
-\frac{1}{\nnu}\sum_{i=1}^{\nnu} \ratiomodel(\boldxnu_i)
 +  \frac{\lambda}{2} R(\ratiomodel)^2
\right],
\label{alpha-uLSIF-general}
\end{align}
where $\calG$ is some function space (i.e., a statistical model)
and $R(\cdot)$ is some regularization functional.

\subsection{Non-Parametric Convergence Analysis}
\label{subsec:nonpara-analysis}
First, we elucidate the non-parametric convergence rate of
the proposed PE estimators.
Here, we practically regard the function space $\calG$ as an infinite-dimensional
\emph{reproducing kernel Hilbert space}
\citep[RKHS;][]{AMS:Aronszajn:1950} such as the Gaussian kernel space,
and $R(\cdot)$ as the associated RKHS norm.

\subsubsection{Theoretical Results}
Let us represent the complexity of the function space $\calG$ 
by $\gamma$ ($0<\gamma<2$);
the larger $\gamma$ is, the more complex the function class $\calG$ is
(see Appendix~\ref{appendix:nonparametric} for its precise definition).
We analyze the convergence rate of our PE divergence estimators
as $\bar{n}:=\min(\np,\nq)$ tends to infinity
for $\lambda=\lambda_{\bar{n}}$
under 
\begin{align*}
  \lambda_{\bar{n}} \to o(1)
  ~~\mbox{and}~~
  \lambda_{\bar{n}}^{-1} = o(\bar{n}^{2/(2+\gamma)}).
\end{align*}
The first condition means that $\lambda_{\bar{n}}$ tends to zero,
but the second condition means that its shrinking speed should not be too fast.

Under several technical assumptions detailed in Appendix~\ref{appendix:nonparametric},
we have the following asymptotic convergence results
for the two PE divergence estimators
$\widehat{\mathrm{PE}}_\alpha$ \eqref{eq:PE1}
and $\widetilde{\mathrm{PE}}_\alpha$ \eqref{eq:PE2}:
\begin{align}
\widehat{\mathrm{PE}}_\alpha-\mathrm{PE}_\alpha
= \calO_p(\bar{n}^{-1/2}c\|\relratio\|_{\infty}+\lambda_{\bar{n}}\max(1,R(\relratio)^2)), 
\label{nonpara-rate-PEhat}
\end{align}
and 
\begin{align}
\widetilde{\mathrm{PE}}_\alpha-\mathrm{PE}_\alpha
&=
\calO_p\Big( 
\lambda_{\bar{n}}^{{1}/{2}}\|\relratio\|_{\infty}^{{1}/{2}}\max\{1, R(\relratio)\}
\nonumber\\
&\phantom{=}~~~~~~~~+
\lambda_{\bar{n}}\max\{1,\|\relratio\|_{\infty}^{(1-{\gamma}/{2})/2},R(\relratio)\|\relratio\|_{\infty}^{(1-{\gamma}/{2})/2} ,R(\relratio)\}
\Big),
\label{nonpara-rate-PEtilde}
\end{align}
where $\calO_p$ denotes the asymptotic order in probability,
\begin{align}
  c &:= (1+\alpha) \sqrt{\mathbbV_{\pnu(\boldx)}[\relratio(\boldx)]}
  +(1-\alpha) \sqrt{\mathbbV_{\pde(\boldx)}[\relratio(\boldx)]},
  \label{nonpara-c}
\end{align}
and $\mathbbV_{p(\boldx)}[f(\boldx)]$ denotes the variance of $f(\boldx)$ under $p(\boldx)$:
\begin{align*}
  \mathbbV_{p(\boldx)}[f(\boldx)]=
  \int\left(f(\boldx)-\int f(\boldx)p(\boldx)\mathrm{d}\boldx\right)^2
  p(\boldx)\mathrm{d}\boldx.
\end{align*}

\subsubsection{Interpretation}
In both Eq.\eqref{nonpara-rate-PEhat} and Eq.\eqref{nonpara-rate-PEtilde},
the coefficients of the leading terms (i.e., the first terms)
of the asymptotic convergence rates
become smaller as $\|\relratio\|_{\infty}$ gets smaller.
Since
\begin{align*}
\textstyle
  \|\relratio\|_{\infty}
  =\left\|\big(\alpha +(1-\alpha)/\ratio(\boldx)\big)^{-1}\right\|_{\infty}
  <\frac{1}{\alpha}~~~\mbox{for }\alpha>0,
\end{align*}
larger $\alpha$ would be more preferable
in terms of the asymptotic approximation error.
Note that when $\alpha=0$, $\|\relratio\|_{\infty}$ can tend to infinity
even under a simple setting that the ratio of two Gaussian functions is considered
\citep[][see also the numerical examples in Section~\ref{subsec:illustration} of this paper]{NIPS2010_0731}.
Thus, our proposed approach of estimating the $\alpha$-relative PE divergence
(with $\alpha>0$) would be more advantageous than the naive approach of
estimating the plain PE divergence (which corresponds to $\alpha=0$)
in terms of the non-parametric convergence rate.

The above results also show that
$\widehat{\mathrm{PE}}_\alpha$ and $\widetilde{\mathrm{PE}}_\alpha$
have different asymptotic convergence rates.
The leading term in Eq.\eqref{nonpara-rate-PEhat}
is of order ${\bar{n}}^{-1/2}$,
while the leading term in Eq.\eqref{nonpara-rate-PEtilde}
is of order $\lambda_{\bar{n}}^{1/2}$,
which is slightly slower (depending on the complexity $\gamma$) than ${\bar{n}}^{-1/2}$.
Thus, $\widehat{\mathrm{PE}}_\alpha$ would be
more accurate than $\widetilde{\mathrm{PE}}_\alpha$ in large sample cases.
Furthermore, when $\pnu(\boldx)=\pde(\boldx)$,
$\mathbbV_{\pnu(\boldx)}[\relratio(\boldx)]=0$ holds and thus $c=0$ holds
(see Eq.\eqref{nonpara-c}).
Then the leading term in Eq.\eqref{nonpara-rate-PEhat}
vanishes and therefore $\widehat{\mathrm{PE}}_\alpha$ 
has the even faster convergence rate of order $\lambda_{\bar{n}}$,
which is slightly slower (depending on the complexity $\gamma$) than ${\bar{n}}^{-1}$.
Similarly, if $\alpha$ is close to $1$,
$\relratio(\boldx)\approx 1$ and thus $c\approx0$ holds.

When ${\bar{n}}$ is not large enough to be able to neglect the terms of
$o({\bar{n}}^{-1/2})$, the terms of $O(\lambda_{\bar{n}})$ matter.
If $\|\relratio\|_{\infty}$ and $R(\relratio)$ are large
(this can happen, e.g., when $\alpha$ is close to $0$),
the coefficient of the $O(\lambda_{\bar{n}})$-term in Eq.\eqref{nonpara-rate-PEhat}
can be larger than that in Eq.\eqref{nonpara-rate-PEtilde}.
Then $\widetilde{\mathrm{PE}}_\alpha$ would be more favorable than
$\widehat{\mathrm{PE}}_\alpha$ in terms of the approximation accuracy.


\begin{figure}[p]
  \centering
  \subfigure[$\Pde=N(0,1)$: $\Pnu$ and $\Pde$ are the same.]{
    \begin{tabular}{@{}c@{}c@{}c@{}c@{}}
 \includegraphics[width=.24\textwidth]{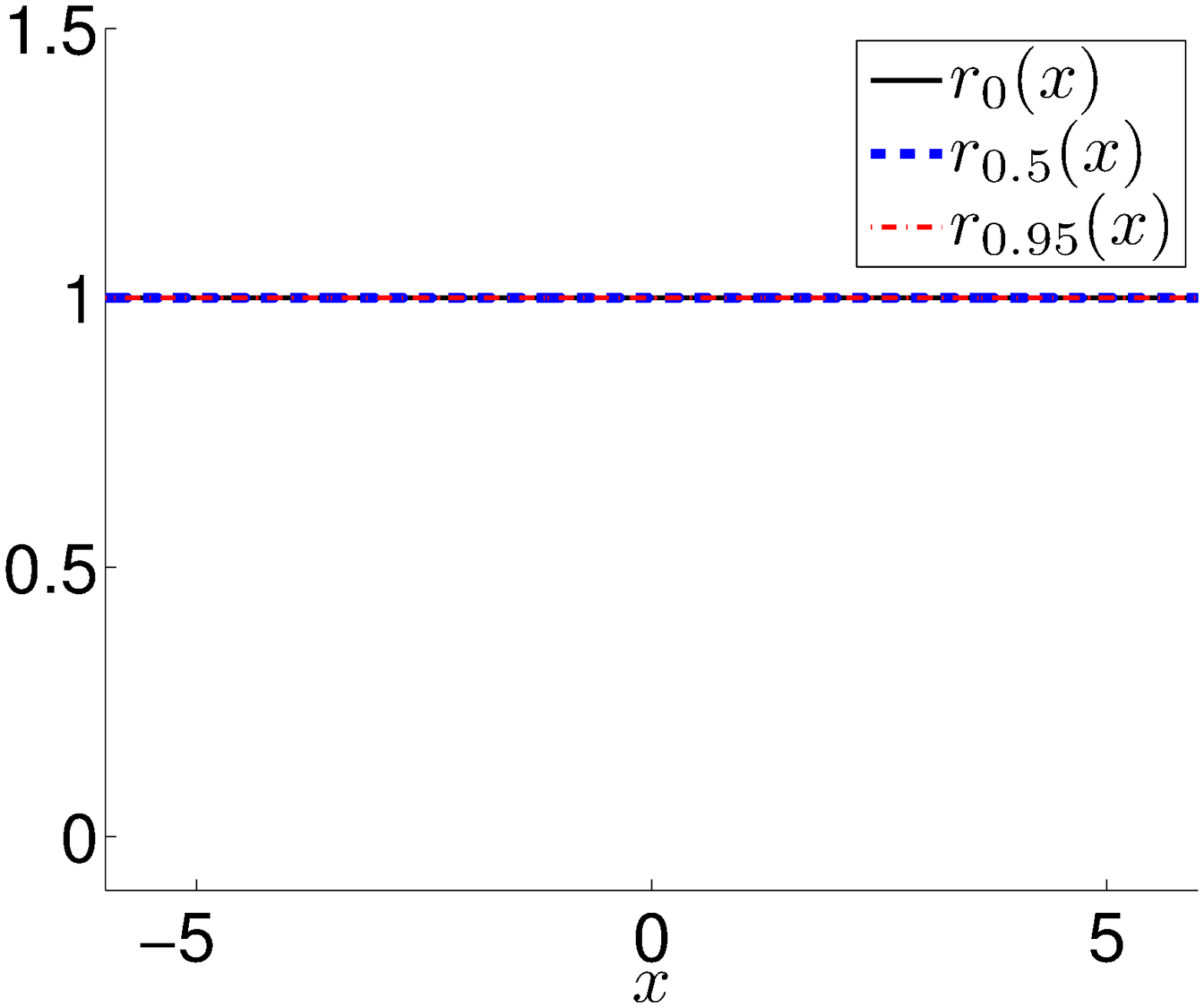}&
 \includegraphics[width=.24\textwidth]{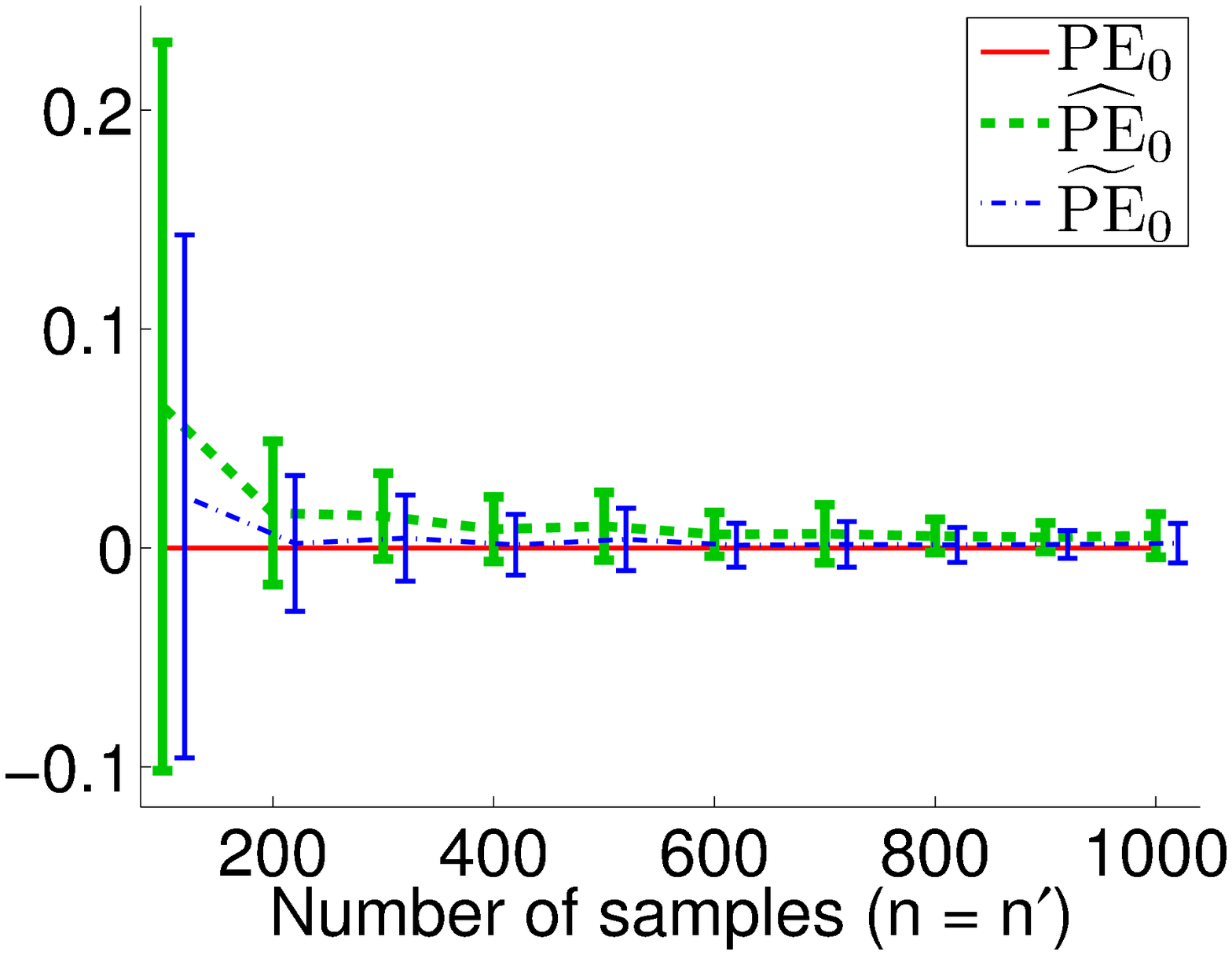}&
 \includegraphics[width=.24\textwidth]{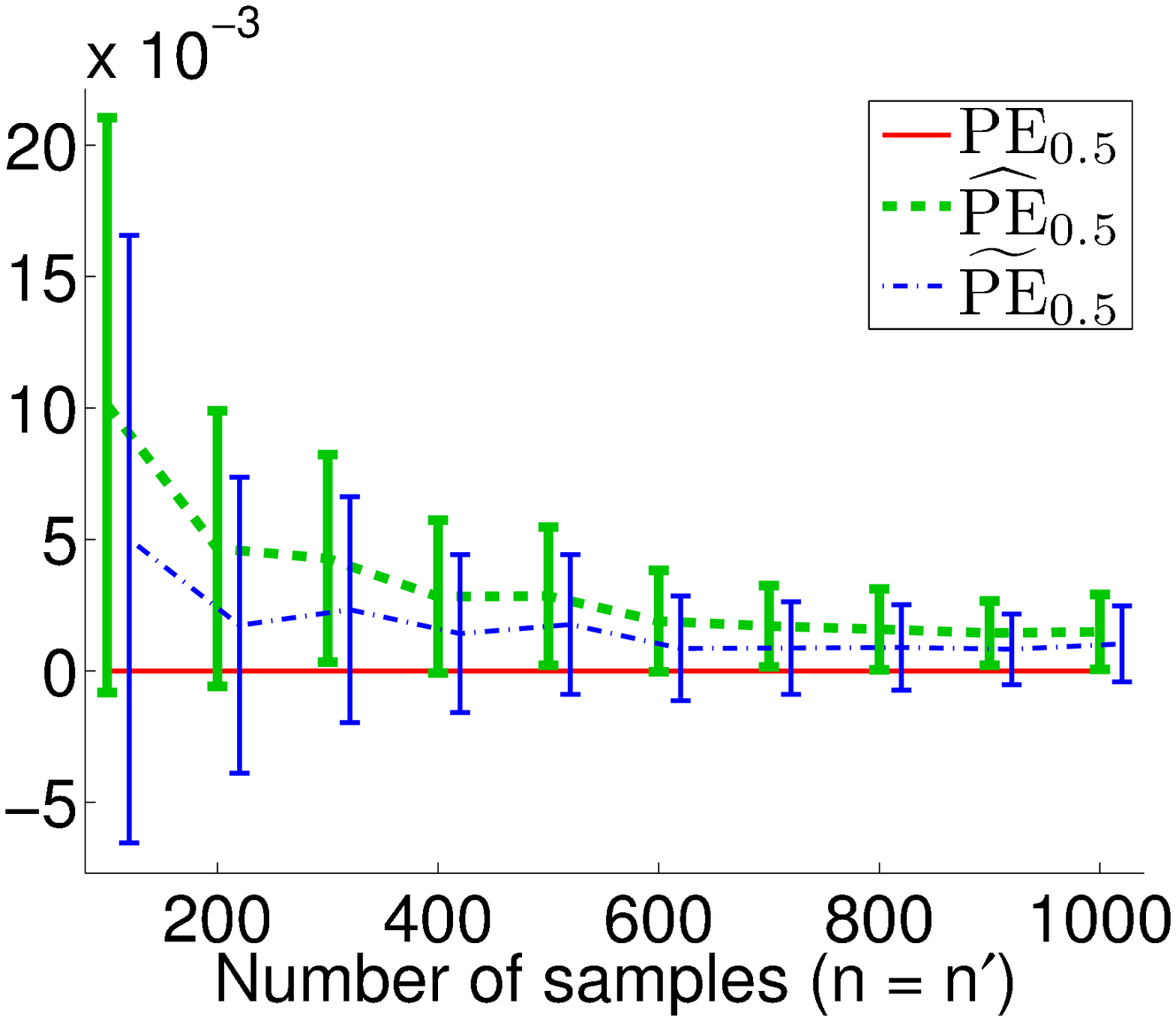}&
 \includegraphics[width=.24\textwidth]{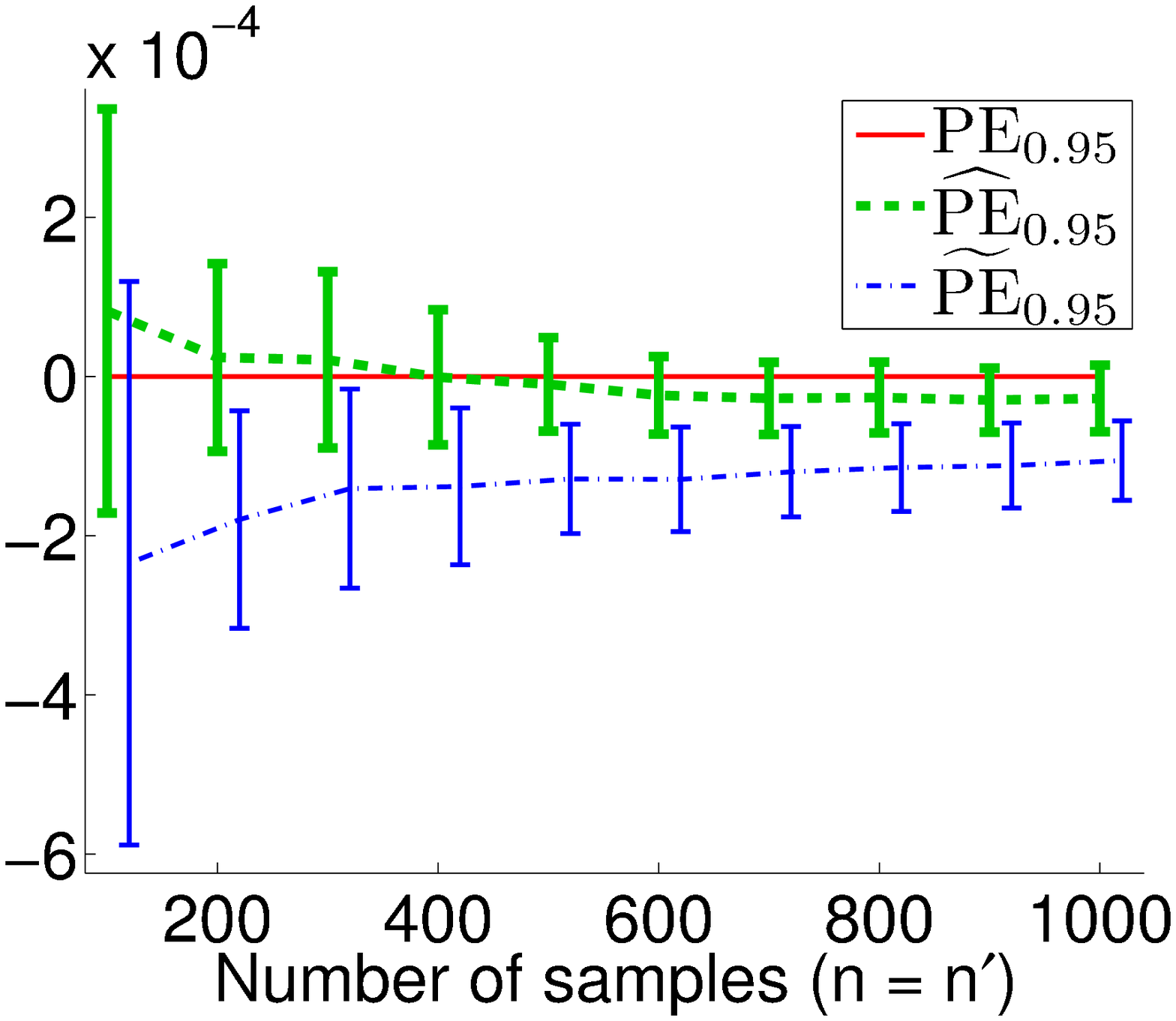}
    \end{tabular}
      }
 \subfigure[$\Pde=N(0,0.6)$: $\Pde$ has smaller standard deviation than $\Pnu$.]{
    \begin{tabular}{@{}c@{}c@{}c@{}c@{}}
 \includegraphics[width=.24\textwidth]{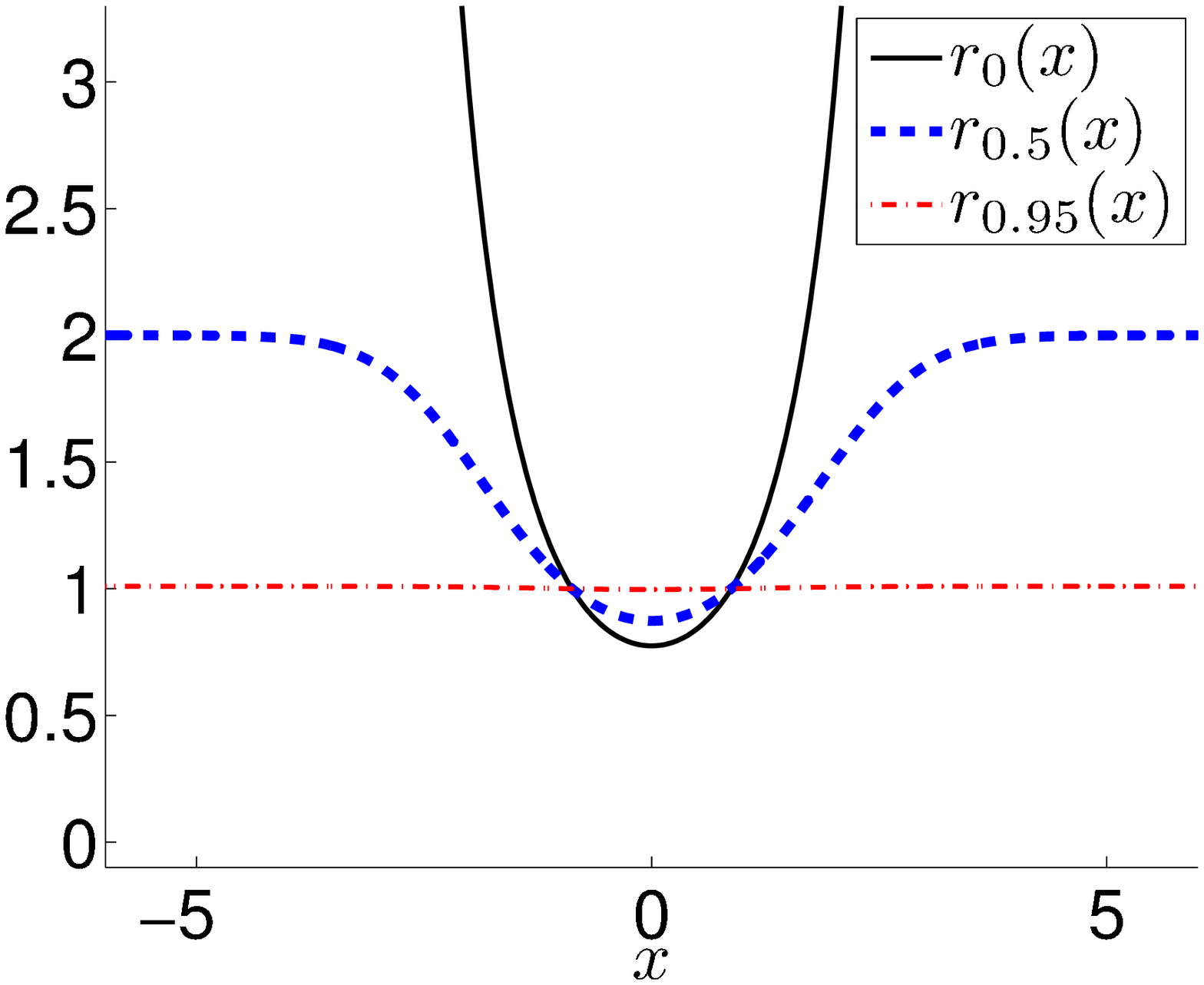}&
 \includegraphics[width=.24\textwidth]{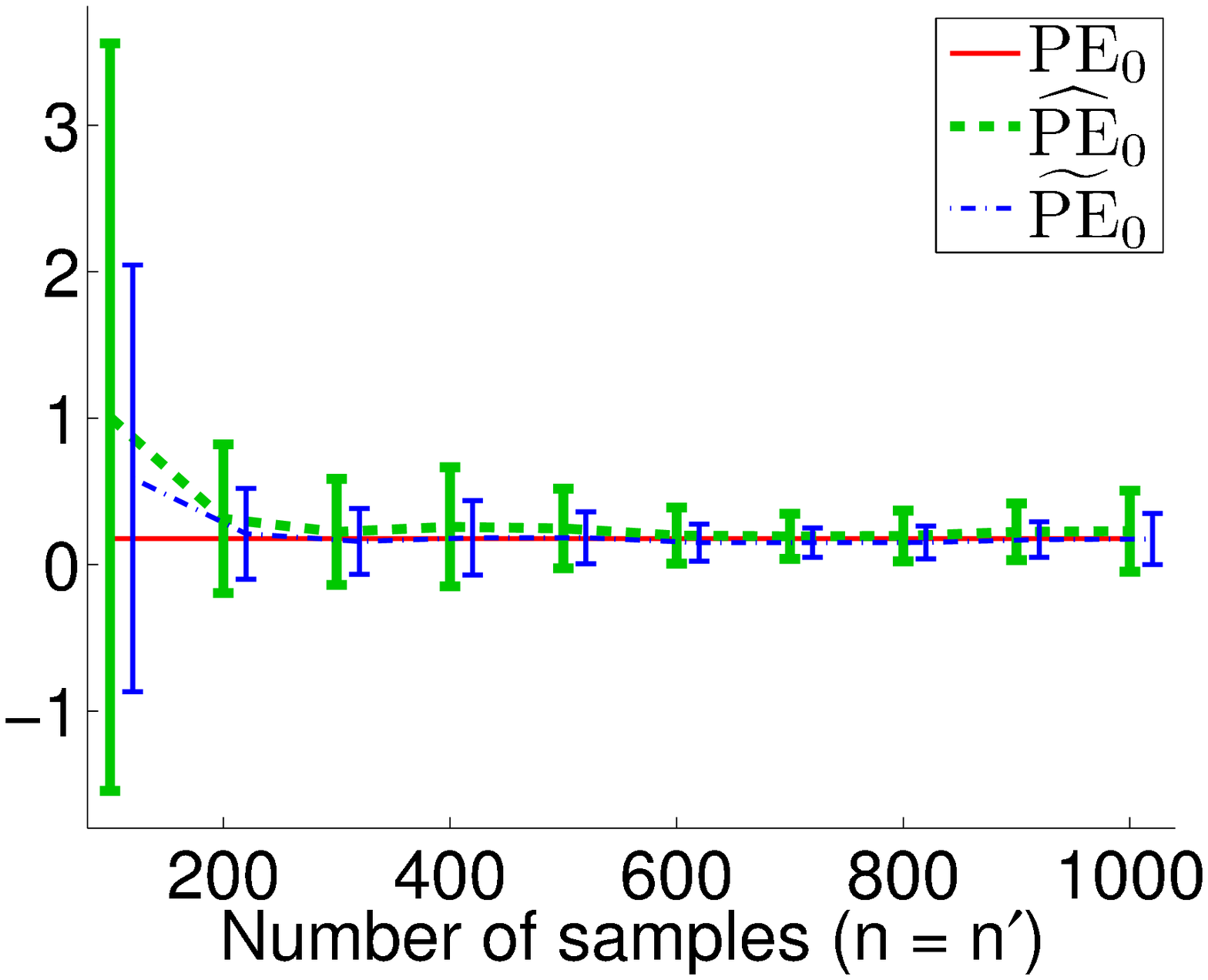}&
 \includegraphics[width=.24\textwidth]{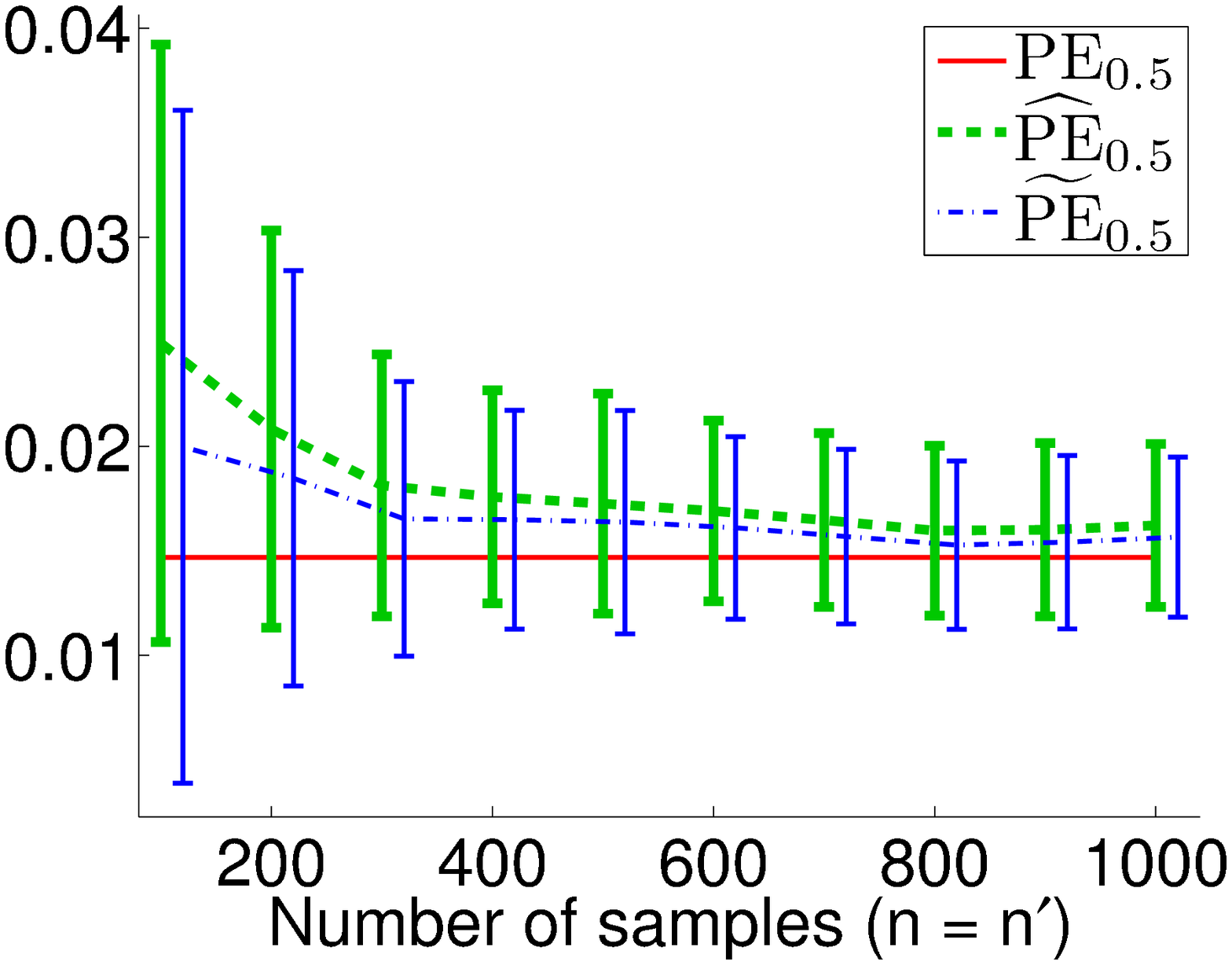}&
 \includegraphics[width=.24\textwidth]{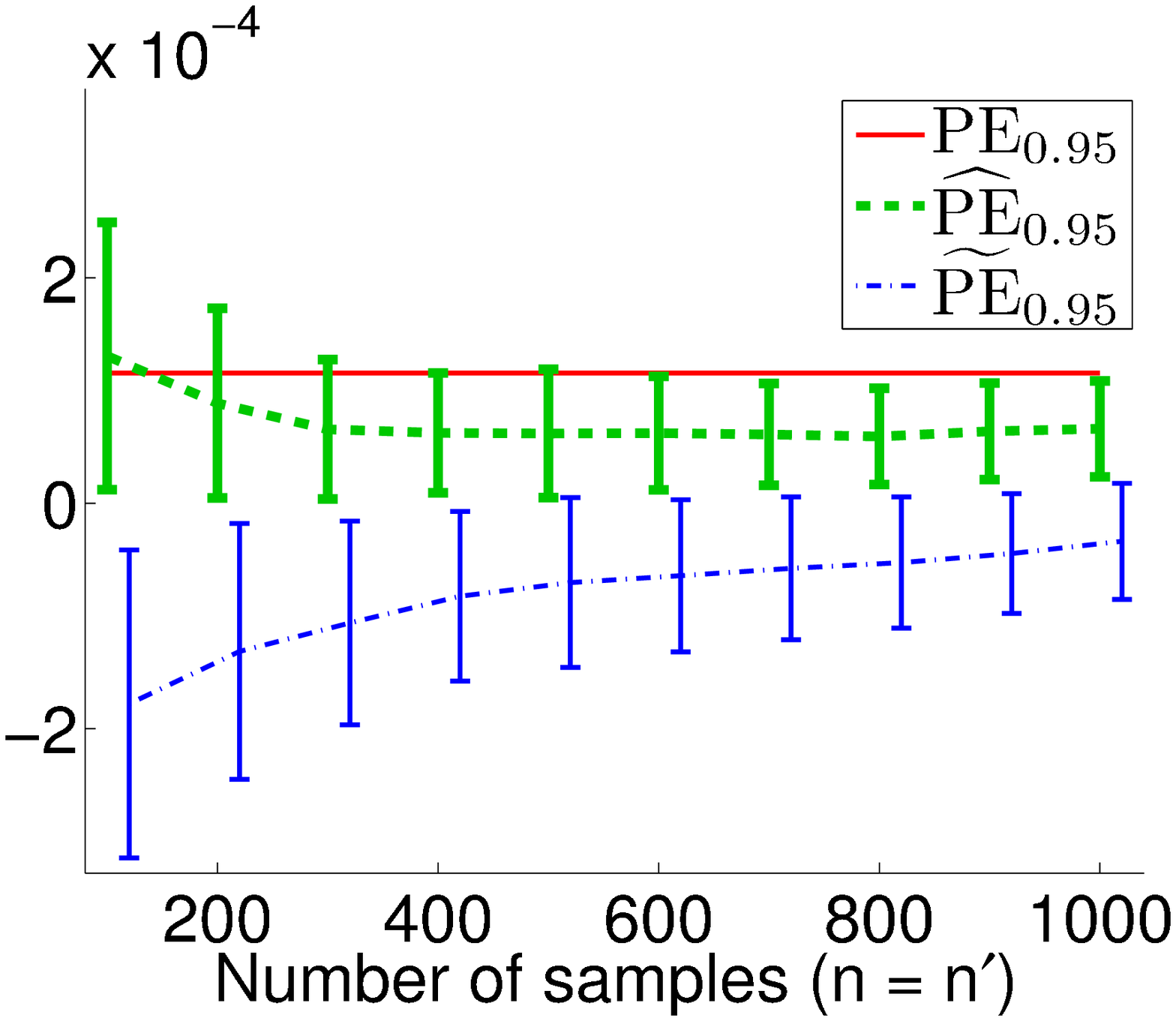}
    \end{tabular}
      }
 \subfigure[$\Pde=N(0,2)$: $\Pde$ has larger standard deviation than $\Pnu$.]{
    \begin{tabular}{@{}c@{}c@{}c@{}c@{}}
 \includegraphics[width=.24\textwidth]{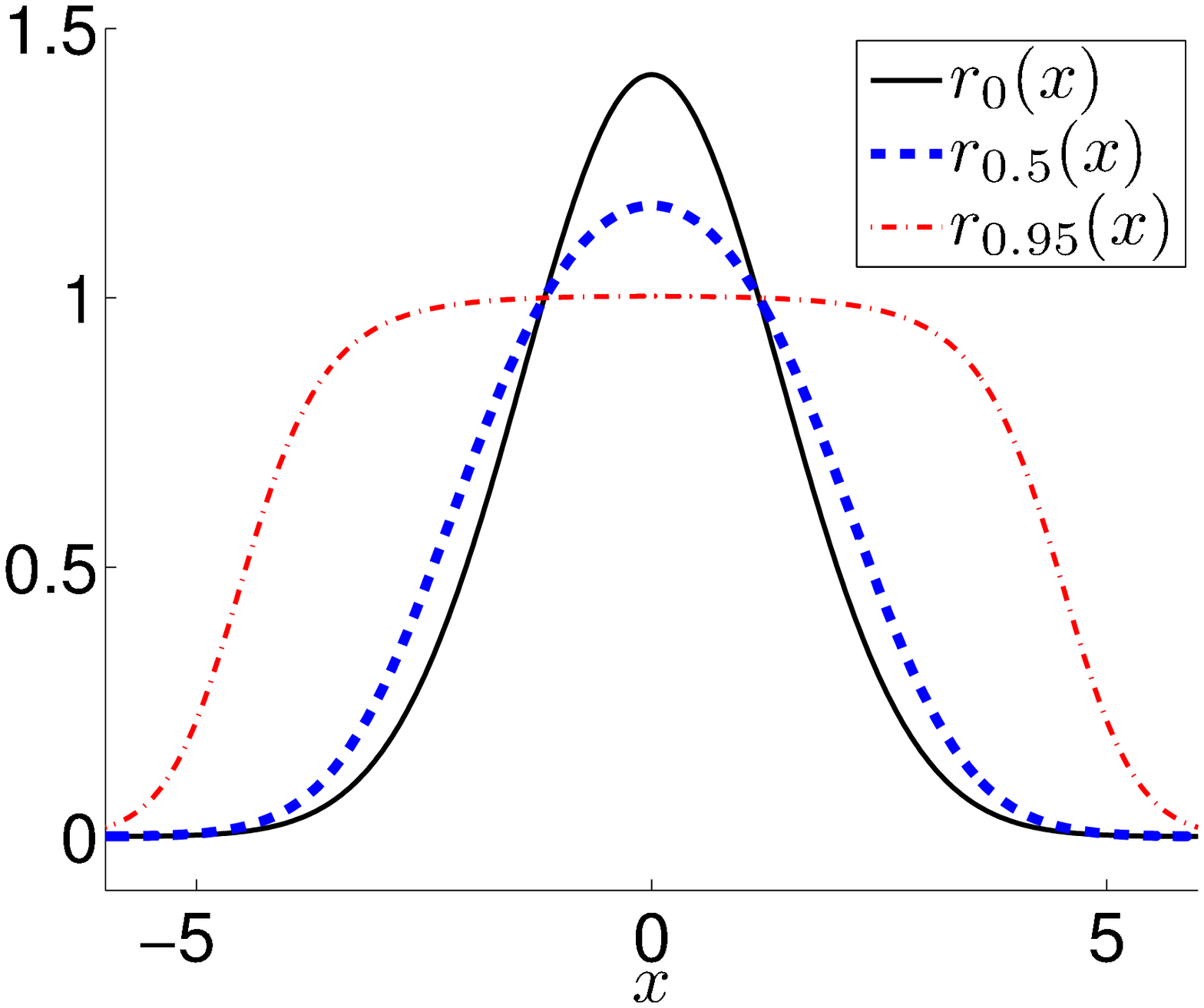}&
 \includegraphics[width=.24\textwidth]{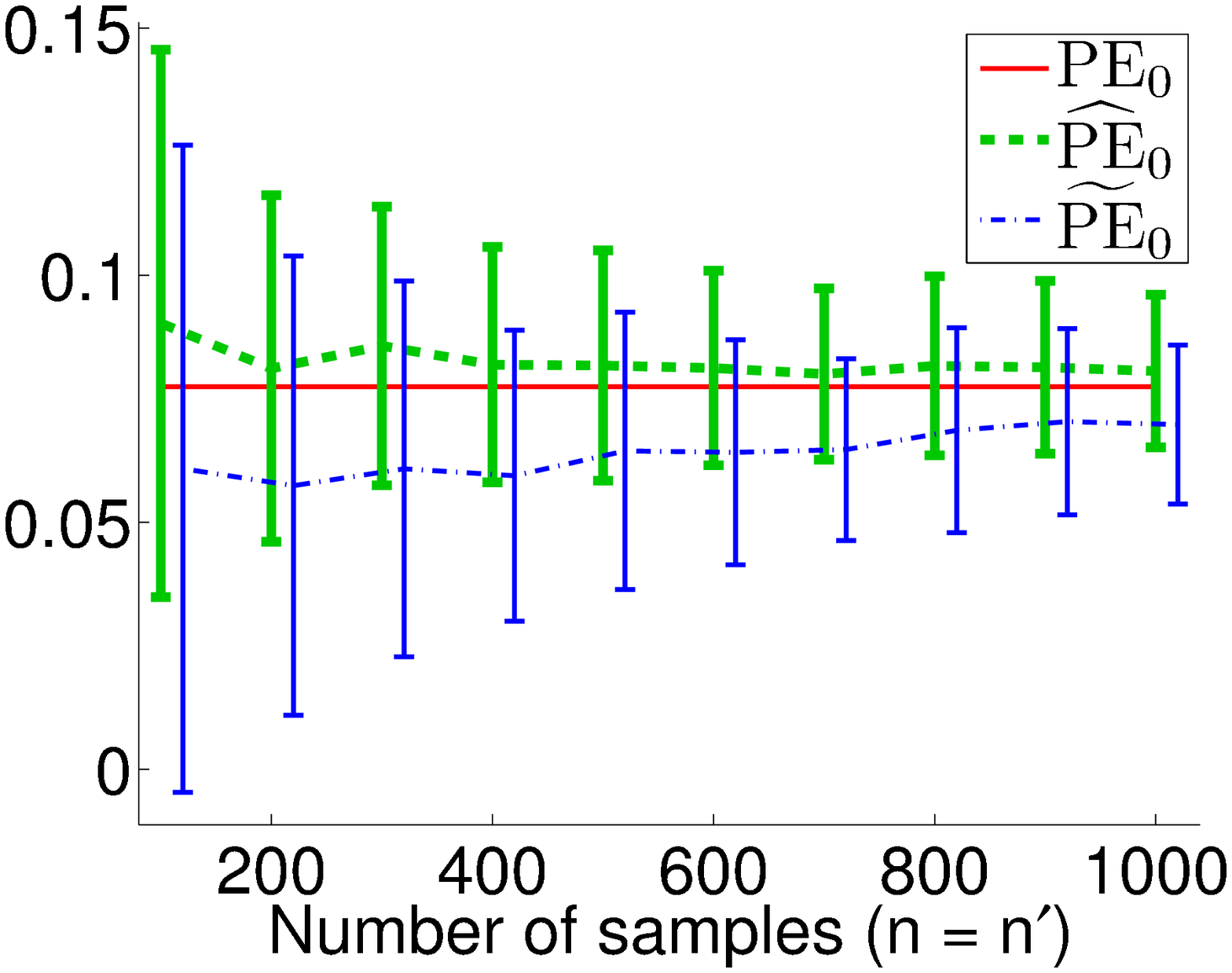}&
 \includegraphics[width=.24\textwidth]{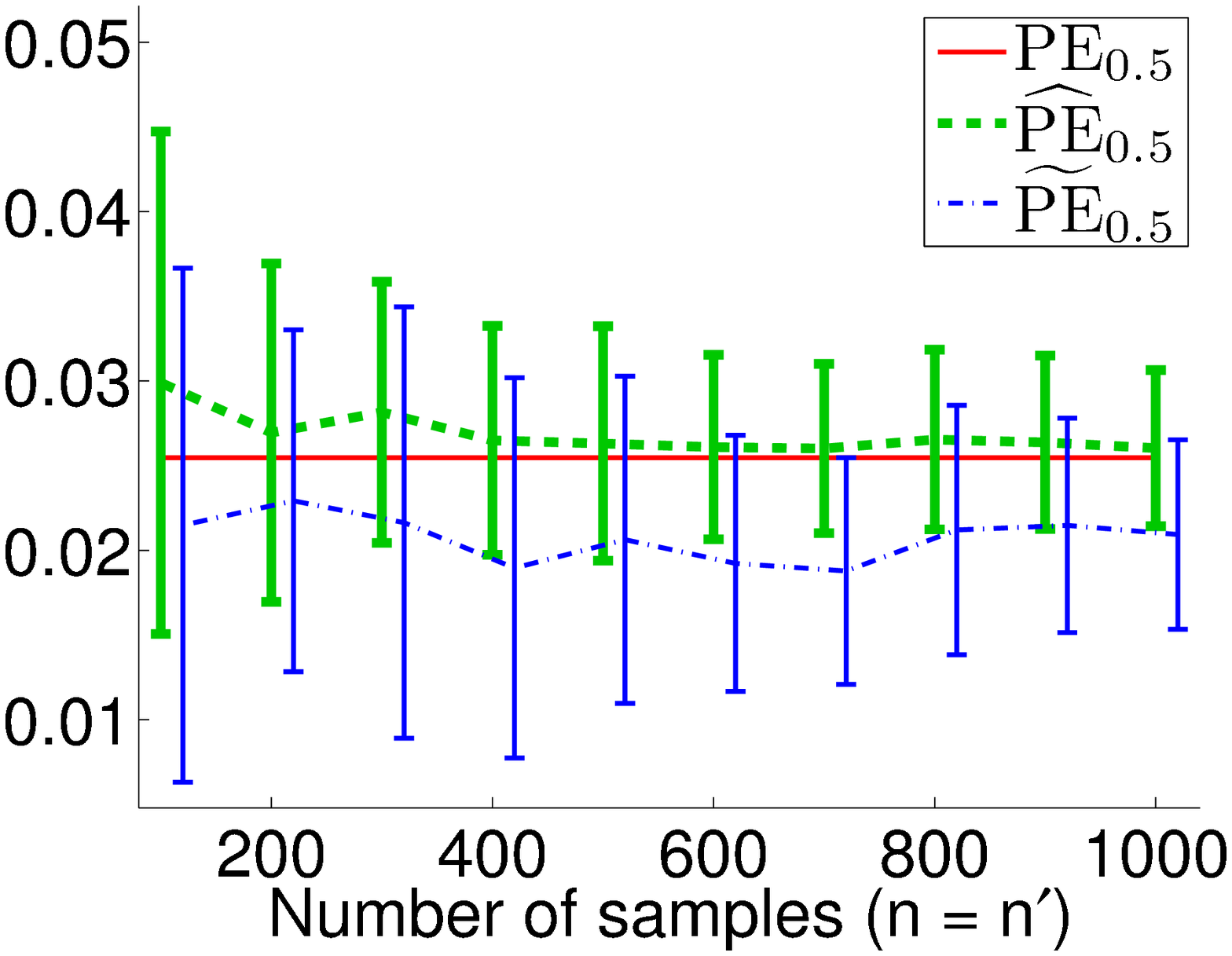}&
 \includegraphics[width=.24\textwidth]{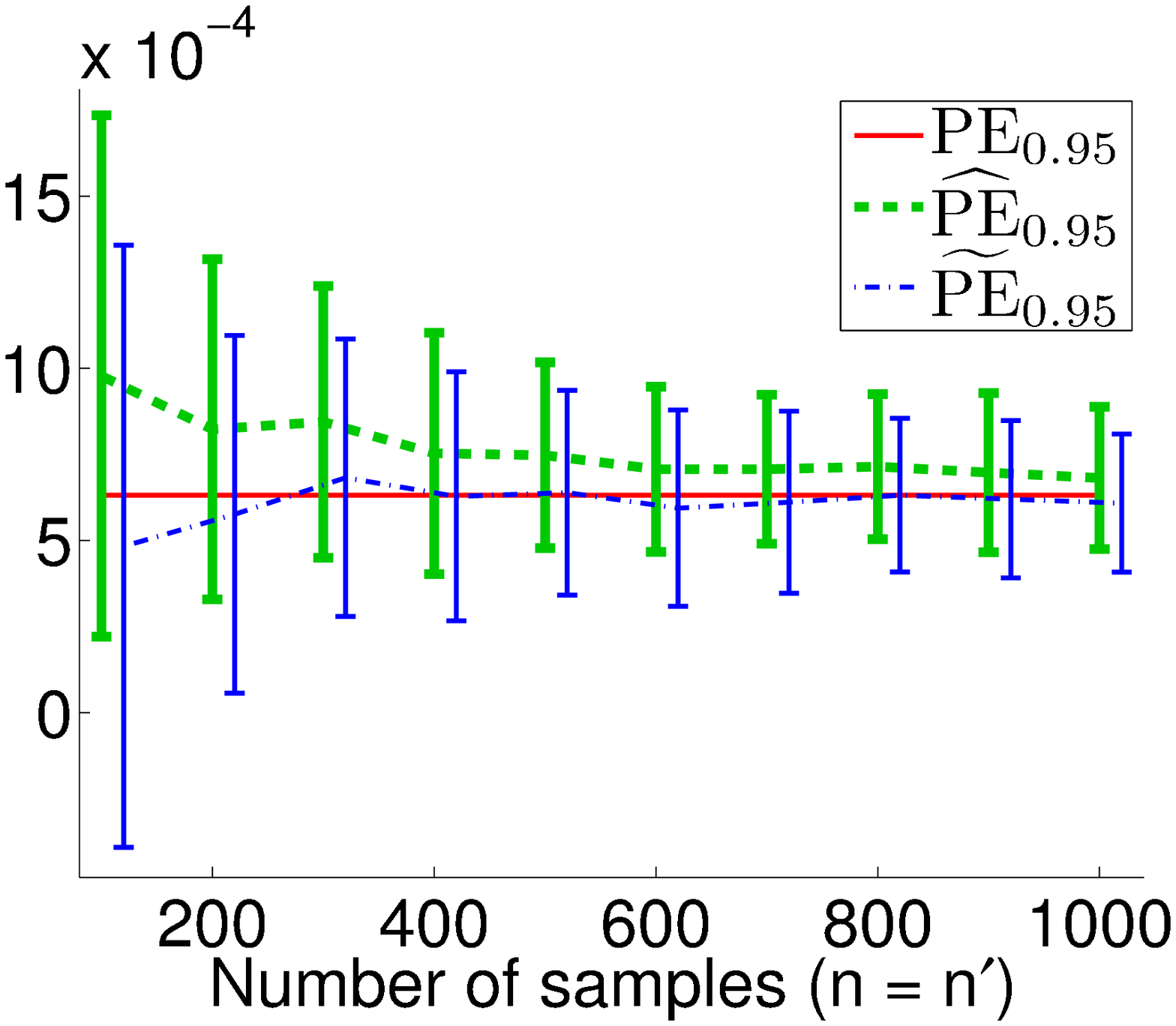}
    \end{tabular}
      }
 \subfigure[$\Pde=N(0.5,1)$: $\Pnu$ and $\Pde$ have different means.]{
    \begin{tabular}{@{}c@{}c@{}c@{}c@{}}
 \includegraphics[width=.24\textwidth]{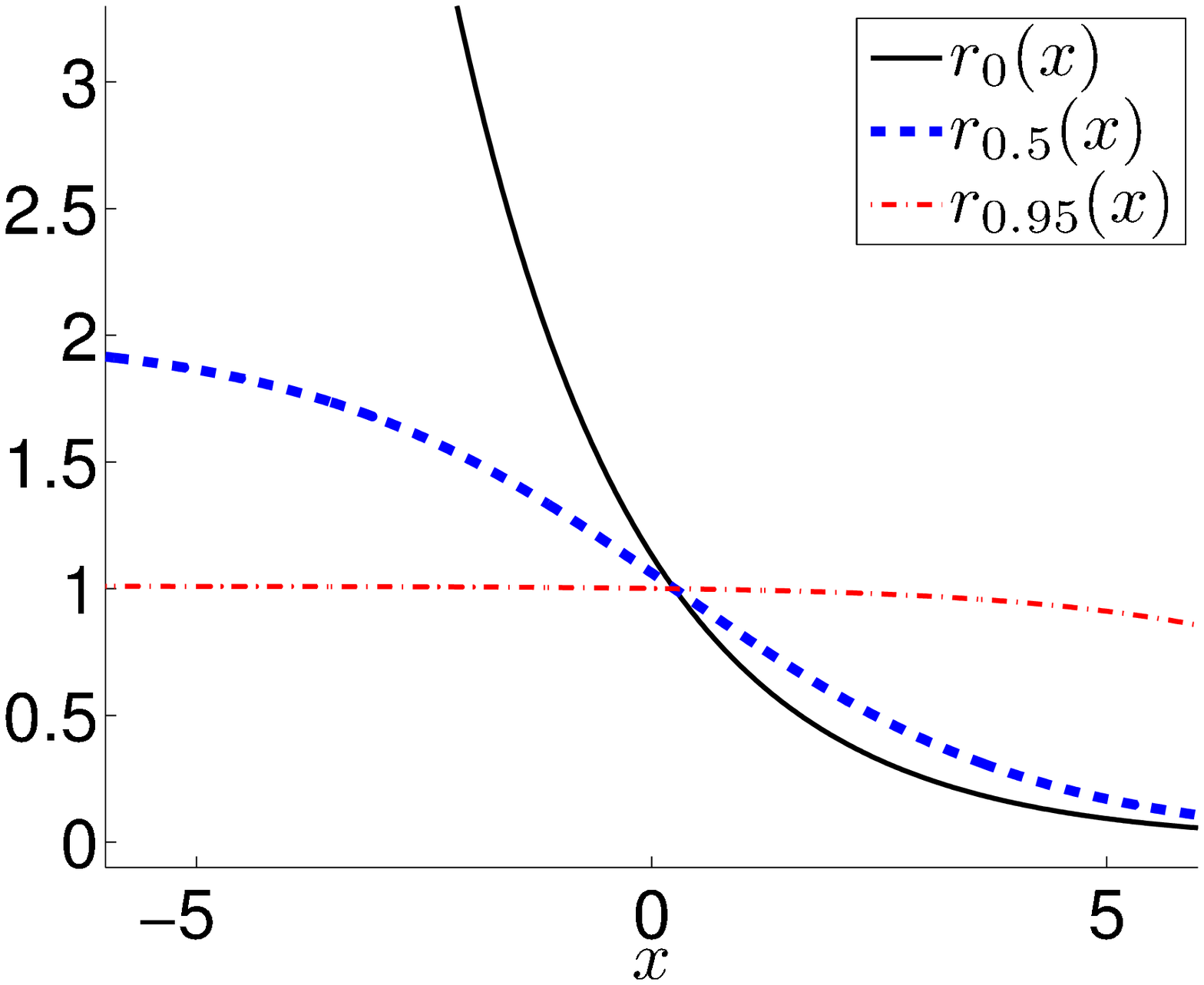}&
 \includegraphics[width=.24\textwidth]{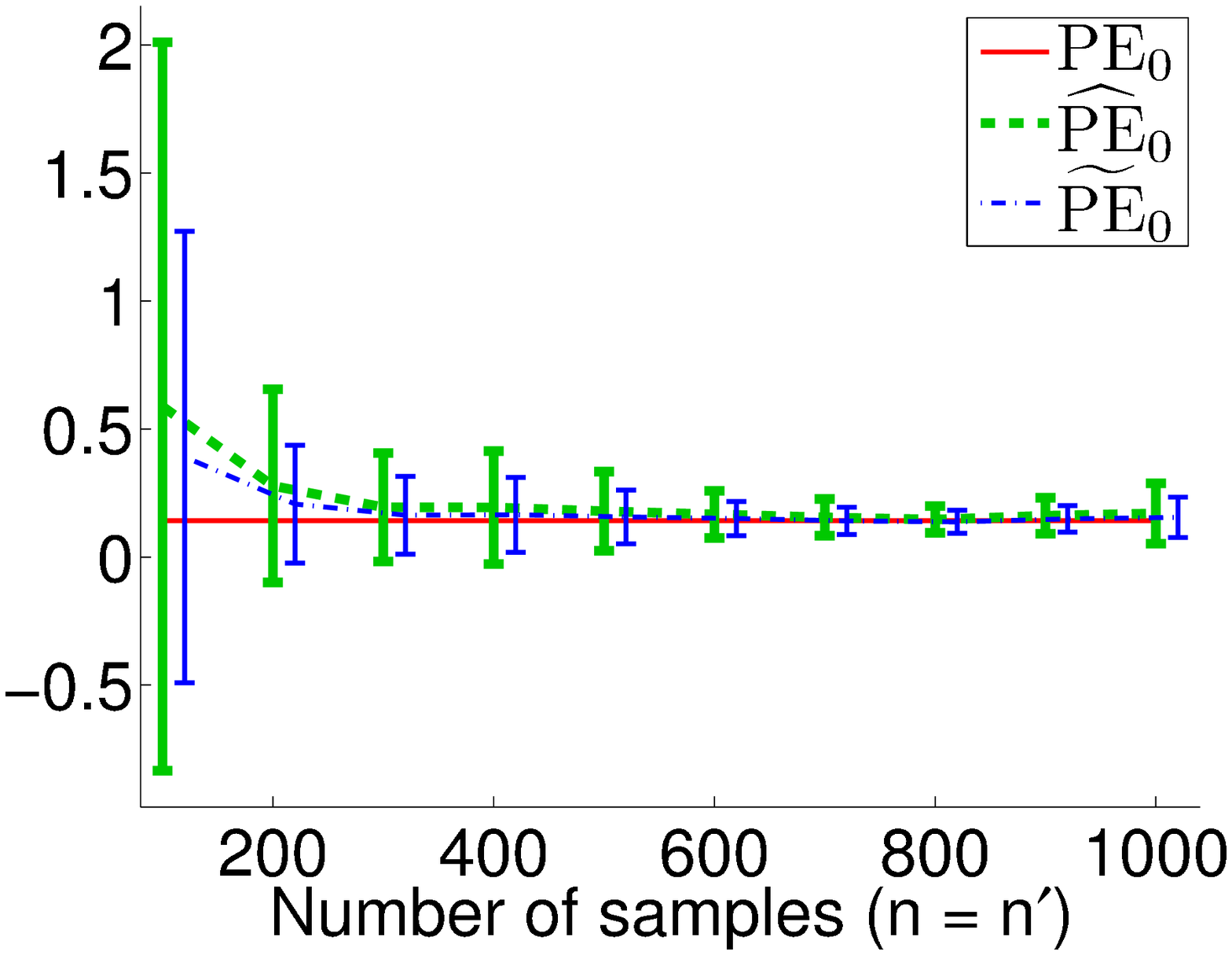}&
 \includegraphics[width=.24\textwidth]{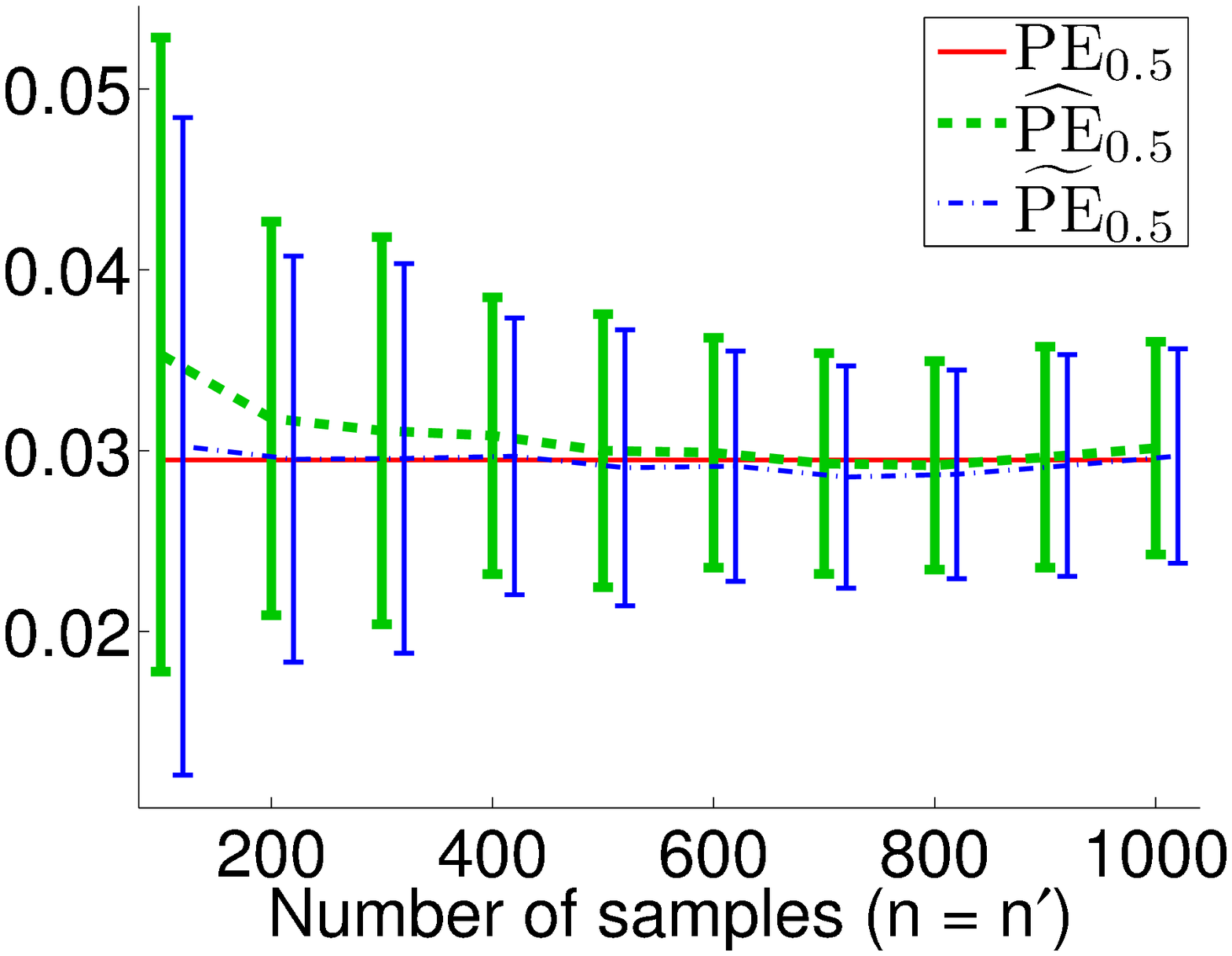}&
 \includegraphics[width=.24\textwidth]{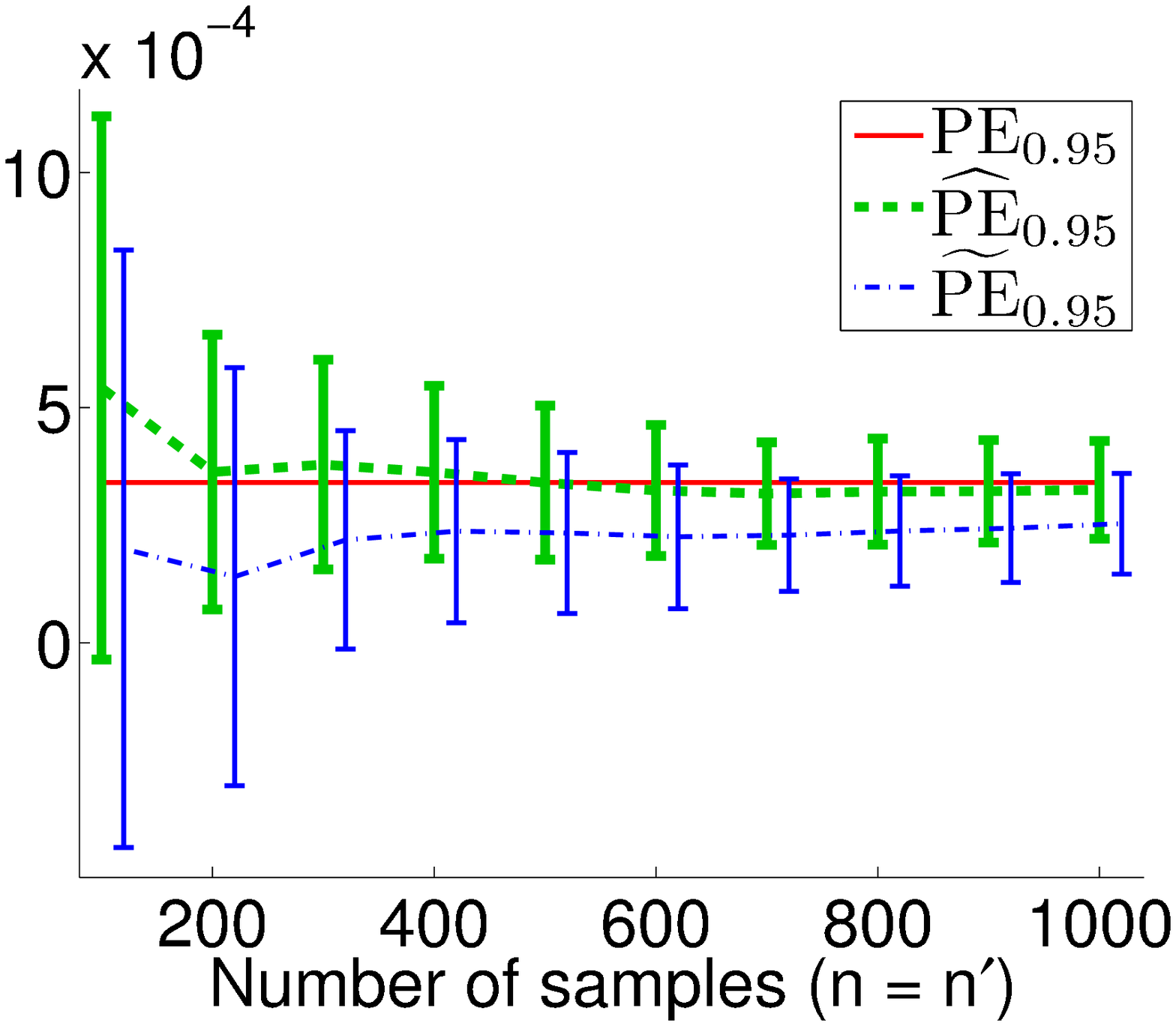}
    \end{tabular}
      }
 \subfigure[$\Pde=0.95N(0,1)+0.05N(3,1)$: $\Pde$ contains an additional component to $\Pnu$.]{
    \begin{tabular}{@{}c@{}c@{}c@{}c@{}}
  \includegraphics[width=.24\textwidth]{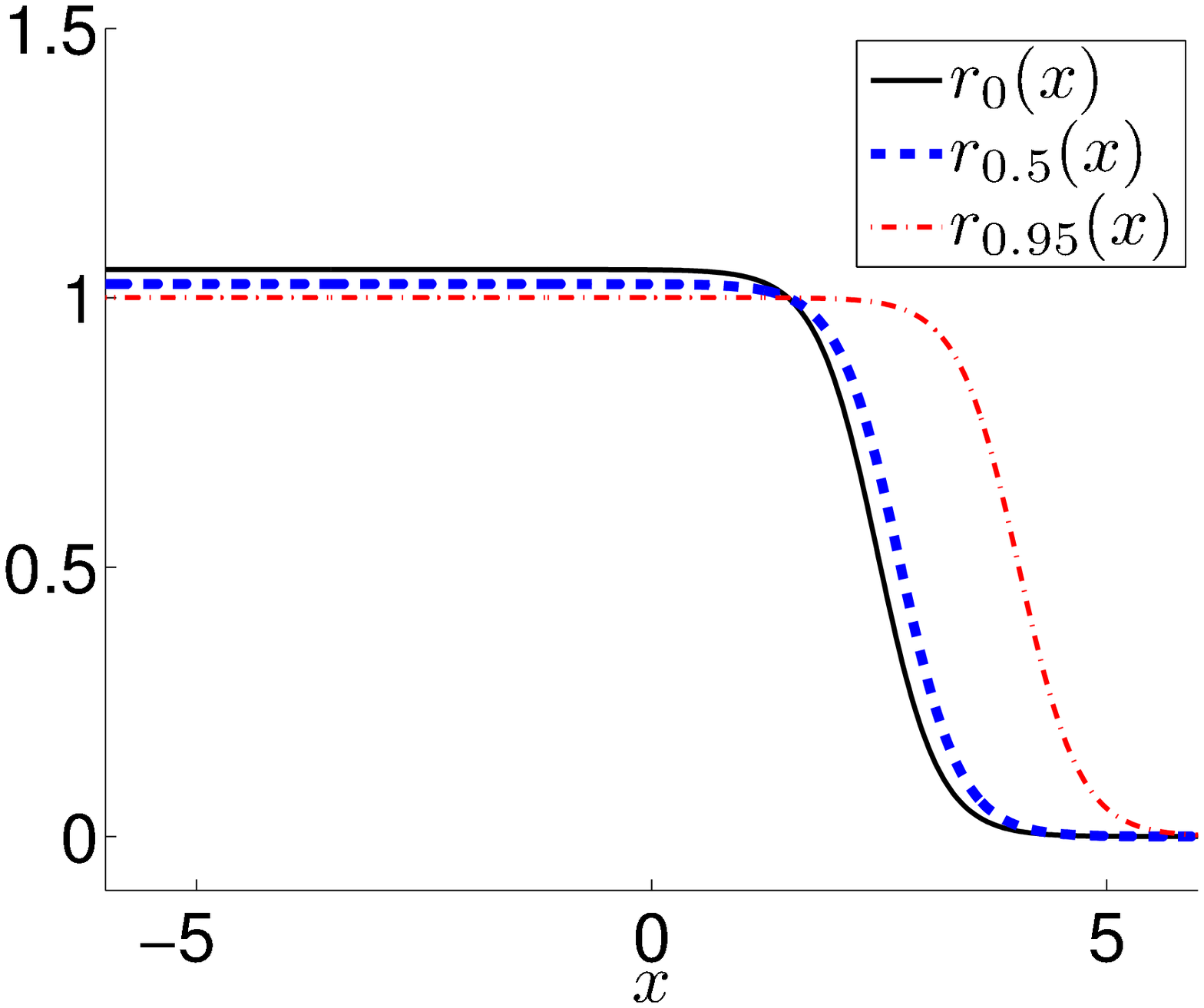}&
  \includegraphics[width=.24\textwidth]{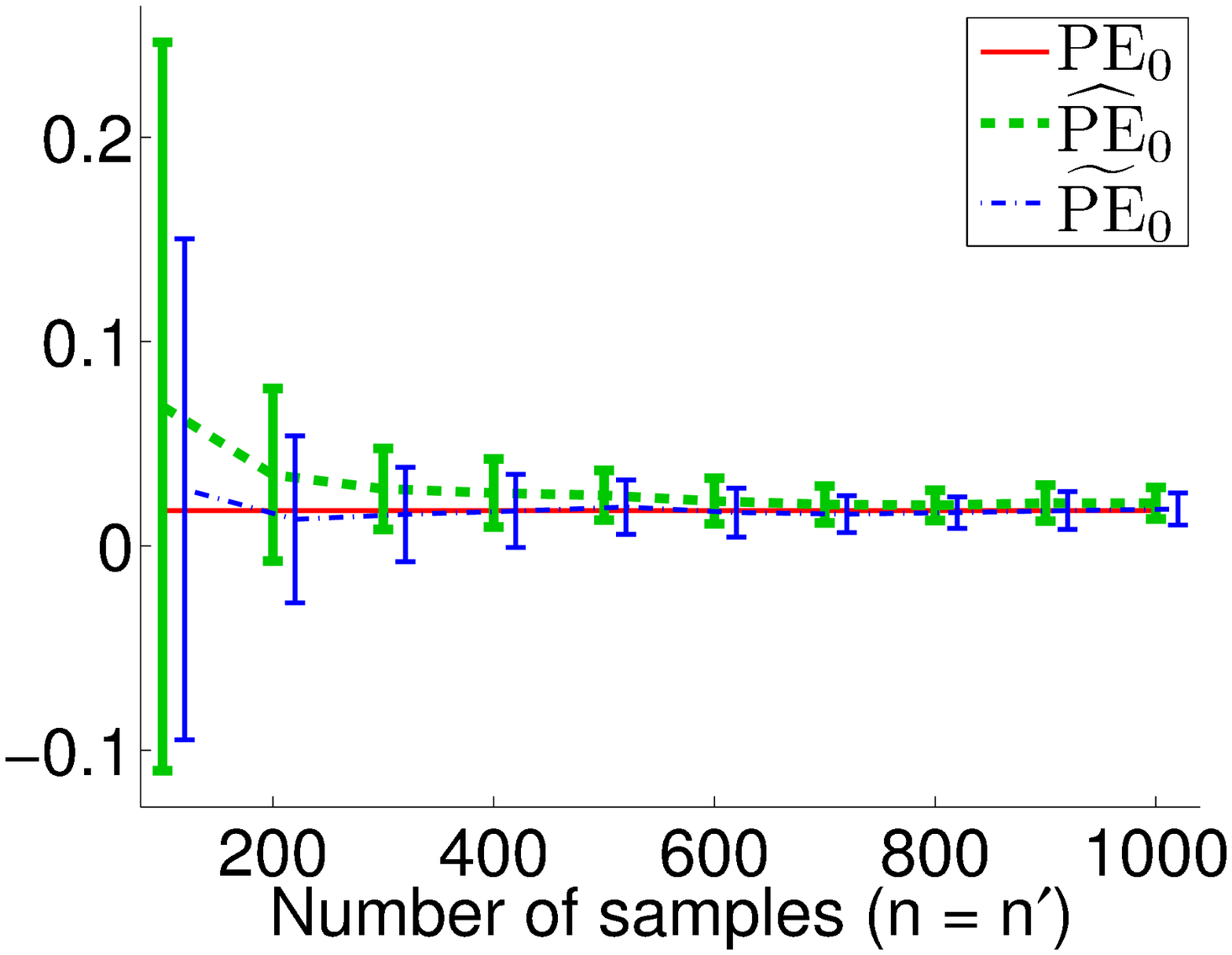}&
  \includegraphics[width=.24\textwidth]{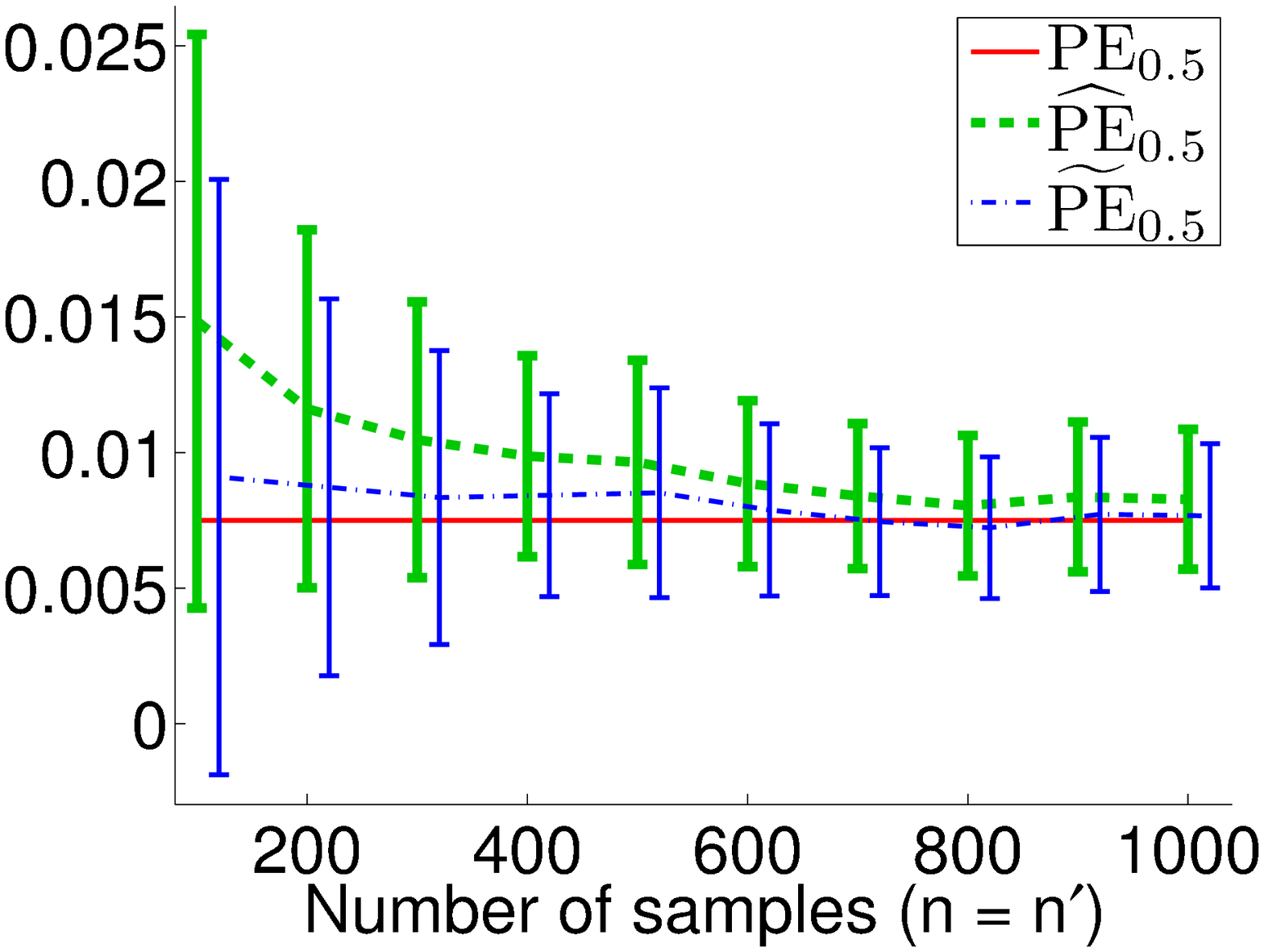}&
  \includegraphics[width=.24\textwidth]{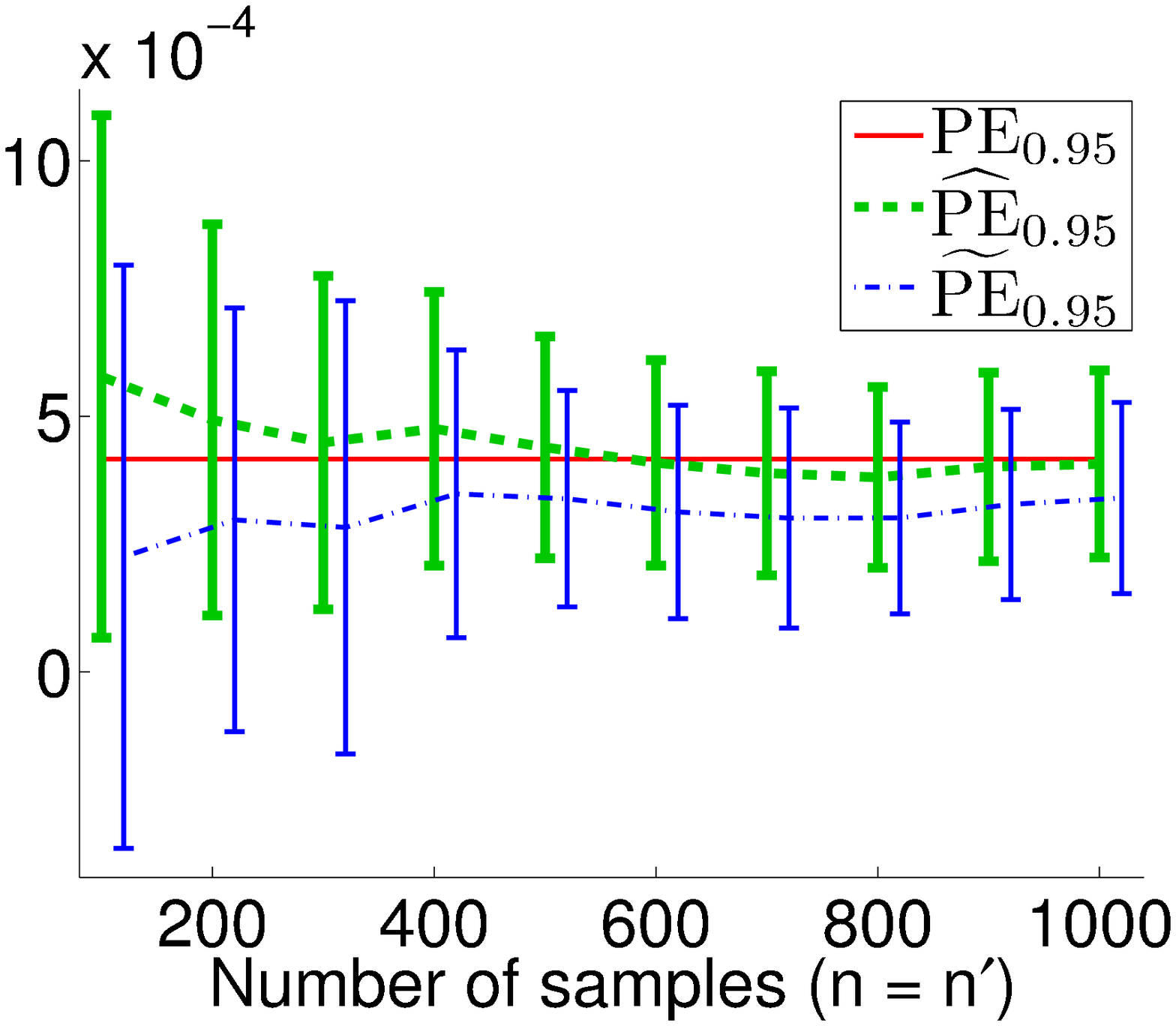}
    \end{tabular}
    \label{fig:illustrative-divergence-outlier}
      }
 \caption{Illustrative examples of divergence estimation by RuLSIF.
   From left to right: true density-ratios for $\alpha=0$, $0.5$, and $0.95$
   ($\Pnu = N(0,1)$),
   and estimation error of PE divergence for $\alpha=0$, $0.5$, and $0.95$.}
    \label{fig:illustrative-divergence}
  \end{figure}

\subsubsection{Numerical Illustration}
Let us numerically investigate the above interpretation
using the same artificial dataset as Section~\ref{subsec:illustration}.

Figure~\ref{fig:illustrative-divergence} shows
the mean and standard deviation of 
$\widehat{\mathrm{PE}}_\alpha$ and $\widetilde{\mathrm{PE}}_\alpha$
over $100$ runs for $\alpha=0$, $0.5$, and $0.95$,
as functions of $\nnu$ ($=\nde$ in this experiment).
The true $\mathrm{PE}_\alpha$ (which was numerically computed)
is also plotted in the graphs.
The graphs show that
both the estimators $\widehat{\mathrm{PE}}_\alpha$ and $\widetilde{\mathrm{PE}}_\alpha$
approach the true $\mathrm{PE}_\alpha$
as the number of samples increases,
and the approximation error tends to be smaller if $\alpha$ is larger.

When $\alpha$ is large,
$\widehat{\mathrm{PE}}_\alpha$ tends to perform
slightly better than $\widetilde{\mathrm{PE}}_\alpha$.
On the other hand, when $\alpha$ is small and the number of samples is small,
$\widetilde{\mathrm{PE}}_\alpha$ slightly compares favorably with
$\widehat{\mathrm{PE}}_\alpha$.
Overall, these numerical results well agree with our theory.

\subsection{Parametric Variance Analysis}
\label{subsec:para-analysis}
Next, we analyze the asymptotic variance of the PE divergence estimator
$\widehat{\mathrm{PE}}_\alpha$ \eqref{eq:PE1} under a parametric setup.

\subsubsection{Theoretical Results}
As the function space $\calG$ in Eq.\eqref{alpha-uLSIF-general},
we consider the following parametric model:
\begin{align*}
  \relratioModel=\{\ratiomodel(\boldx;\boldtheta)~|~\boldtheta\in\Theta\subset\Rbb^\nparam\},
\end{align*}
where $\nparam$ is a finite number.
Here we assume that the above parametric model is \emph{correctly specified},
i.e., it includes the true relative density-ratio function $\relratio(\boldx)$:
there exists $\boldtheta^*$ such that
\begin{align*}
  \ratiomodel(\boldx;\boldtheta^*)=\relratio(\boldx).
\end{align*}
Here, we use RuLSIF without regularization,
i.e., $\lambda=0$ in Eq.\eqref{alpha-uLSIF-general}.

Let us denote the variance of $\widehat{\mathrm{PE}}_\alpha$ \eqref{eq:PE1}
by $\mathbbV[\hatPEest]$,
where randomness comes from 
the draw of samples $\{\boldxnu_i\}_{i=1}^{\nnu}$ and $\{\boldxde_j\}_{j=1}^{\nde}$.
Then, under a standard regularity condition for the asymptotic normality
\cite[see Section 3 of][]{Book:VanDerVaart:AsymptoticStat},
$\mathbbV[\hatPEest]$ can be expressed and upper-bounded as
 \begin{align}
 \Vbb[\hatPEest]
  &=
 \frac{1}{\mnu}\Vnu\bigg[\relratio-\frac{\alpha{}\relratio(\boldx)^2}{2}\bigg]
 +\frac{1}{\mde}\Vde\bigg[\frac{(1-\alpha)\relratio(\boldx)^2}{2}\bigg]
 +o\bigg(\frac{1}{\mnu},\,\frac{1}{\mde}\bigg)
  \label{eqn:theorem_val-main}\\
 &\leq 
  \frac{\|\relratio\|_\infty^2}{\mnu}
 +\frac{\alpha^2\|\relratio\|_\infty^4}{4\mnu}
 +\frac{(1-\alpha)^2\|\relratio\|_\infty^4}{4\mde}
 +o\!\left(\frac{1}{\mnu},\frac{1}{\mde}\right). 
  \label{eqn:theorem_val_upper-main}
\end{align}

Let us denote the variance of $\tildePEest$ by $\Vbb[\tildePEest]$. 
Then, under a standard regularity condition for the asymptotic normality
\cite[see Section 3 of][]{Book:VanDerVaart:AsymptoticStat},
the variance of $\tildePEest$ is asymptotically expressed as
 \begin{align}
 \Vbb[\tildePEest]
  &=
  \frac{1}{\mnu}\Vnu\bigg[\frac{\relratio+(1-\alpha\relratio)
  \Enu[\nabla{g}]^\top\boldU_\alpha^{-1}\nabla{g}}{2}\bigg]\nonumber\\
&\phantom{=}
+\frac{1}{\mde}\Vde\bigg[\frac{(1-\alpha)\relratio\Enu[\nabla{g}]^\top\boldU_\alpha^{-1}\nabla{g}}{2}\bigg]
 +o\bigg(\frac{1}{\mnu},\frac{1}{\mde}\bigg),
  \label{eqn:variance_PE2-main}
\end{align} 
where $\nabla{g}$ is the gradient vector of $g$ with respect to $\boldtheta$ at
$\boldtheta=\boldtheta^*$, i.e., 
\begin{align*}
  (\nabla{g}(\boldx;\boldtheta^*))_j=\frac{\partial{g}(\boldx;\boldtheta^*)}{\partial\theta_j}.
\end{align*}
The matrix $\boldU_\alpha$ is defined by 
 \begin{align*}
  \boldU_\alpha=\alpha\Enu[\nabla{g}\nabla{g}^\top]+(1-\alpha)\Ede[\nabla{g}\nabla{g}^\top]. 
 \end{align*}

\subsubsection{Interpretation}
Eq.\eqref{eqn:theorem_val-main} shows that,
up to $O\!\left(\frac{1}{\mnu},\frac{1}{\mde}\right)$,
the variance of $\hatPEest$ depends only on
the true relative density-ratio $\relratio(\boldx)$,
not on the estimator of $\relratio(\boldx)$.
This means that the model complexity does not affect the asymptotic variance.
Therefore, \emph{overfitting} would hardly occur in the estimation of the 
relative PE divergence even when complex models are used.
We note that the above superior property is applicable only
to relative PE divergence estimation,
not to relative density-ratio estimation.
This implies that overfitting occurs in relative density-ratio estimation,
but the approximation error cancels out in relative PE divergence estimation.

On the other hand, Eq.\eqref{eqn:variance_PE2-main} shows that
the variance of $\tildePEest$ is affected by the model $\relratioModel$, since the
factor $\Enu[\nabla{g}]^\top\boldU_\alpha^{-1}\nabla{g}$ depends on the model complexity in general. 
When the equality 
\begin{align*}
  \Enu[\nabla{g}]^\top\boldU_\alpha^{-1}\nabla{g}(\boldx;\boldtheta^*)=\relratio(\boldx)
\end{align*}
holds, the variances of $\tildePEest$ and $\hatPEest$ are asymptotically the same. 
However, in general, the use of $\hatPEest$ would be more recommended.

Eq.\eqref{eqn:theorem_val_upper-main} shows that
the variance $\Vbb[\hatPEest]$ can be upper-bounded by the quantity
depending on $\|\relratio\|_\infty$, which is monotonically lowered
if $\|\relratio\|_\infty$ is reduced.
Since $\|\relratio\|_\infty$ monotonically decreases as $\alpha$ increases,
our proposed approach of estimating the $\alpha$-relative PE divergence
(with $\alpha>0$) would be more advantageous than the naive approach of
estimating the plain PE divergence (which corresponds to $\alpha=0$)
in terms of the parametric asymptotic variance.

\begin{figure}[p]
\centering
\footnotesize
\begin{tabular}{@{}c@{}c@{}}
\includegraphics[width=.37\textwidth]{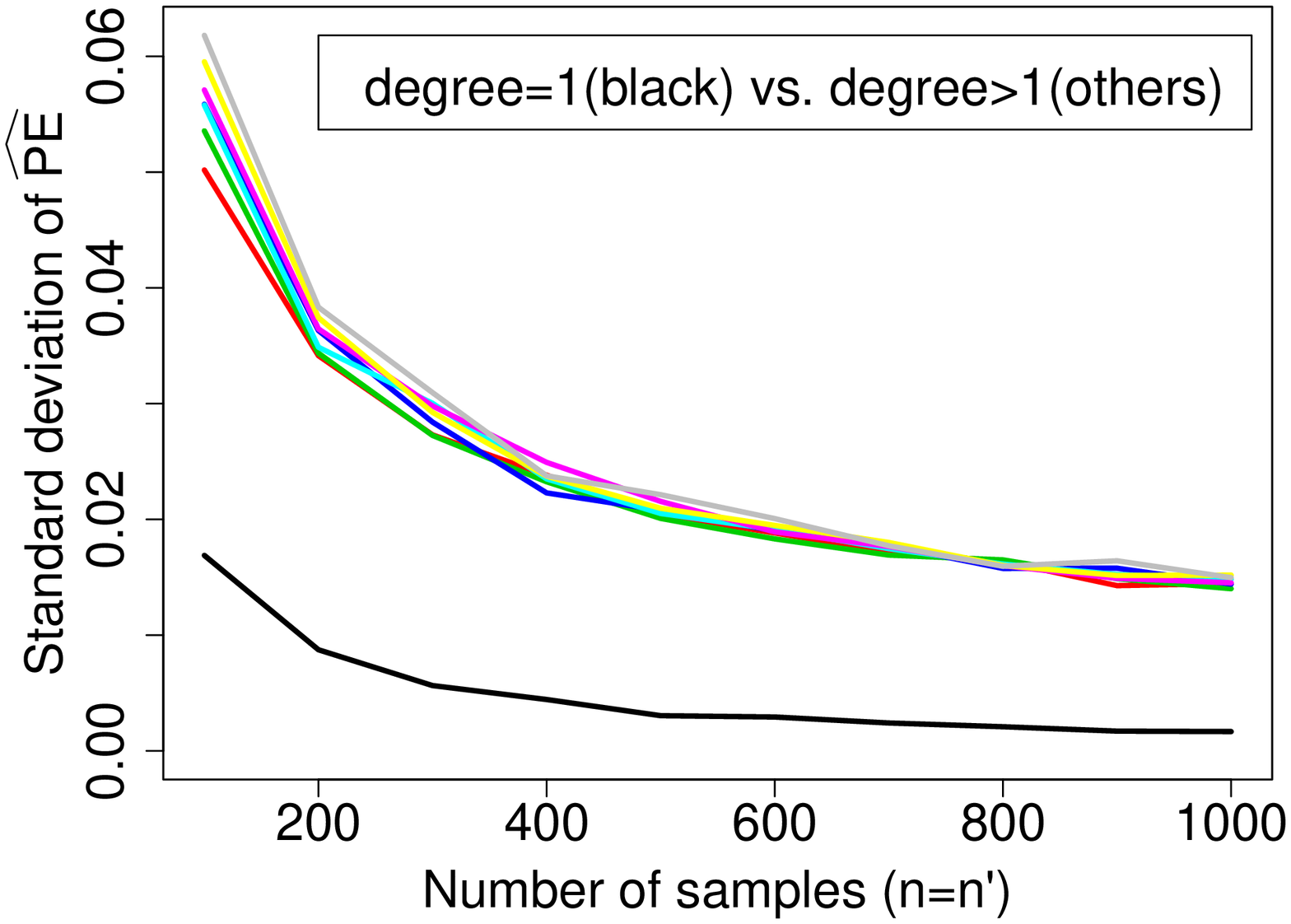}&
\includegraphics[width=.37\textwidth]{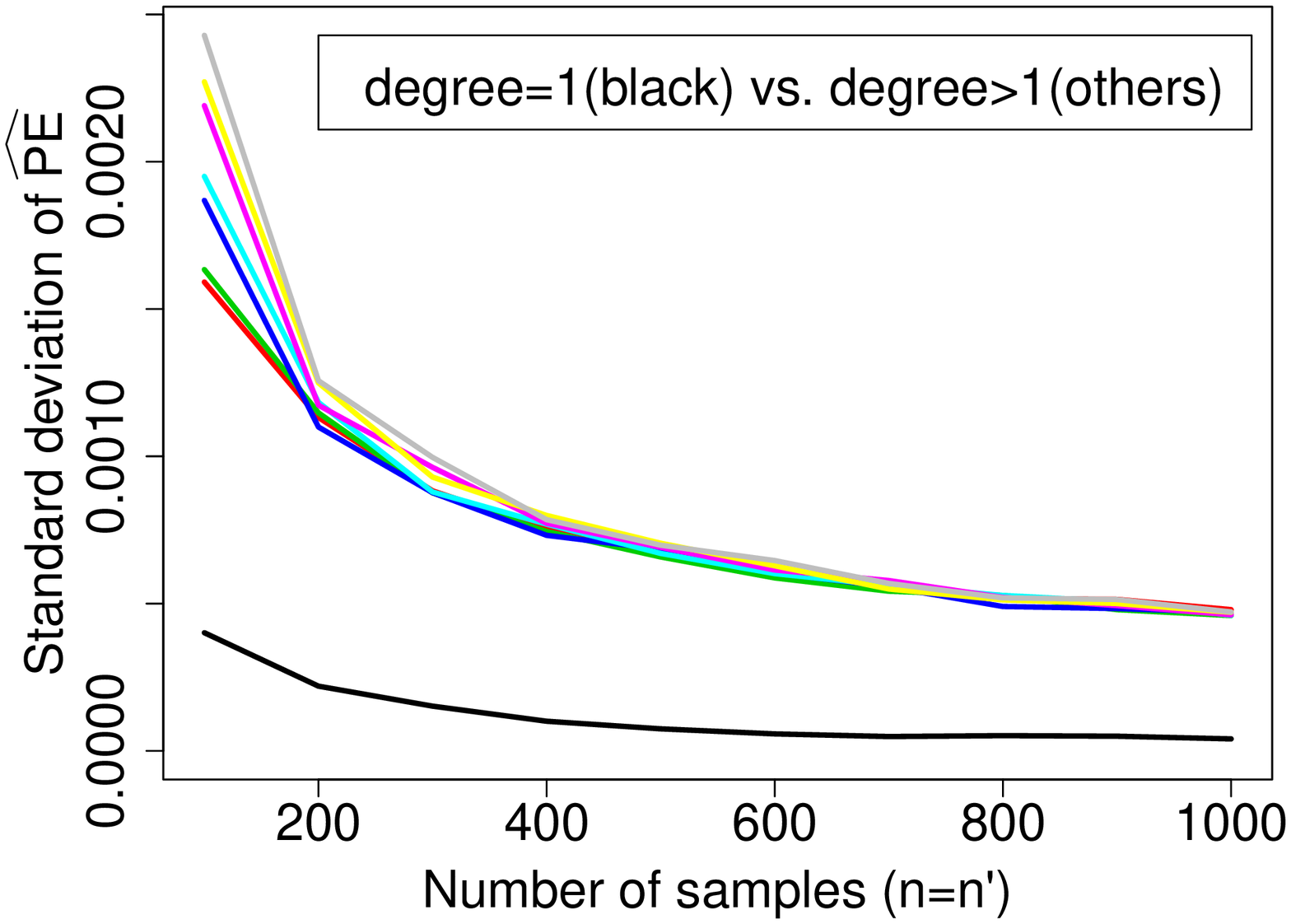}\\[-1mm]
  $\hatPEest$ with $\alpha=0.2$&
  $\hatPEest$ with $\alpha=0.8$\\
\includegraphics[width=.37\textwidth]{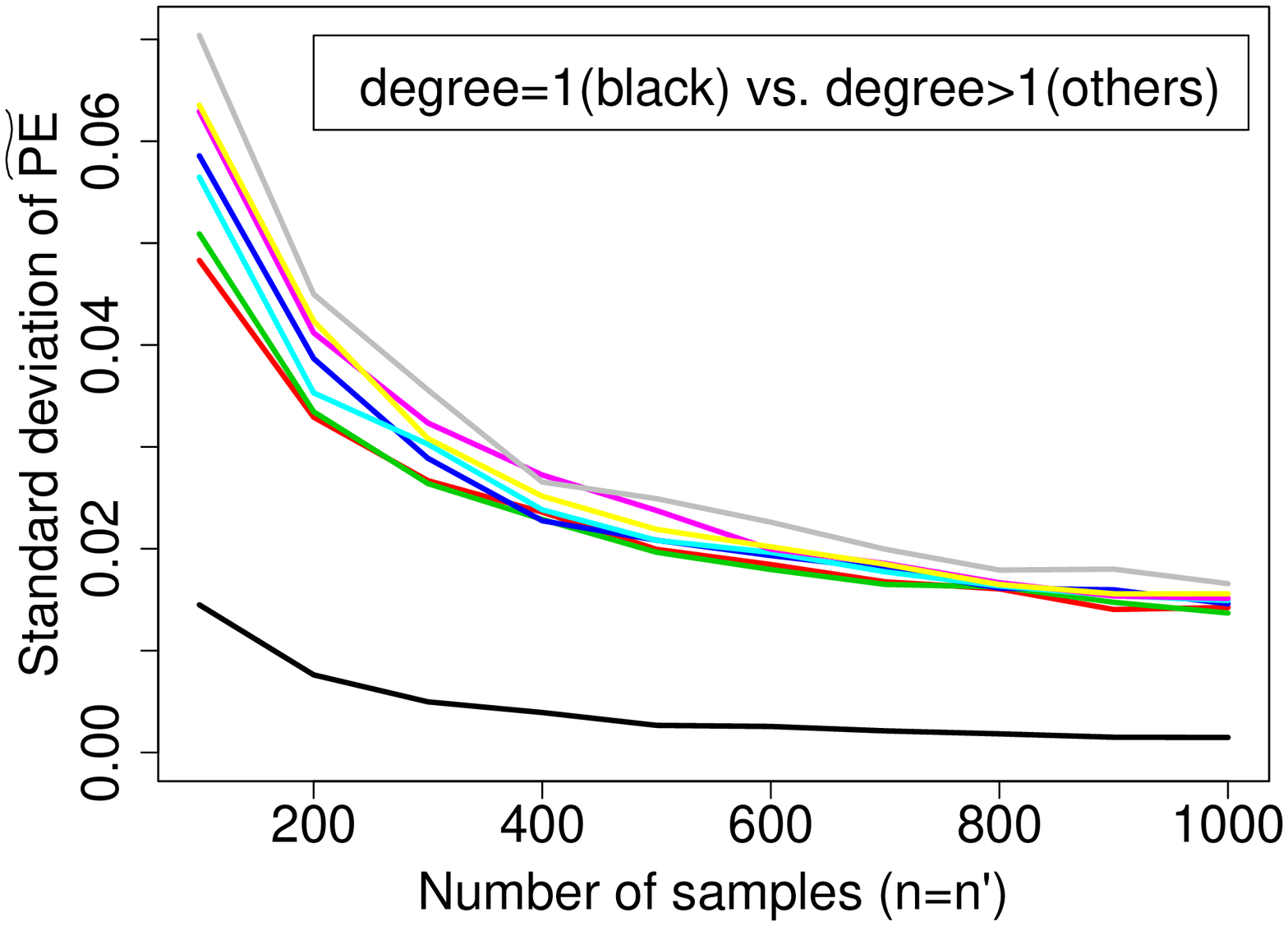}&
\includegraphics[width=.37\textwidth]{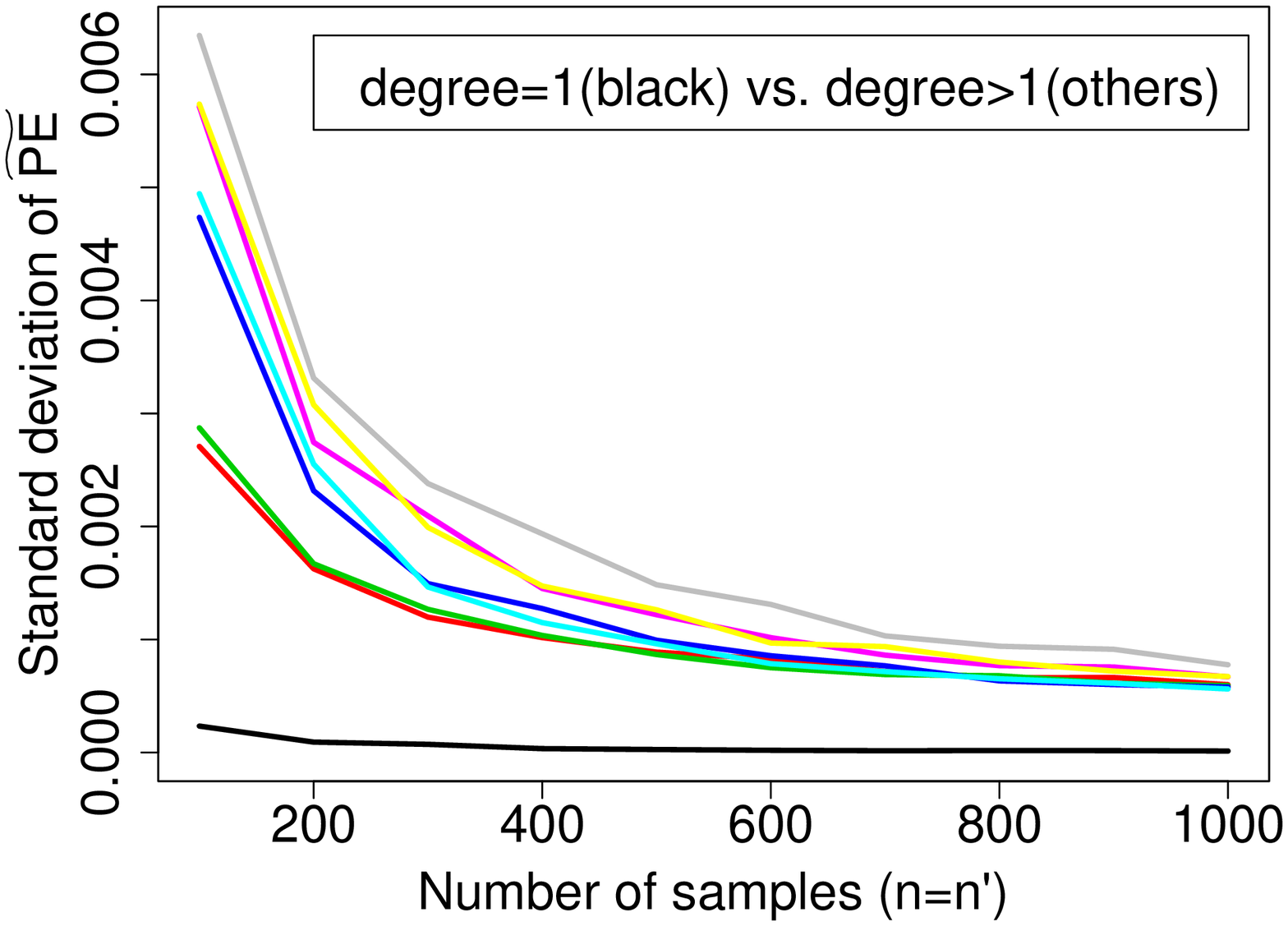}\\[-1mm]
  $\tildePEest$ with $\alpha=0.2$&
  $\tildePEest$ with $\alpha=0.8$\\
\end{tabular}
 \caption{Standard deviations of PE estimators for
   dataset (b) (i.e., $\Pnu = N(0,1)$ and $\Pde=N(0,0.6)$)
   as functions of the sample size $\mnu=\mde$.
 }
 \label{fig:Rplot-sim-exp2}
\begin{tabular}{@{}c@{}c@{}}
\includegraphics[width=.37\textwidth]{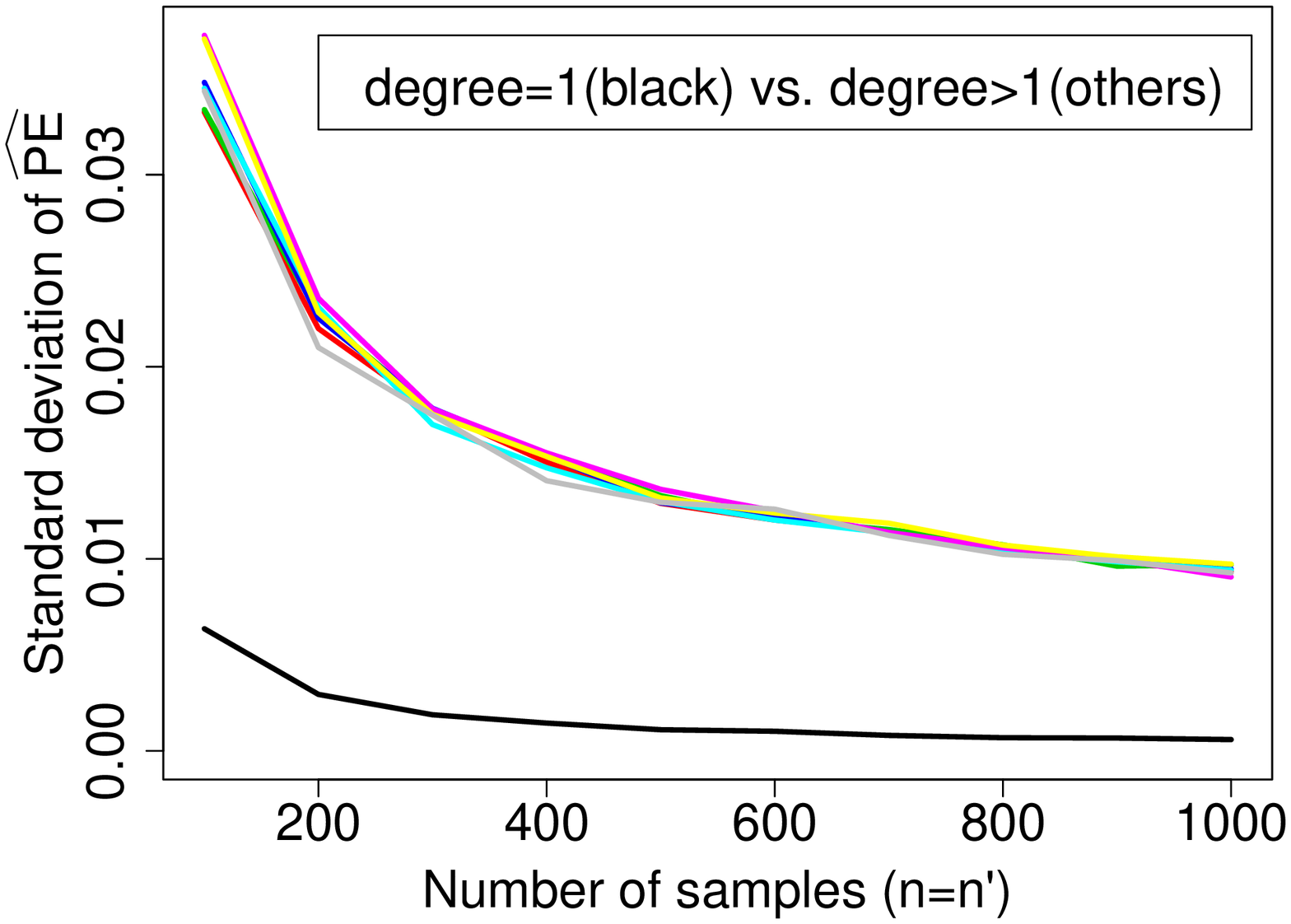}&
\includegraphics[width=.37\textwidth]{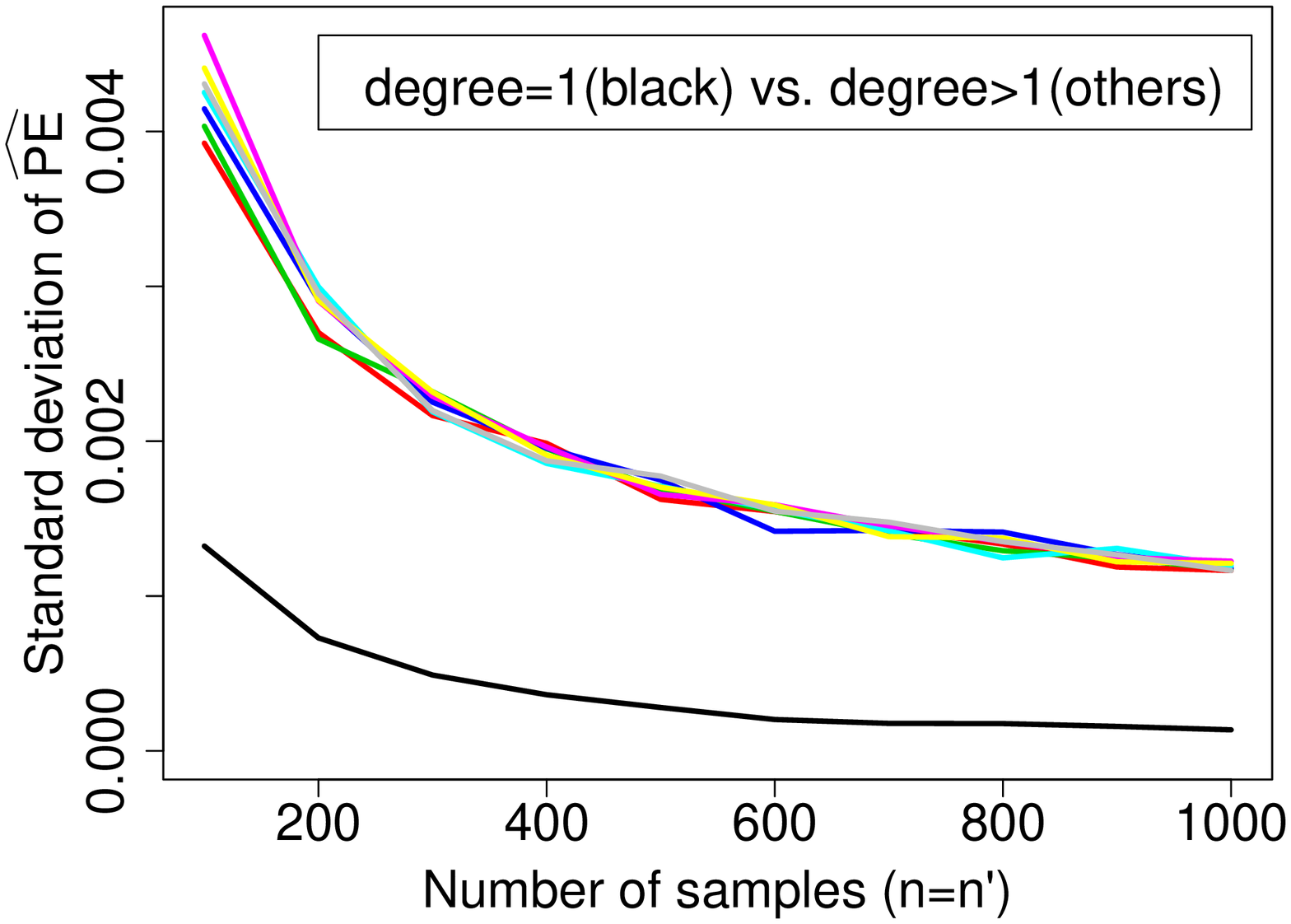}\\[-1mm]
  $\hatPEest$ with $\alpha=0.2$&
  $\hatPEest$ with $\alpha=0.8$\\
\includegraphics[width=.37\textwidth]{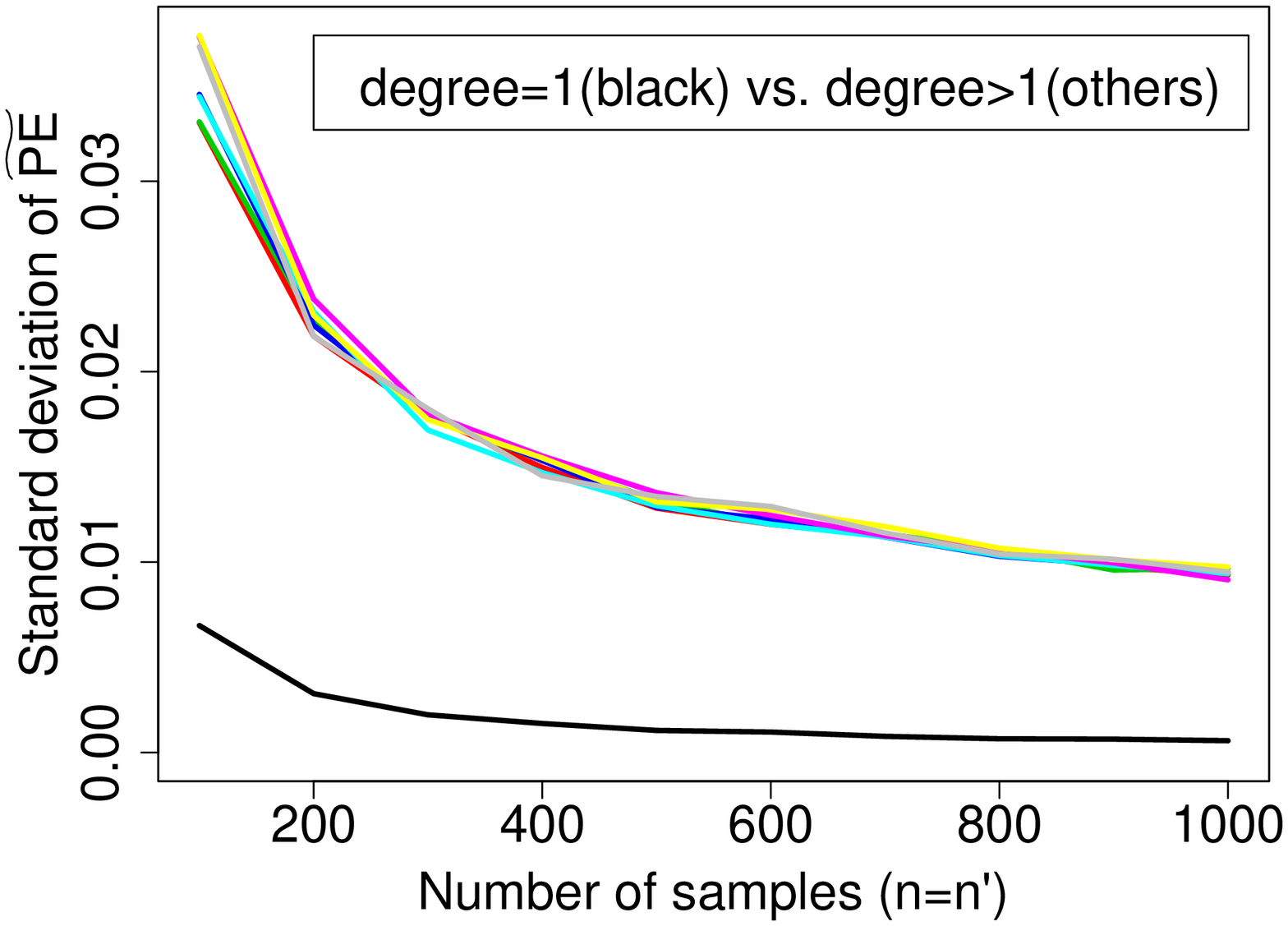}&
\includegraphics[width=.37\textwidth]{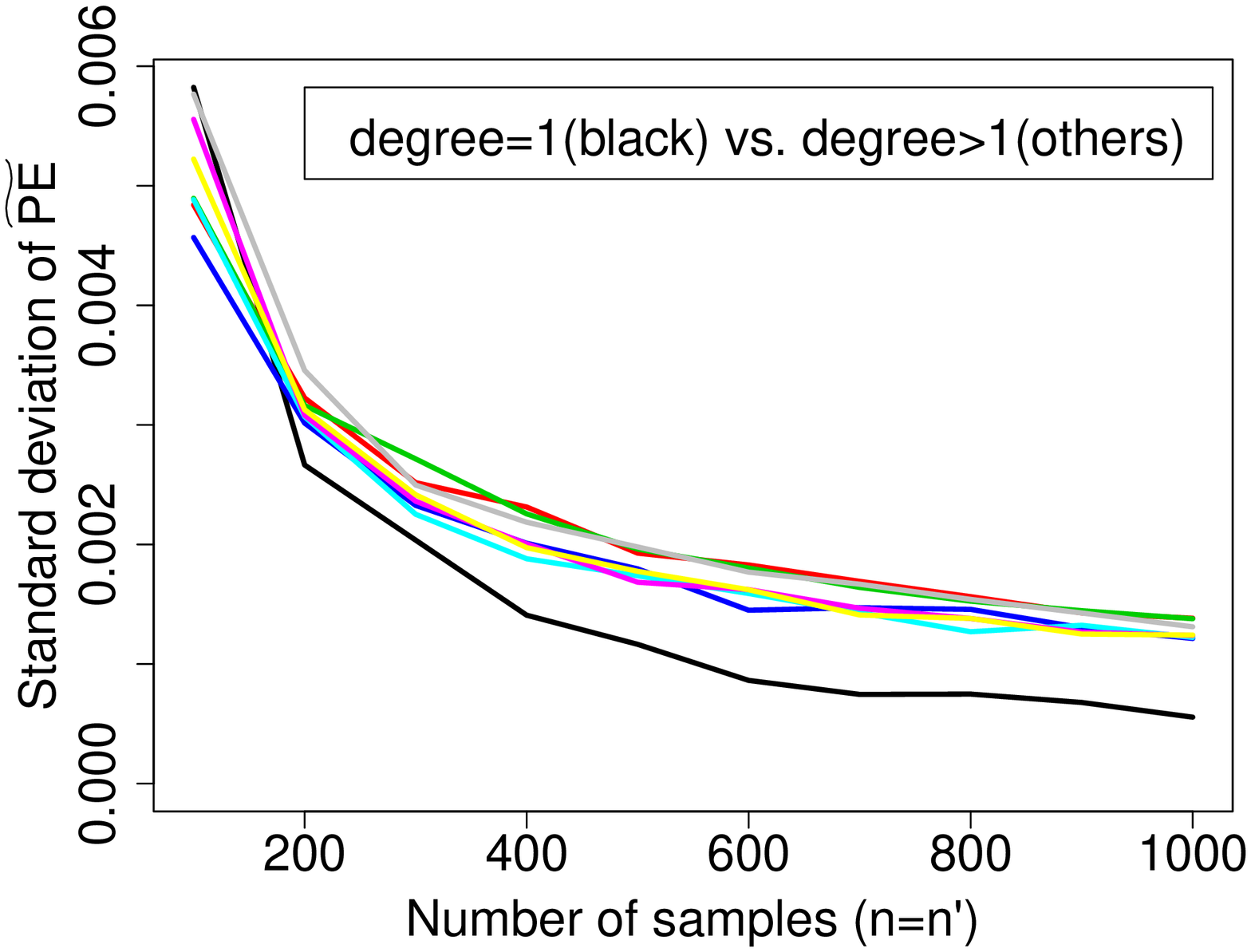}\\[-1mm]
  $\tildePEest$ with $\alpha=0.2$&
  $\tildePEest$ with $\alpha=0.8$\\
\end{tabular}
 \caption{Standard deviations of PE estimators for
   dataset (c) (i.e., $\Pnu = N(0,1)$ and $\Pde=N(0,2)$)
   as functions of the sample size $\mnu=\mde$.
 }
 \label{fig:Rplot-sim-exp3}
\end{figure}

\subsubsection{Numerical Illustration}
Here, we show some numerical results for illustrating the above theoretical results
using the one-dimensional datasets (b) and (c) in Section~\ref{subsec:illustration}.
Let us define the parametric model as
\begin{align}
  \relratioModel_k=
  \left\{
    g(x;\boldtheta)=\frac{\ratio(x;\boldtheta)}{\alpha\ratio(x;\boldtheta)+1-\alpha}~\bigg|~
    \ratio(x;\boldtheta)=\exp\Bigg(\sum_{\ell=0}^k\theta_\ell x^\ell\Bigg),\,
    \boldtheta\in\Rbb^{k+1}
  \right\}. 
  \label{para-simulation-model}
 \end{align}
The dimension of the model $\relratioModel_k$ is equal to $k+1$.
The $\alpha$-relative density-ratio $\relratio(x)$
can be expressed using the ordinary density-ratio $\ratio(x)=\pnu(x)/\pde(x)$ as
\begin{align*}
 \relratio(x)=\frac{\ratio(x)}{\alpha \ratio(x)+1-\alpha}. 
\end{align*}
Thus, when $k>1$, the above model $\relratioModel_k$
includes the true relative density-ratio $\relratio(x)$
of the datasets (b) and (c).
We test RuLSIF with $\alpha=0.2$ and $0.8$
for the model \eqref{para-simulation-model} with degree $k=1,2,\ldots,8$. 
The parameter $\boldtheta$ is learned so that
Eq.\eqref{alpha-uLSIF-general} is minimized by a quasi-Newton method.

The standard deviations of $\hatPEest$ and $\tildePEest$
for the datasets (b) and (c)
are depicted in Figure~\ref{fig:Rplot-sim-exp2}
and Figure~\ref{fig:Rplot-sim-exp3}, respectively.
The graphs show that the degree of models does not
significantly affect the standard deviation of $\hatPEest$
(i.e., no overfitting),
as long as the model includes the true relative density-ratio (i.e., $k>1$). 
On the other hand, 
bigger models tend to produce larger standard deviations in $\tildePEest$. 
Thus, the standard deviation of $\tildePEest$
more strongly depends on the model complexity.

\section{Experiments}
\label{sec:experiments}
In this section, we experimentally evaluate the performance of
the proposed method
in two-sample homogeneity test, outlier detection,
and transfer learning tasks.

\subsection{Two-Sample Homogeneity Test}
First, we apply the proposed divergence estimator to two-sample homogeneity test.

\subsubsection{Divergence-Based Two-Sample Homogeneity Test}
Given two sets of samples $\calXnu=\{\boldxnu_i\}_{i=1}^{\nnu}\iid\Pnu$
and $\calXde=\{\boldxde_j\}_{j=1}^{\nde}\iid\Pde$, 
the goal of the two-sample homogeneity test is to test the \emph{null hypothesis}
that the probability distributions $\Pnu$ and $\Pde$ are the same
against its complementary alternative (i.e., the distributions are different).

By using an estimator $\widehat{\mathrm{Div}}$ of some divergence between the two distributions
$\Pnu$ and $\Pde$,
homogeneity of two distributions can be tested
based on the \emph{permutation test} procedure \citep{book:Efron+Tibshirani:1993}
as follows:
\begin{itemize}
\item Obtain a divergence estimate $\widehat{\mathrm{Div}}$
  using the original datasets $\calXnu$ and $\calXde$.
\item Randomly permute the $|\calXnu\cup\calXde|$ samples,
  and assign the first $|\calXnu|$ samples to a set $\calXnut$
  and the remaining $|\calXde|$ samples to another set $\calXdet$.
\item Obtain a divergence estimate $\widetilde{\mathrm{Div}}$
  using the randomly shuffled datasets $\calXnut$ and $\calXdet$
  (note that, since $\calXnut$ and $\calXdet$
  can be regarded as being drawn from the same distribution,
  $\widetilde{\mathrm{Div}}$ tends to be close to zero).
\item Repeat this random shuffling procedure many times,
  and construct the empirical distribution of $\widetilde{\mathrm{Div}}$
  under the null hypothesis that the two distributions are the same.
\item Approximate the p-value by evaluating the relative ranking of
  the original $\widehat{\mathrm{Div}}$
  in the distribution of $\widetilde{\mathrm{Div}}$.
\end{itemize}

When an asymmetric divergence such as the KL divergence
 \citep{Annals-Math-Stat:Kullback+Leibler:1951}
or the PE divergence \citep{PhMag:Pearson:1900} is adopted
for two-sample homogeneity test,
the test results depend on the choice of \emph{directions}:
a divergence from $\Pnu$ to $\Pde$ or from $\Pde$ to $\Pnu$.
\citet{NN:Sugiyama+etal:2011b} proposed 
to choose the direction that gives a smaller p-value---it
was experimentally shown that,
when the uLSIF-based PE divergence estimator is used
for the two-sample homogeneity test
(which is called the \emph{least-squares two-sample homogeneity test}; LSTT),
the heuristic of choosing the direction with a smaller p-value
contributes to reducing the \emph{type-II error}
(the probability of accepting incorrect null-hypotheses,
i.e., two distributions are judged to be the same
when they are actually different),
while the increase of the \emph{type-I error} 
(the probability of rejecting correct null-hypotheses,
i.e., two distributions are judged to be different when they are actually the same)
is kept moderate.

Below, we refer to LSTT with $\pnu(\boldx)/\pde(\boldx)$ as the \emph{plain LSTT},
LSTT with $\pde(\boldx)/\pnu(\boldx)$ as the \emph{reciprocal LSTT},
and LSTT with heuristically choosing the one with a smaller p-value
as the \emph{adaptive LSTT}.

\begin{figure}[p]
  \centering
  \subfigure[$\Pde=N(0,1)$: $\Pnu$ and $\Pde$ are the same.]{
    \begin{tabular}{@{}c@{}c@{}c@{}c@{}}
      \includegraphics[width=.24\textwidth]{dataset1-density.eps}&
      \includegraphics[width=.24\textwidth]{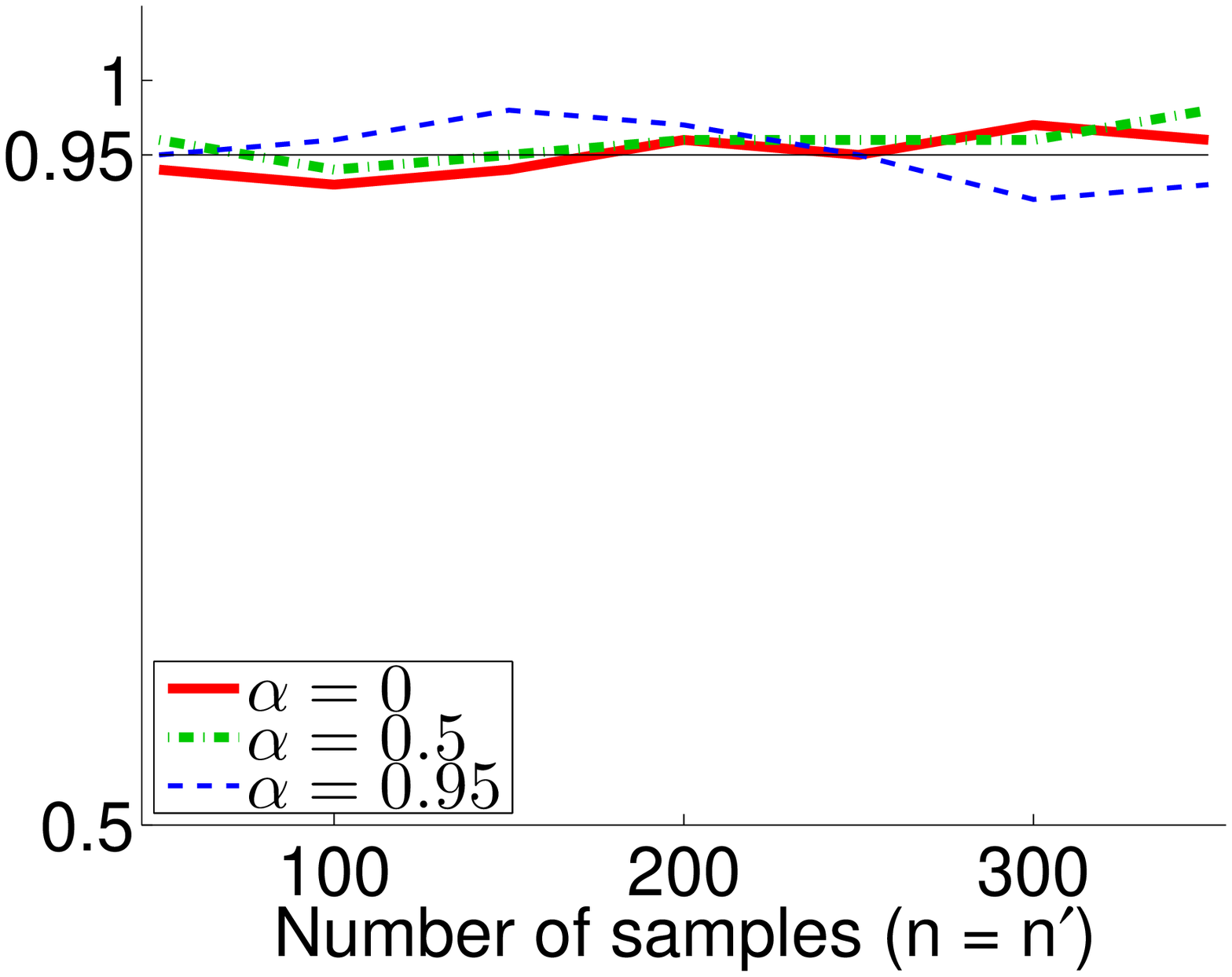}&
      \includegraphics[width=.24\textwidth]{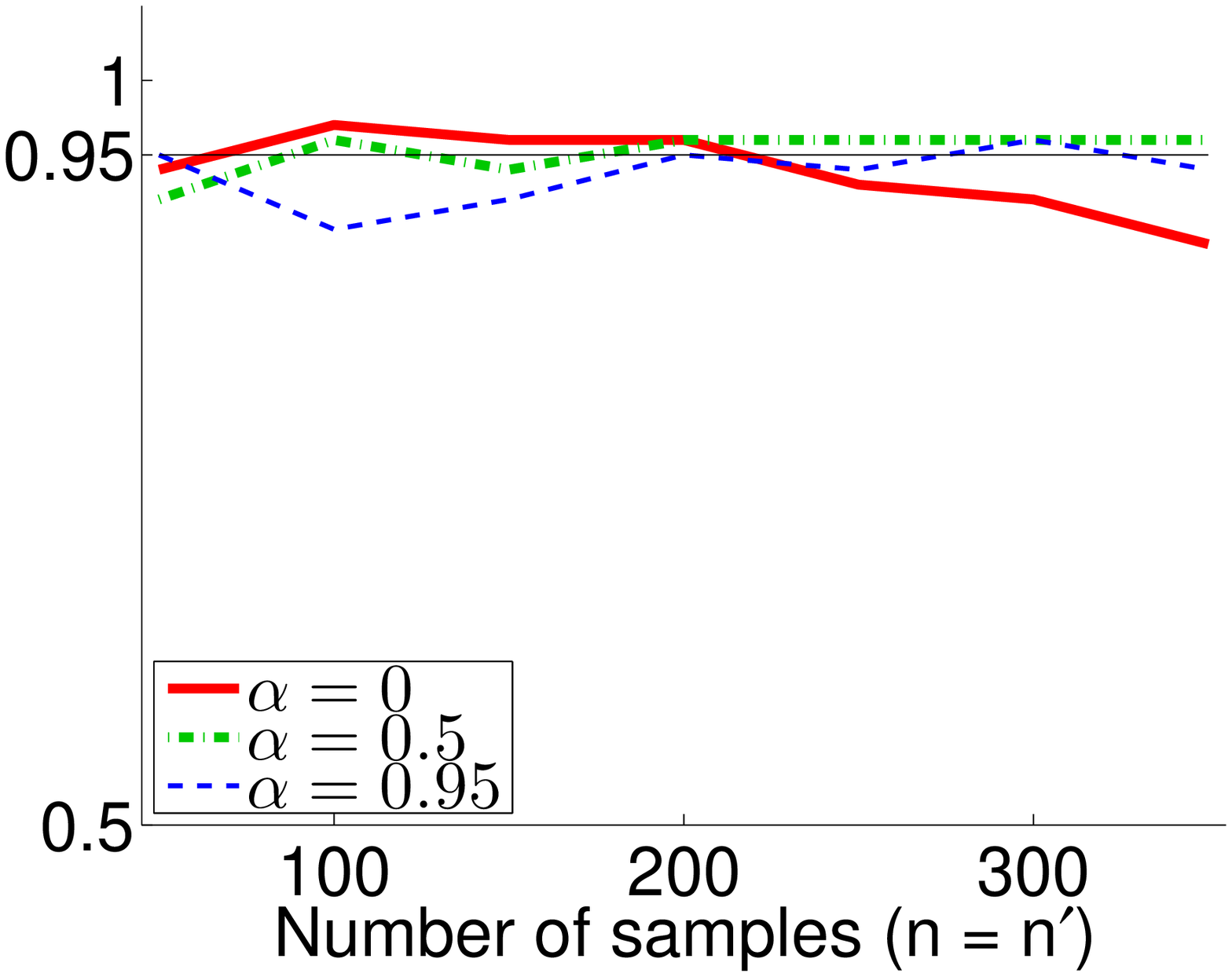}&
      \includegraphics[width=.24\textwidth]{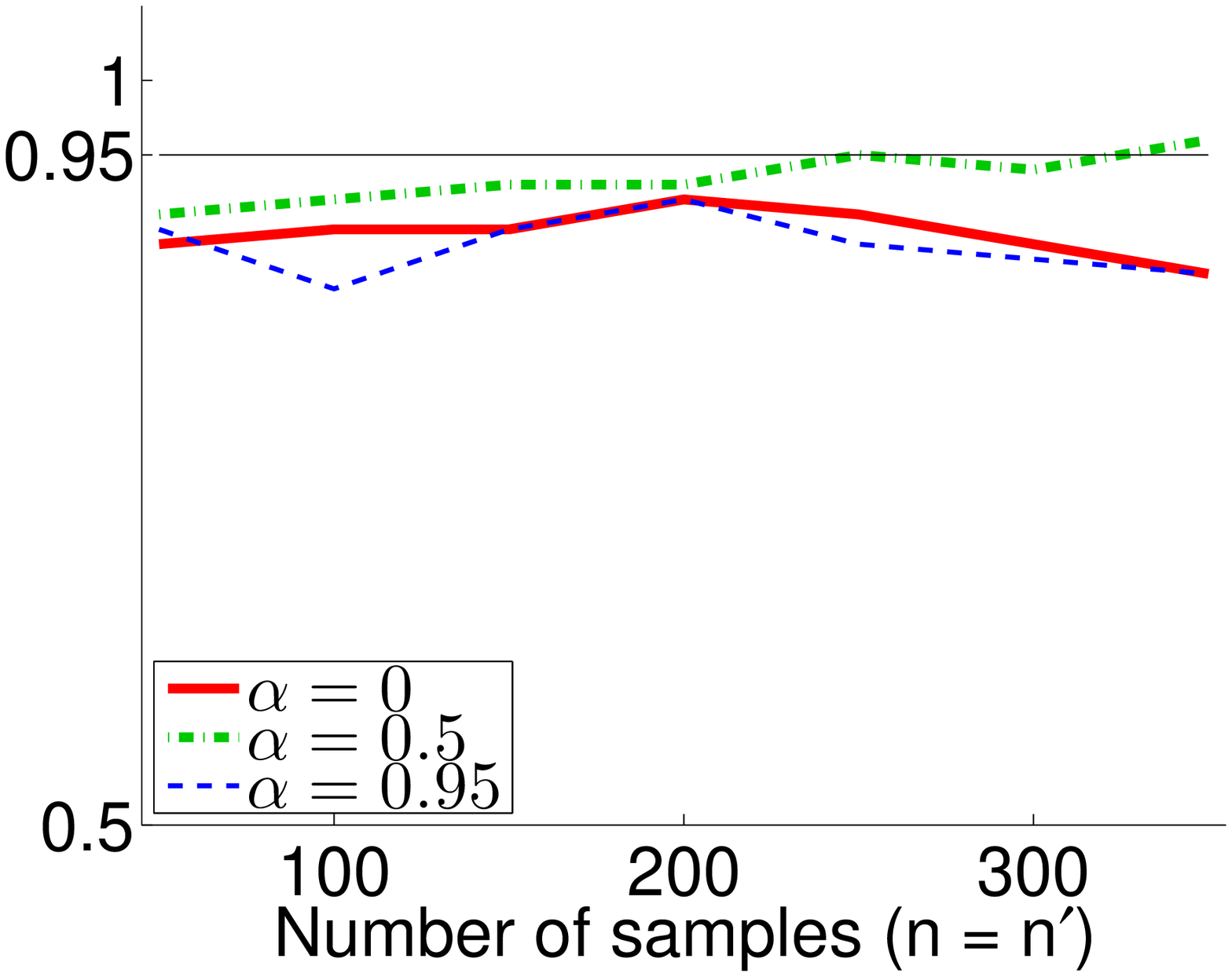}
    \end{tabular}
  }
  \subfigure[$\Pde=N(0,0.6)$: $\Pde$ has smaller standard deviation than $\Pnu$.]{
    \begin{tabular}{@{}c@{}c@{}c@{}c@{}}
      \includegraphics[width=.24\textwidth]{dataset2-density.eps}&
      \includegraphics[width=.24\textwidth]{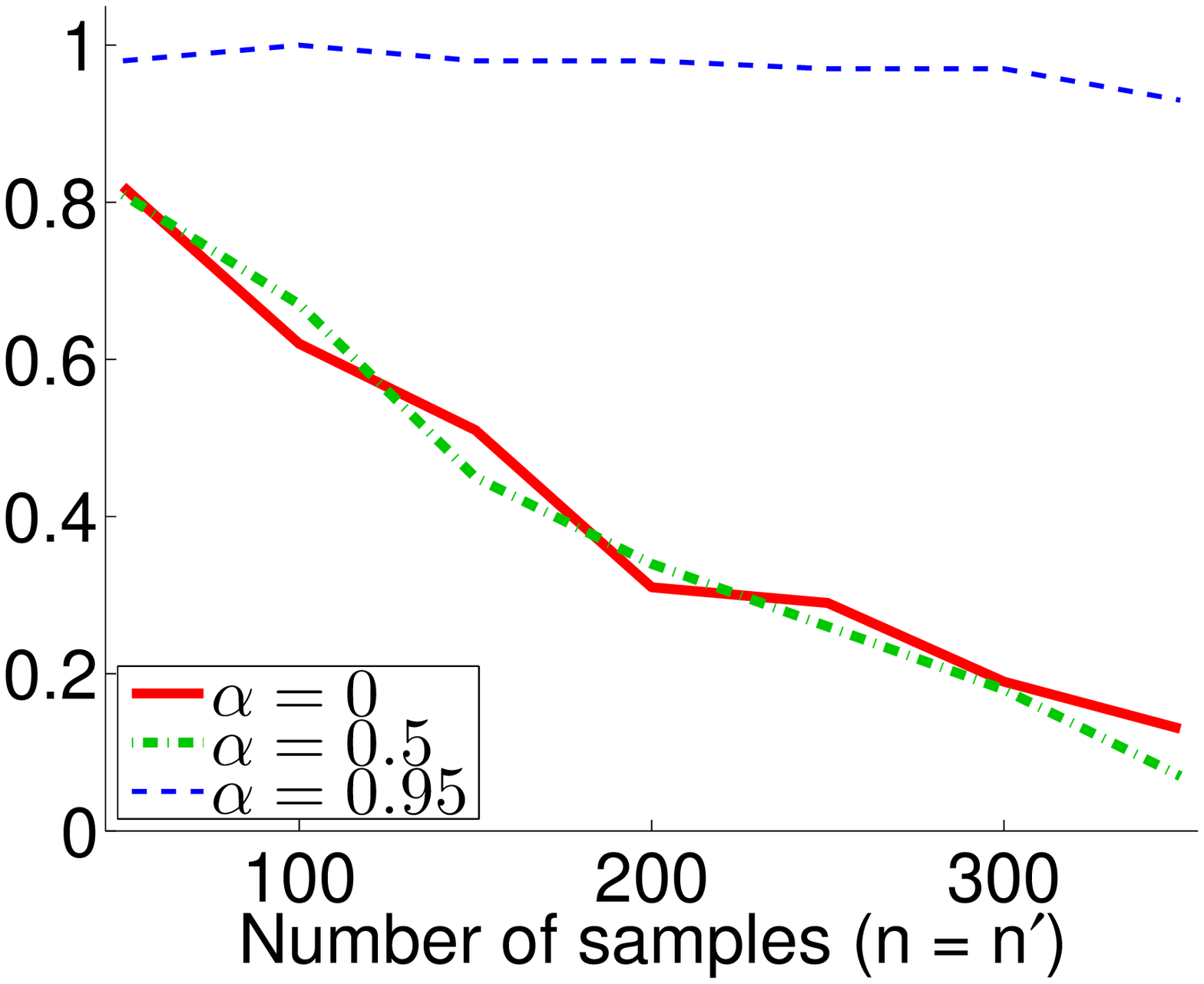}&
      \includegraphics[width=.24\textwidth]{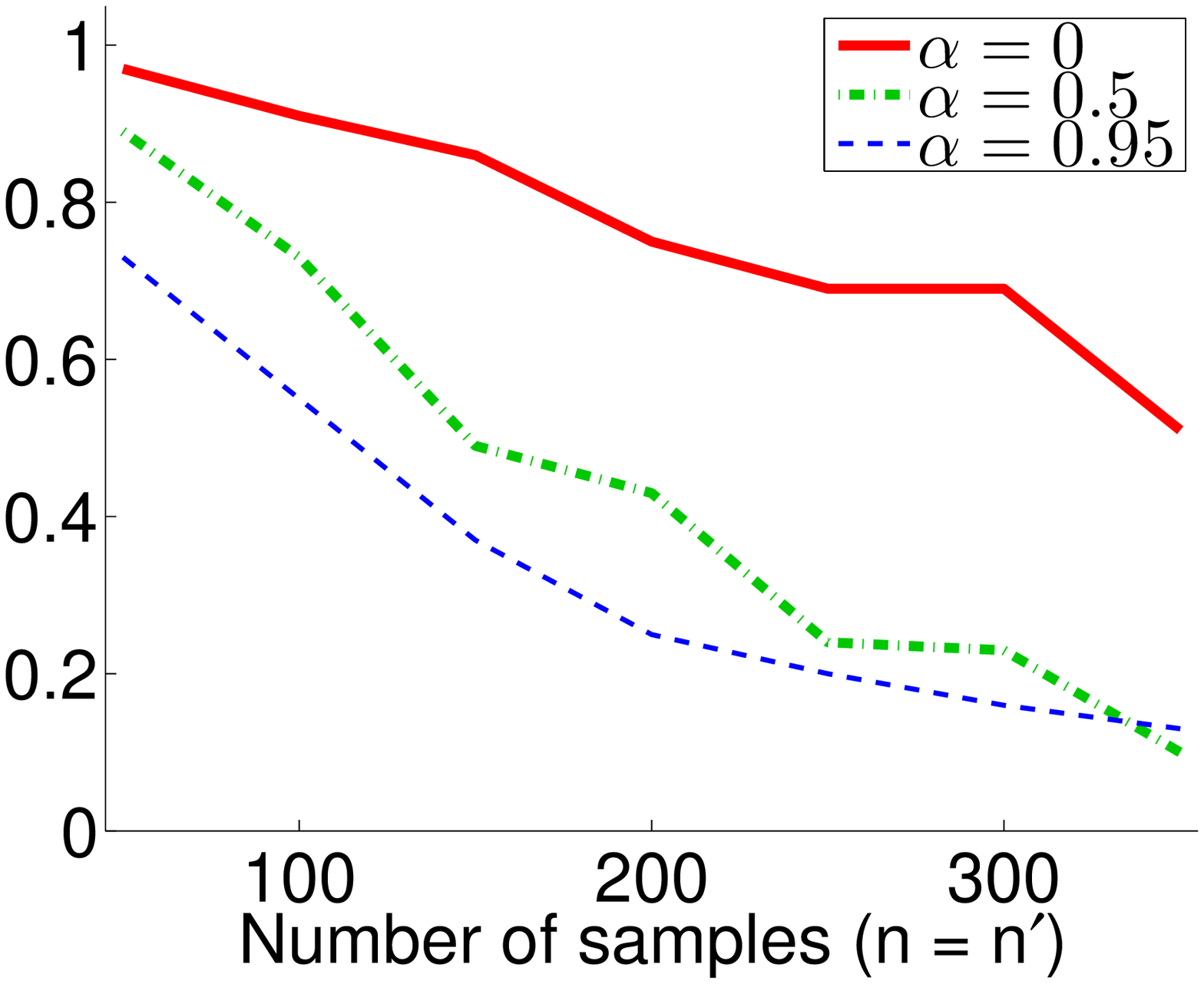}&
      \includegraphics[width=.24\textwidth]{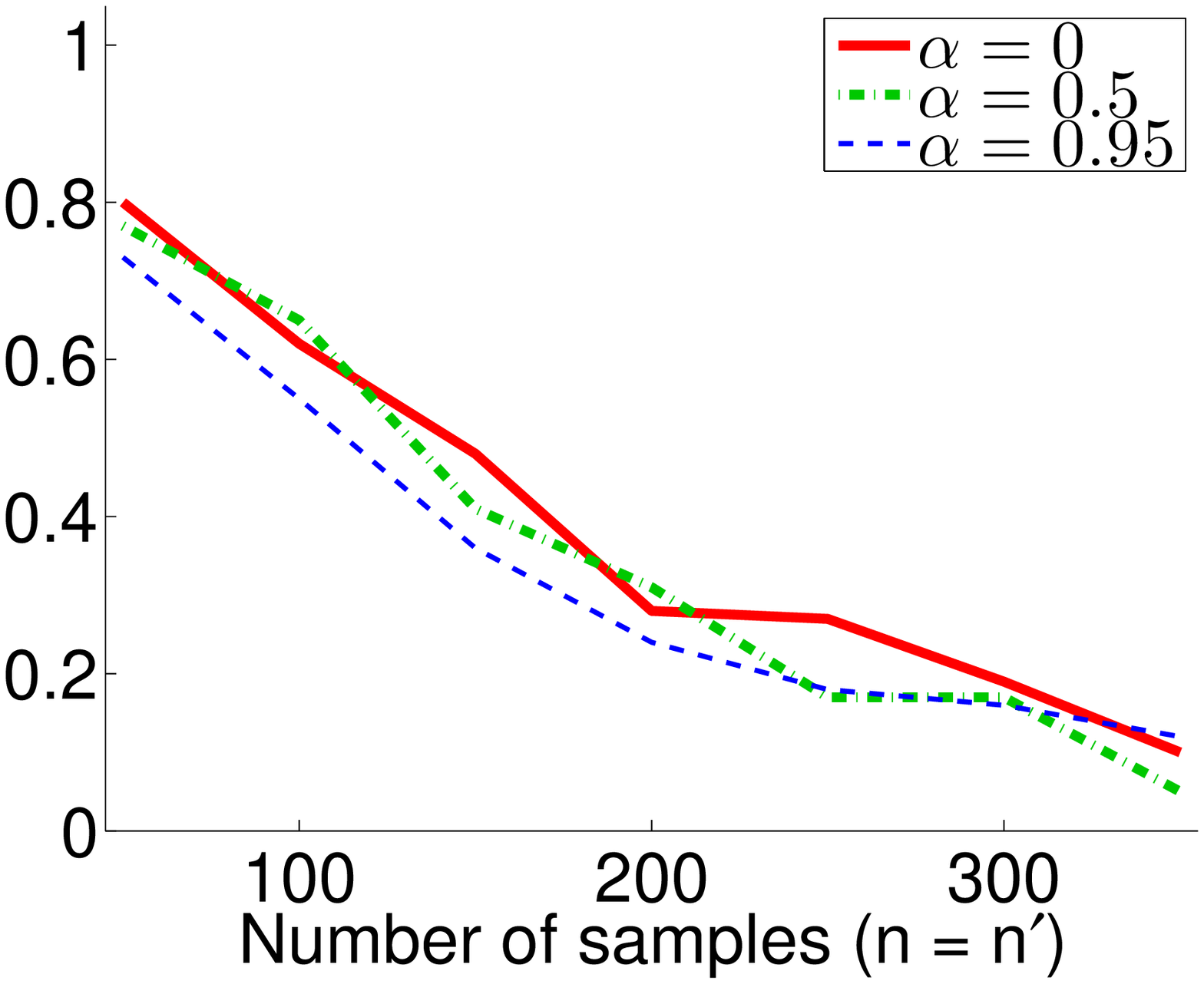}
    \end{tabular}
  }
  \subfigure[$\Pde=N(0,2)$: $\Pde$ has larger standard deviation than $\Pnu$.]{
    \begin{tabular}{@{}c@{}c@{}c@{}c@{}}
      \includegraphics[width=.24\textwidth]{dataset3-density.eps}&
      \includegraphics[width=.24\textwidth]{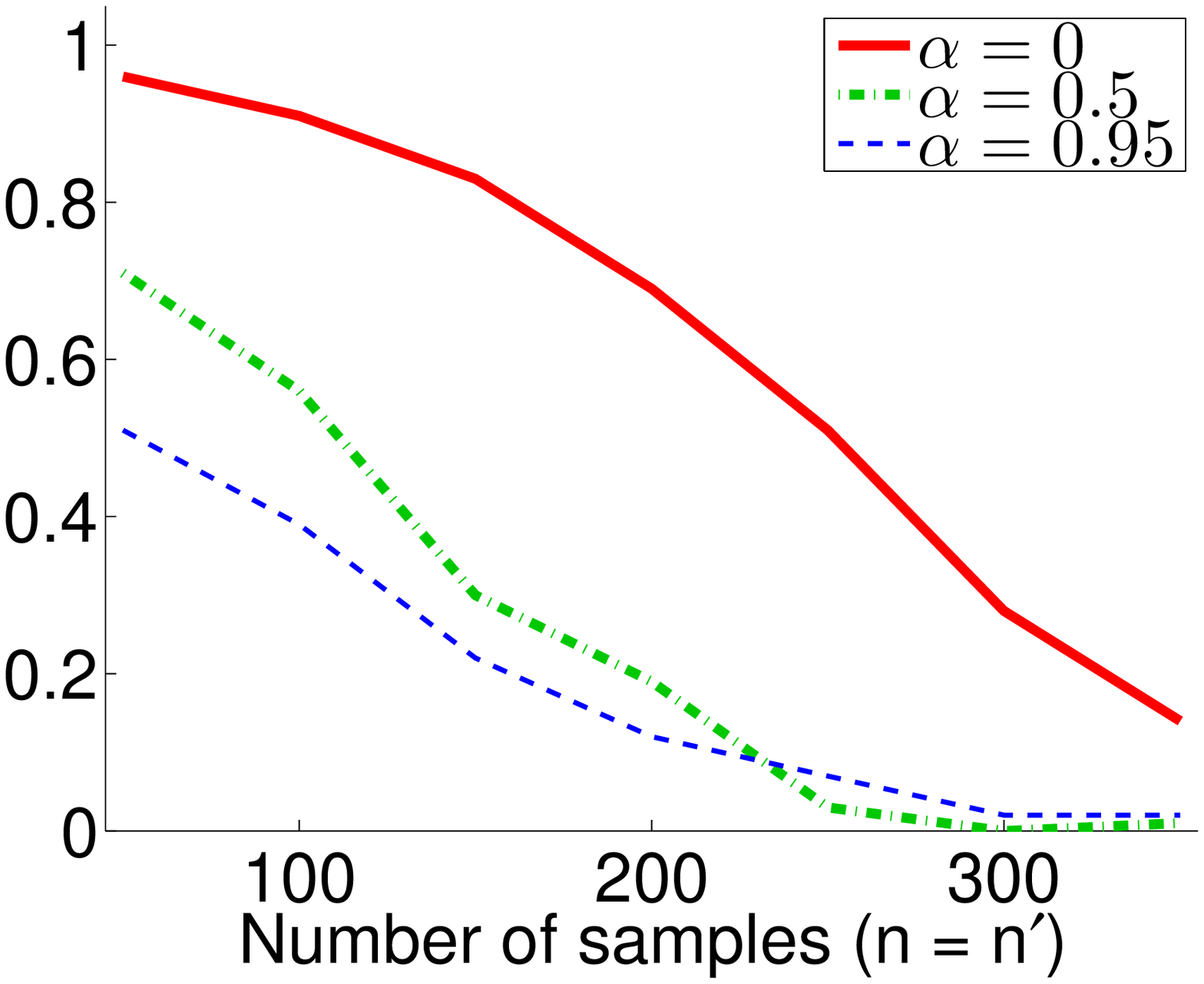}&
      \includegraphics[width=.24\textwidth]{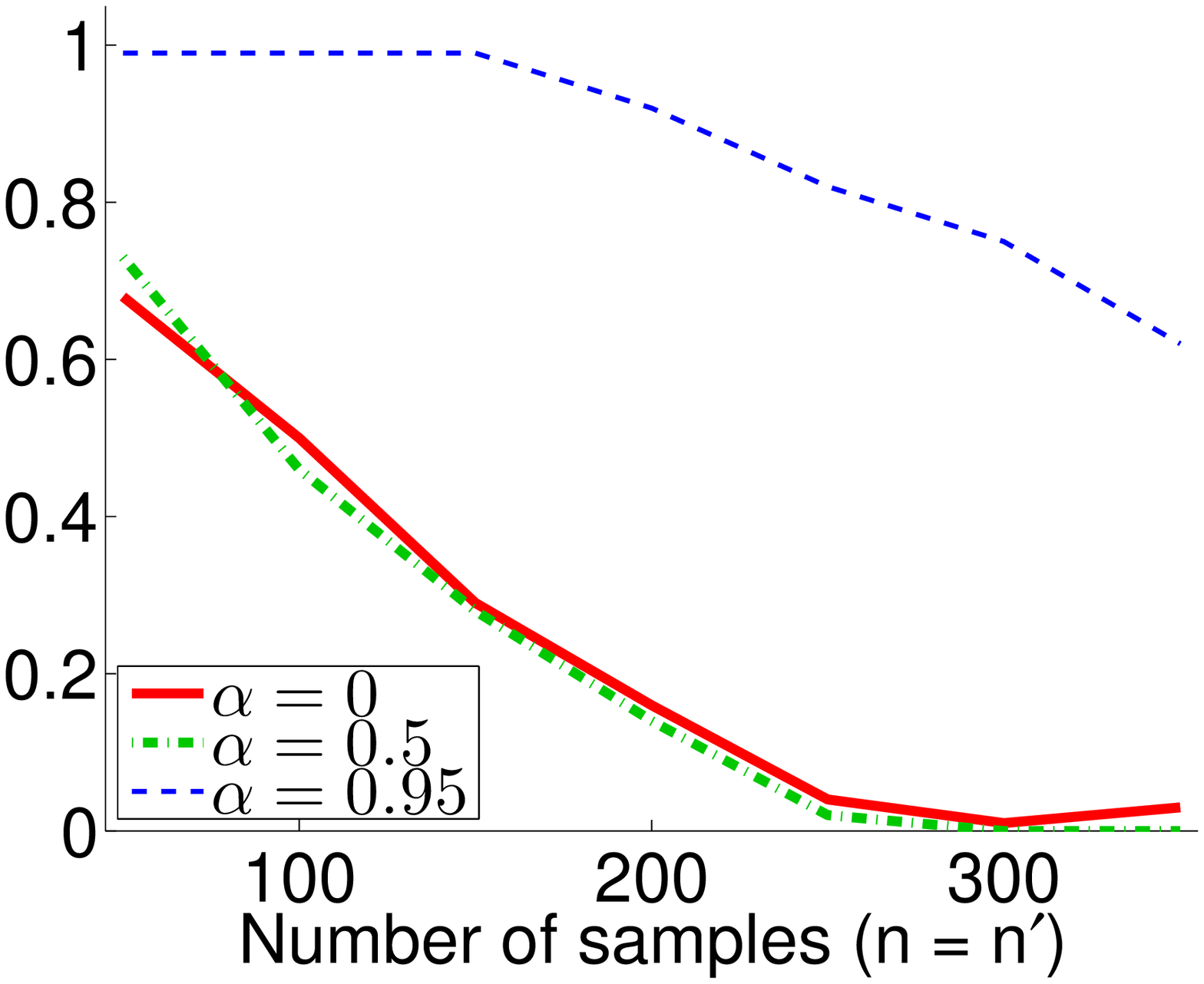}&
      \includegraphics[width=.24\textwidth]{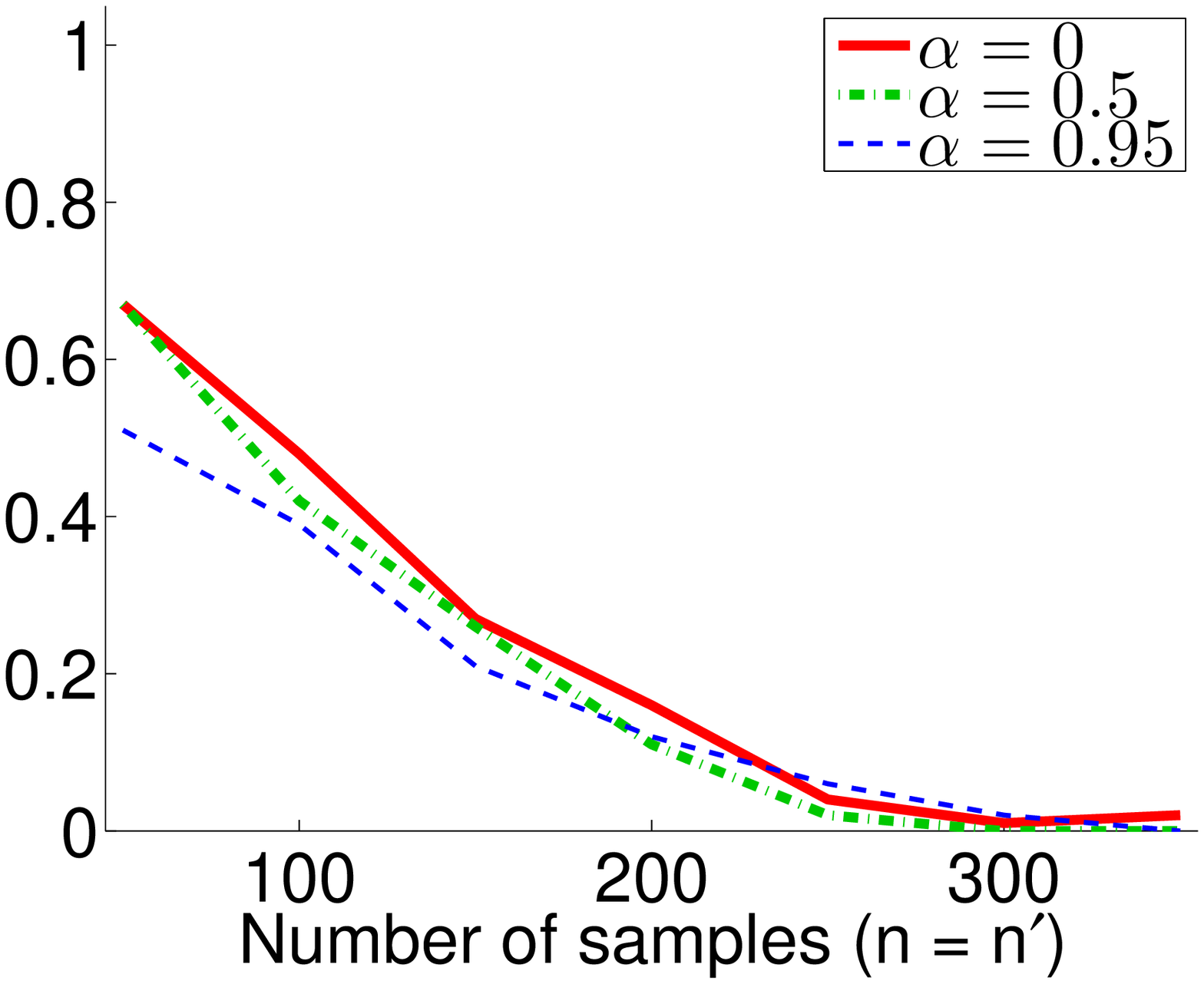}
    \end{tabular}
  }
  \subfigure[$\Pde=N(0.5,1)$: $\Pnu$ and $\Pde$ have different means.]{
    \begin{tabular}{@{}c@{}c@{}c@{}c@{}}
      \includegraphics[width=.24\textwidth]{dataset4-density.eps}&
      \includegraphics[width=.24\textwidth]{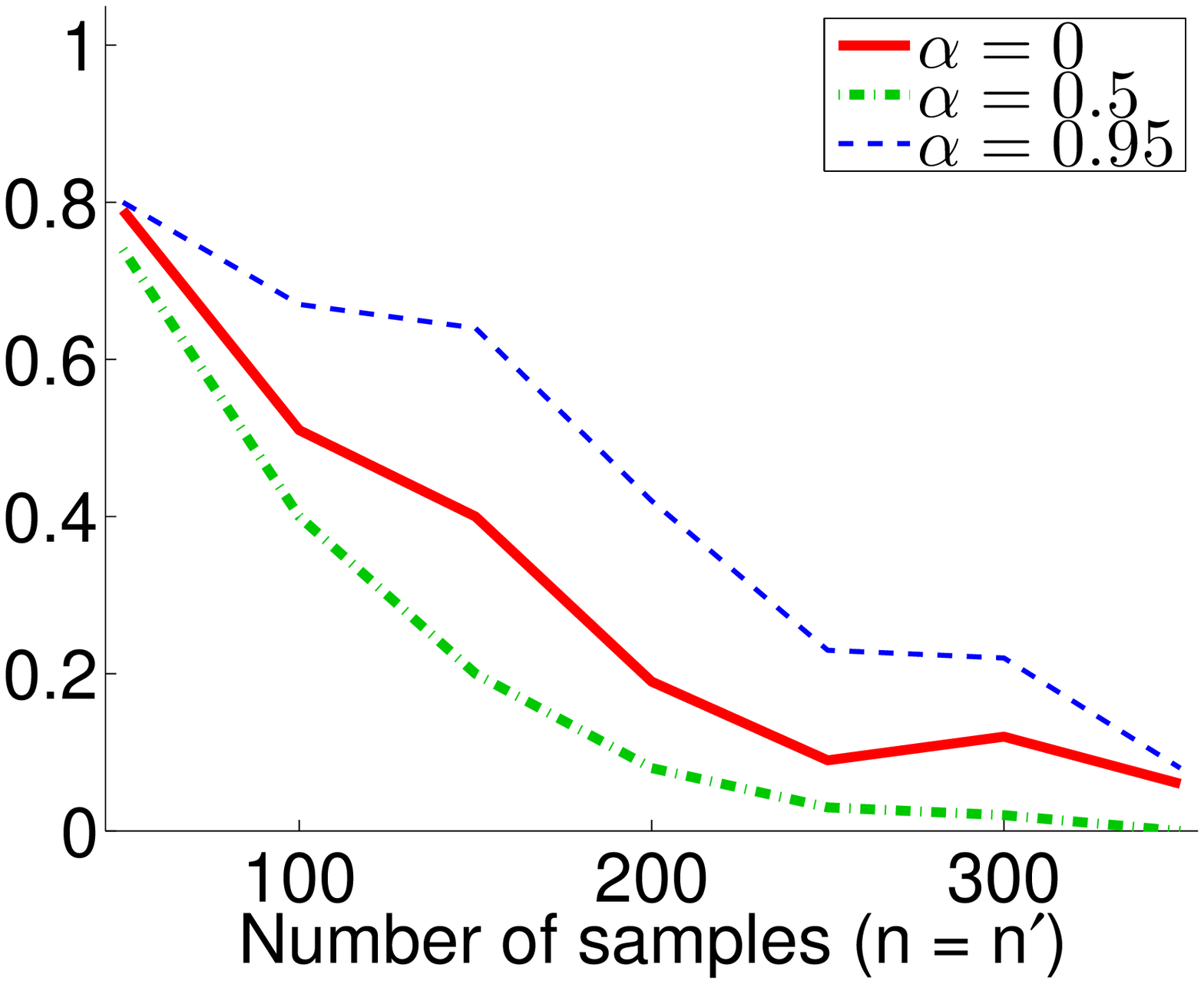}&
      \includegraphics[width=.24\textwidth]{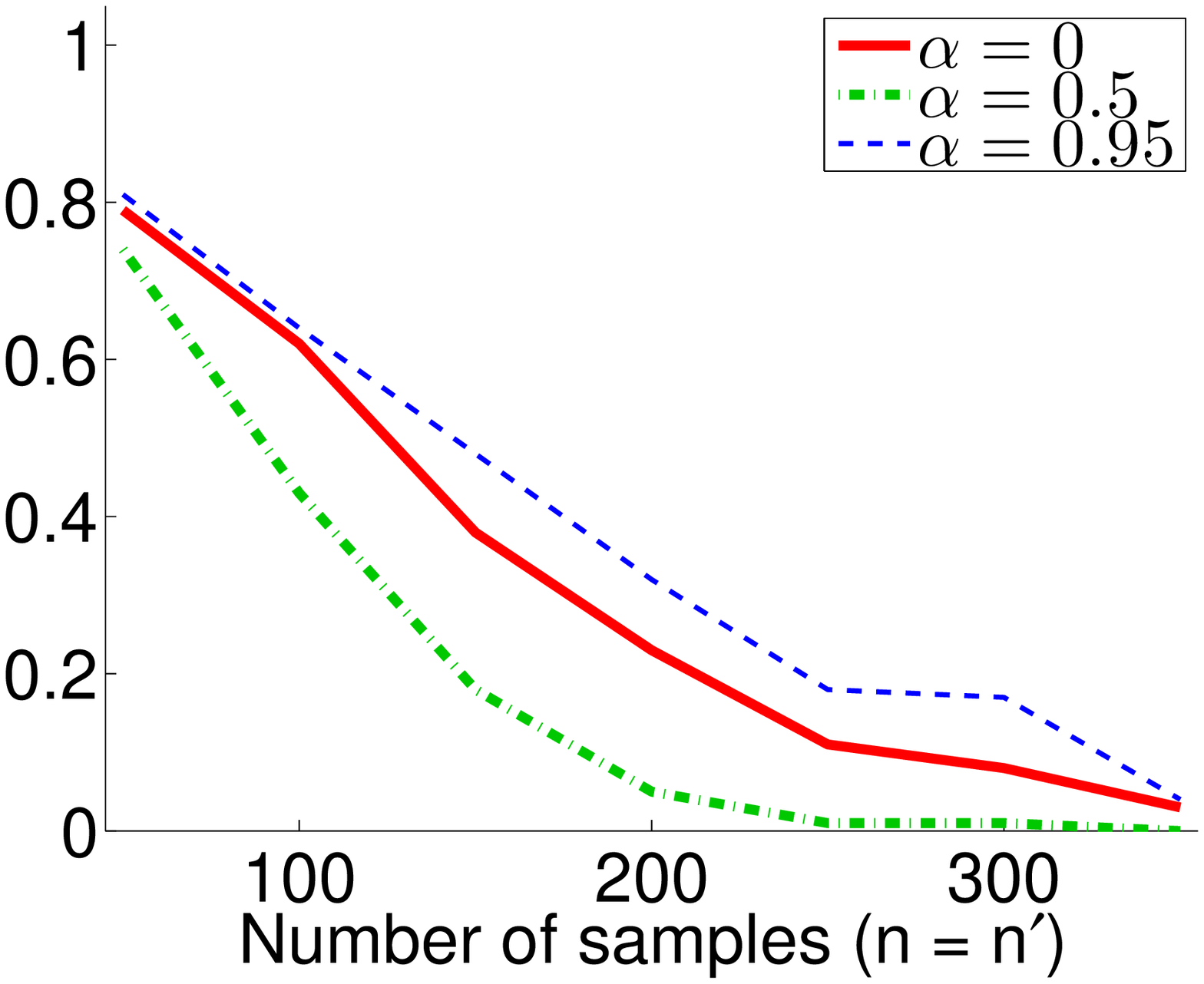}&
      \includegraphics[width=.24\textwidth]{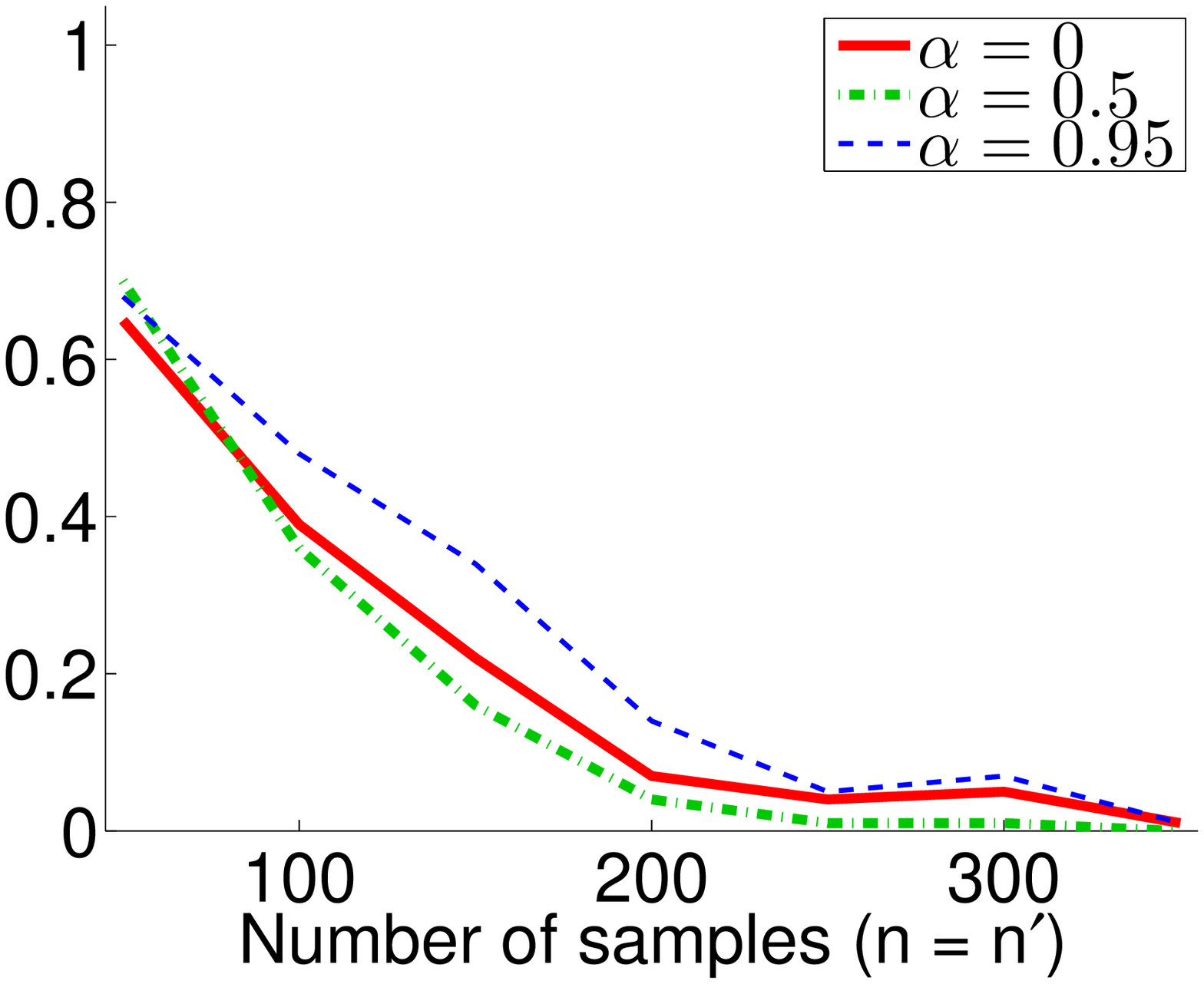}
    \end{tabular}
  }
  \caption{Illustrative examples of two-sample homogeneity test
    based on relative divergence estimation. 
    From left to right: true densities ($\Pnu = N(0,1)$),
    the acceptance rate of the null hypothesis 
    under the significance level $5\%$
    by plain LSTT, reciprocal LSTT, and adaptive LSTT.
  }
  \label{fig:illustrative-two-sample}
\end{figure}

\subsubsection{Artificial Datasets}
We illustrate how the proposed method behaves
in two-sample homogeneity test scenarios using 
the artificial datasets (a)--(d) described in Section~\ref{subsec:illustration}.
We test the plain LSTT, reciprocal LSTT, and adaptive LSTT
for $\alpha=0$, $0.5$, and $0.95$,
with significance level $5\%$.

The experimental results are shown in Figure~\ref{fig:illustrative-two-sample}.
For the dataset (a) where $\Pnu=\Pde$ (i.e., the null hypothesis is correct),
the plain LSTT and reciprocal LSTT
correctly accept the null hypothesis with probability approximately $95\%$.
This means that the type-I error is properly controlled in these methods.
On the other hand, the adaptive LSTT tends to give slightly lower acceptance rates
than $95\%$ for this toy dataset,
but the adaptive LSTT with $\alpha=0.5$ still works reasonably well.
This implies that the heuristic of choosing the method with a smaller p-value
does not have critical influence on the type-I error.

In the datasets (b), (c), and (d), $\Pnu$ is different from $\Pde$
(i.e., the null hypothesis is not correct),
and thus we want to reduce the acceptance rate of the incorrect null-hypothesis
as much as possible.
In the plain setup for the dataset (b) and the reciprocal setup for the dataset (c),
the true density-ratio functions with $\alpha=0$ diverge to infinity,
and thus larger $\alpha$ makes the density-ratio approximation more reliable.
However, $\alpha=0.95$ does not work well
because it produces an overly-smoothed density-ratio function
and thus it is hard to be distinguished from the completely constant density-ratio function
(which corresponds to $\Pnu=\Pde$).
On the other hand,
in the reciprocal setup for the dataset (b) and the plain setup for the dataset (c),
small $\alpha$ performs poorly since density-ratio functions with large $\alpha$ 
can be more accurately approximated than those with small $\alpha$
(see Figure~\ref{fig:illustrative-ratio}).
In the adaptive setup, large $\alpha$ tends to perform slightly
better than small $\alpha$ for the datasets (b) and (c).

In the dataset (d), the true density-ratio function with $\alpha=0$
diverges to infinity for both the plain and reciprocal setups.
In this case, middle $\alpha$ performs the best,
which well balances the trade-off between
high distinguishability from the completely constant density-ratio function
(which corresponds to $\Pnu=\Pde$)
and easy approximability.
The same tendency that middle $\alpha$ works well
can also be mildly observed in the adaptive LSTT
for the dataset (d).

Overall, if the plain LSTT (or the reciprocal LSTT) is used,
small $\alpha$ (or large $\alpha$) sometimes works excellently.
However, it performs poorly in other cases
and thus the performance is unstable depending on the true distributions.
The plain LSTT (or the reciprocal LSTT) with middle $\alpha$
tends to perform reasonably well for all datasets.
On the other hand, the adaptive LSTT was shown to nicely 
overcome the above instability problem
when $\alpha$ is small or large.
However, when $\alpha$ is set to be a middle value,
the plain LSTT and the reciprocal LSTT both give similar results
and thus the adaptive LSTT provides only a small amount of improvement.

Our empirical finding is that,
if we have prior knowledge that
one distribution has a wider support than the other distribution,
assigning the distribution with a wider support to $\Pde$
and setting $\alpha$ to be a large value
seem to work well.
If there is no knowledge on the true distributions
or two distributions have less overlapped supports,
using middle $\alpha$ in the adaptive setup seems to be a reasonable choice.

We will systematically investigate this issue using 
more complex datasets below.

\begin{table}[p]
\centering
\caption{Experimental results of two-sample homogeneity test for the IDA datasets.
  The mean (and standard deviation in the bracket) 
  rate of accepting the null hypothesis (i.e., $\Pnu = \Pde$)
  under the significance level $5\%$ is reported.
  The two sets of samples are both taken from the positive training set
  (i.e., the null hypothesis is correct).
  The best method having the highest mean acceptance rate and comparable methods 
  according to the \emph{t-test} at the significance level $5\%$ are specified by bold face.}
\label{tb:twosample_result_P=P'}
\begin{tabular}{|l|c|c||l@{}r|l@{}r|l@{}r|l@{}r|l@{}r|l@{}r|l@{}r}
\hline
\multirow{2}{*}{Datasets}  & \multirow{2}{*}{$d$} & \multirow{2}{*}{$n = n'$} & \multicolumn{2}{c|}{\multirow{2}{*}{MMD}}  & \multicolumn{2}{c|}{LSTT} &  \multicolumn{2}{c|}{LSTT} &  \multicolumn{2}{c|}{LSTT}  \\
    &    &    &   &            & \multicolumn{2}{c|}{($\alpha$ = 0.0)}   & \multicolumn{2}{c|}{($\alpha$ = 0.5)}          &  \multicolumn{2}{c|}{($\alpha$ = 0.95)} \\
\hline \hline%
banana&2&100&\textbf{0.98}&\textbf{(0.14)}&\textbf{0.93}&\textbf{(0.26)}&\textbf{0.92}&\textbf{(0.27)}&\textbf{0.92}&\textbf{(0.27)}\\\cline{1-11}
thyroid&5&19&\textbf{0.98}&\textbf{(0.14)}&\textbf{0.95}&\textbf{(0.22)}&\textbf{0.95}&\textbf{(0.22)}&0.88&(0.33)\\\hline
titanic&5&21&\textbf{0.92}&\textbf{(0.27)}&\textbf{0.86}&\textbf{(0.35)}&\textbf{0.92}&\textbf{(0.27)}&\textbf{0.89}&\textbf{(0.31)}\\\cline{1-11}
diabetes&8&85&\textbf{0.95}&\textbf{(0.22)}&0.87&(0.34)&\textbf{0.91}&\textbf{(0.29)}&0.82&(0.39)\\\hline
breast-cancer&9&29&\textbf{0.98}&\textbf{(0.14)}&0.91&(0.29)&\textbf{0.94}&\textbf{(0.24)}&\textbf{0.92}&\textbf{(0.27)}\\\cline{1-11}
flare-solar&9&100&\textbf{0.93}&\textbf{(0.26)}&\textbf{0.91}&\textbf{(0.29)}&\textbf{0.95}&\textbf{(0.22)}&\textbf{0.93}&\textbf{(0.26)}\\\hline
heart&13&38&\textbf{0.96}&\textbf{(0.20)}&0.85&(0.36)&\textbf{0.91}&\textbf{(0.29)}&\textbf{0.93}&\textbf{(0.26)}\\\cline{1-11}
german&20&100&\textbf{0.93}&\textbf{(0.26)}&\textbf{0.91}&\textbf{(0.29)}&\textbf{0.92}&\textbf{(0.27)}&\textbf{0.89}&\textbf{(0.31)}\\\hline
ringnorm&20&100&\textbf{0.95}&\textbf{(0.22)}&\textbf{0.93}&\textbf{(0.26)}&\textbf{0.91}&\textbf{(0.29)}&0.85&(0.36)\\\cline{1-11}
waveform&21&66&\textbf{0.93}&\textbf{(0.26)}&\textbf{0.92}&\textbf{(0.27)}&\textbf{0.93}&\textbf{(0.26)}&\textbf{0.88}&\textbf{(0.33)}\\\cline{1-11}
\end{tabular}
\vspace*{10mm}
\caption{Experimental results of two-sample homogeneity test for the IDA datasets.
  The mean (and standard deviation in the bracket) 
  rate of accepting the null hypothesis (i.e., $\Pnu = \Pde$)
  under the significance level $5\%$ is reported.
  The set of samples corresponding to the numerator of the density ratio
  is taken from the positive training set
  and the set of samples corresponding to the denominator of the density ratio
  is taken from the positive training set and the negative training set
  (i.e., the null hypothesis is not correct).
  The best method having the lowest mean acceptance rate and comparable methods 
  according to the \emph{t-test} at the significance level $5\%$ are specified by bold face.}
\label{tb:twosample_result_PneqP'}
\begin{tabular}{|l|c|c||l@{}r|l@{}r|l@{}r|l@{}r|l@{}r|l@{}r|l@{}r}
\hline
\multirow{2}{*}{Datasets}  & \multirow{2}{*}{$d$} & \multirow{2}{*}{$n = n'$} & \multicolumn{2}{c|}{\multirow{2}{*}{MMD}}  & \multicolumn{2}{c|}{LSTT} &  \multicolumn{2}{c|}{LSTT} &  \multicolumn{2}{c|}{LSTT}  \\
    &    &    &   &            & \multicolumn{2}{c|}{($\alpha$ = 0.0)}   & \multicolumn{2}{c|}{($\alpha$ = 0.5)}          &  \multicolumn{2}{c|}{($\alpha$ = 0.95)} \\
\hline \hline%
banana&2&100&0.80&(0.40)&\textbf{0.10}&\textbf{(0.30)}&\textbf{0.02}&\textbf{(0.14)}&\textbf{0.17}&\textbf{(0.38)}\\\cline{1-11}
thyroid&5&19&0.72&(0.45)&0.81&(0.39)&\textbf{0.65}&\textbf{(0.48)}&0.80&(0.40)\\\hline
titanic&5&21&\textbf{0.79}&\textbf{(0.41)}&\textbf{0.86}&\textbf{(0.35)}&\textbf{0.87}&\textbf{(0.34)}&\textbf{0.88}&\textbf{(0.33)}\\\cline{1-11}
diabetes&8&85&\textbf{0.38}&\textbf{(0.49)}&\textbf{0.42}&\textbf{(0.50)}&0.47&(0.50)&0.57&(0.50)\\\hline
breast-cancer&9&29&0.91&(0.29)&\textbf{0.75}&\textbf{(0.44)}&\textbf{0.80}&\textbf{(0.40)}&\textbf{0.79}&\textbf{(0.41)}\\\cline{1-11}
flare-solar&9&100&\textbf{0.59}&\textbf{(0.49)}&0.81&(0.39)&\textbf{0.55}&\textbf{(0.50)}&\textbf{0.66}&\textbf{(0.48)}\\\hline
heart&13&38&\textbf{0.47}&\textbf{(0.50)}&\textbf{0.28}&\textbf{(0.45)}&\textbf{0.40}&\textbf{(0.49)}&0.62&(0.49)\\\cline{1-11}
german&20&100&0.59&(0.49)&0.55&(0.50)&\textbf{0.44}&\textbf{(0.50)}&0.68&(0.47)\\\hline
ringnorm&20&100&\textbf{0.00}&\textbf{(0.00)}&\textbf{0.00}&\textbf{(0.00)}&\textbf{0.00}&\textbf{(0.00)}&\textbf{0.02}&\textbf{(0.14)}\\\cline{1-11}
waveform&21&66&\textbf{0.00}&\textbf{(0.00)}&\textbf{0.00}&\textbf{(0.00)}&\textbf{0.02}&\textbf{(0.14)}&\textbf{0.00}&\textbf{(0.00)}\\\cline{1-11}
\end{tabular}
\end{table}

\subsubsection{Benchmark Datasets}
Here, we apply the proposed two-sample homogeneity test to 
the binary classification datasets 
taken from the \emph{IDA repository} \citep{mach:raetsch+onoda+mueller:2001}.

We test the adaptive LSTT with the RuLSIF-based PE divergence estimator 
for $\alpha=0$, $0.5$, and $0.95$;
we also test the \emph{maximum mean discrepancy}
\citep[MMD;][]{Bioinformatics:Borgwardt+etal:2006},
which is a kernel-based two-sample homogeneity test method.
The performance of MMD depends on the choice of the Gaussian kernel width.
Here, we adopt a version proposed by \citet{NIPS2009_0893},
which automatically optimizes the Gaussian kernel width.
The p-values of MMD are computed in the same way as LSTT
based on the permutation test procedure.

First, we investigate the rate of accepting the null hypothesis
when the null hypothesis is correct (i.e., the two distributions are the same).
We split all the positive training samples into two sets
and perform two-sample homogeneity test for the two sets of samples.
The experimental results are summarized in Table~\ref{tb:twosample_result_P=P'},
showing that the adaptive LSTT with $\alpha=0.5$ compares
favorably with that with $\alpha=0$ and $1$.
LSTT with $\alpha=0.5$ and MMD are comparable to each other 
in terms of the type-I error.

Next, we consider the situation where
the null hypothesis is not correct (i.e., the two distributions are different).
The numerator samples are generated in the same way as above,
but a half of denominator samples are replaced with negative training samples.
Thus, while the numerator sample set contains only positive training samples,
the denominator sample set includes both positive and negative training samples.
The experimental results are summarized in Table~\ref{tb:twosample_result_PneqP'},
showing that the adaptive LSTT with $\alpha=0.5$ again compares
favorably with that with $\alpha=0$ and $1$.
Furthermore, LSTT with $\alpha=0.5$ tends to outperform
MMD in terms of the type-II error.

Overall, LSTT with $\alpha=0.5$ is shown to be a useful method for two-sample homogeneity test.

\subsection{Inlier-Based Outlier Detection}
Next, we apply the proposed method to outlier detection.

\subsubsection{Density-Ratio Approach to Inlier-Based Outlier Detection}
Let us consider an outlier detection problem of
finding irregular samples in a dataset (called an ``evaluation dataset'')
based on another dataset (called a ``model dataset'') that only contains regular samples.
Defining the density ratio over the two sets of samples,
we can see that the density-ratio values for regular samples are close to one,
while those for outliers tend to be significantly deviated from one.
Thus, density-ratio values
could be used as an index of the degree of outlyingness
\citep{AISTATS:Smola+etal:2009,KAIS:Hido+etal:2011}.

Since the evaluation dataset usually has a wider support than the model dataset,
we regard the evaluation dataset as samples
corresponding to the denominator density $\pde(\boldx)$,
and the model dataset as samples
corresponding to the numerator density $\pnu(\boldx)$.
Then, outliers tend to have smaller density-ratio values (i.e., close to zero).
As such, density-ratio approximators can be used for outlier detection.

When evaluating the performance of outlier detection methods, it is
important to take into account both the \emph{detection rate} (i.e.,
the amount of true outliers an outlier detection algorithm can find)
and the \emph{detection accuracy} (i.e., the amount of true inliers an
outlier detection algorithm misjudges as outliers). Since there is a
trade-off between the detection rate and the detection accuracy, we
adopt the \emph{area under the ROC curve} (AUC) as our error metric
\citep{PR:KBradley:1997}.

\begin{table}[t]
\centering
\caption{Mean AUC score (and the standard deviation in the bracket)
over $1000$ trials for the artificial outlier-detection dataset.
The best method in terms of the mean AUC score and
comparable methods according to the \emph{t-test} at the significance
level $5\%$ are specified by bold face.}
\label{tb:toy_outlier_result}
\begin{tabular}{|c|l@{}r|l@{}r|l@{}r|}
\hline
  \begin{tabular}{@{}c@{}}
Input\\
dimensionality $d$
\end{tabular}
& \multicolumn{2}{c|}{
  \begin{tabular}{@{}c@{}}
    RuLSIF \\
    ($\alpha=0$)    
  \end{tabular}
}
& \multicolumn{2}{c|}{
  \begin{tabular}{@{}c@{}}
    RuLSIF \\
    ($\alpha=0.5$)    
  \end{tabular}
}
& \multicolumn{2}{c|}{
  \begin{tabular}{@{}c@{}}
    RuLSIF \\
    ($\alpha=0.95$)    
  \end{tabular}
} \\
\hline \hline%
1 & {\bf .933} & {\bf (.089)} & {\bf .926} & {\bf (.100)} & .896 & (.124) \\ \hline
5 & {\bf .882} & {\bf (.099)} & {\bf .891} & {\bf (.091)} & {\bf .894} & {\bf (.086)} \\ \hline
10& .842 & (.107) & {\bf .850} & {\bf (.103)} & {\bf .859} & {\bf (.092)} \\
\hline
\end{tabular}
\end{table}


\subsubsection{Artificial Datasets}
\label{sec:experiment-outlier-artificial}
First, we illustrate how the proposed method behaves
in outlier detection scenarios using artificial datasets.

Let 
\begin{align*}
  \Pnu&=N(0,\boldI_\inputdim),\\
  \Pde&=0.95 N(0,\boldI_\inputdim)+0.05 N(3\inputdim^{-1/2}\boldone_\inputdim,\boldI_\inputdim),
\end{align*}
where $\inputdim$ is the dimensionality of $\boldx$
and $\boldone_\inputdim$ is the $\inputdim$-dimensional vector with all one.
Note that this setup is the same as 
the dataset (e) described in Section~\ref{subsec:illustration}
when $\inputdim=1$.
Here, the samples drawn from $N(0,\boldI_\inputdim)$ are regarded as inliers,
while the samples drawn from $N(\inputdim^{-1/2}\boldone_\inputdim,\boldI_\inputdim)$
are regarded as outliers.
We use $\nnu=\nde=100$ samples.

Table~\ref{tb:toy_outlier_result} 
describes the AUC values for input dimensionality $\inputdim=1$, $5$, and $10$
for RuLSIF with $\alpha=0$, $0.5$, and $0.95$.
This shows that, as the input dimensionality $\inputdim$ increases,
the AUC values overall get smaller.
Thus, outlier detection becomes more challenging in high-dimensional cases.

The result also shows that
RuLSIF with small $\alpha$ tends to work well when the input dimensionality is low,
and RuLSIF with large $\alpha$ works better as the input dimensionality increases.
This tendency can be interpreted as follows:
If $\alpha$ is small, the density-ratio function tends to have sharp `hollow'
for outlier points (see the leftmost graph in Figure~\ref{fig:illustrative-divergence-outlier}).
Thus, as long as the true density-ratio function can be accurately estimated,
small $\alpha$ would be preferable in outlier detection.
When the data dimensionality is low,
density-ratio approximation is rather easy and thus
small $\alpha$ tends to perform well.
However, as the data dimensionality increases,
density-ratio approximation gets harder,
and thus large $\alpha$ which produces a smoother density-ratio function
is more favorable since such a smoother function can be more easily
approximated than a `bumpy' one produced by small $\alpha$.

\subsubsection{Real-World Datasets}

Next, we evaluate the proposed outlier detection method using various real-world datasets:
\begin{description}
\item[IDA repository:]
The \emph{IDA repository} \citep{mach:raetsch+onoda+mueller:2001}
contains various binary classification tasks.
Each dataset consists of positive/negative and training/test samples.
We use positive training samples as inliers in the ``model'' set.
In the ``evaluation'' set,
we use at most $100$ positive test samples as inliers and the first $5\%$
of negative test samples as outliers. 
Thus, the positive samples are treated as inliers
and the negative samples are treated as outliers.


\item[Speech dataset:]
  An in-house speech dataset, which contains short utterance samples
recorded from $2$ male subjects speaking in French with sampling rate
$44.1$kHz. From each utterance sample, we extracted a $50$-dimensional
\emph{line spectral frequencies} vector \citep{ICASSP:Kain+etal:1998}. We
randomly take $200$ samples from one class and assign them to the model
dataset. Then we randomly take $200$ samples from the same class
and $10$ samples from the other class.

\item[20 Newsgroup dataset:]
 The \emph{20-Newsgroups} dataset\footnote{
\url{http://people.csail.mit.edu/jrennie/20Newsgroups/}}
contains $20000$ newsgroup documents,
which contains the following $4$ top-level categories: `comp', `rec', `sci', and
`talk'. Each document is expressed by a $100$-dimensional bag-of-words
vector of term-frequencies.  We randomly take $200$ samples from the
`comp' class and assign them to the model dataset. Then we randomly
take $200$ samples from the same class and $10$
samples from one of the other classes for the evaluation dataset.

\item[The USPS hand-written digit dataset:]
The \emph{USPS} hand-written digit dataset\footnote{
\url{http://www.gaussianprocess.org/gpml/data/}}
contains $9298$ digit images.
Each image consists of $256$ (= $16\times 16$) pixels and each pixel
takes an integer value between $0$ and $255$ as the intensity level. We
regard samples in one class as inliers and samples in other classes as
outliers. We randomly take $200$ samples from the inlier class and assign them
to the model dataset. Then we randomly take $200$ samples from the
same inlier class and $10$ samples from
one of the other classes for the evaluation dataset. 

\end{description}

We compare the AUC scores of RuLSIF with $\alpha=0$,
$0.5$, and $0.95$, and \emph{one-class support vector machine (OSVM)}
with the Gaussian kernel \citep{NC:Scholkopf+etal:2001}. 
We used the \emph{LIBSVM} implementation of OSVM \citep{CC01a}.
The Gaussian width is set to the median distance between samples,
which has been shown to be a useful heuristic \citep{NC:Scholkopf+etal:2001}.
Since there is no systematic method to determine
the tuning parameter $\nu$ in OSVM,
we report the results for $\nu=0.05$ and $0.1$.

The mean and standard deviation of the AUC scores over $100$ runs with
random sample choice are summarized in
Table~\ref{tb:outlier_result}, showing that RuLSIF overall
compares favorably with OSVM. 
Among the RuLSIF methods,
small $\alpha$ tends to perform well for low-dimensional datasets,
and large $\alpha$ tends to work well for high-dimensional datasets.
This tendency well agrees with that for the artificial datasets
(see Section~\ref{sec:experiment-outlier-artificial}).

\begin{table}[p]
\centering
\caption{Experimental results of outlier detection
  for various for real-world datasets.
  Mean AUC score (and standard deviation in the bracket) over
  $100$ trials is reported. 
  The best method having the highest mean AUC score and
  comparable methods according to the \emph{t-test} at the significance
  level $5\%$ are specified by bold face.
  The datasets are sorted in the ascending order of the input dimensionality $\inputdim$.
}
\label{tb:outlier_result}
\begin{tabular}{|@{\ }l@{\ }|@{\ }c@{\ }||l@{}r|l@{}r|l@{}r|l@{}r|l@{}r|l@{}r|l@{}r}
\hline
Datasets  & $\inputdim$ 
&\multicolumn{2}{c|}{
  \begin{tabular}{@{}c@{}}
    OSVM \\
    ($\nu=0.05$)    
  \end{tabular}
} 
&  \multicolumn{2}{c|}{
  \begin{tabular}{@{}c@{}}
    OSVM \\
    ($\nu=0.1$)    
  \end{tabular}
}
& \multicolumn{2}{c|}{
  \begin{tabular}{@{}c@{}}
    RuLSIF \\
    ($\alpha=0$)    
  \end{tabular}
}
& \multicolumn{2}{c|}{
  \begin{tabular}{@{}c@{}}
    RuLSIF \\
    ($\alpha=0.5$)    
  \end{tabular}
}
& \multicolumn{2}{c|}{
  \begin{tabular}{@{}c@{}}
    RuLSIF \\
    ($\alpha=0.95$)    
  \end{tabular}
} \\
\hline \hline
IDA:banana   & 2  & {\bf .668} & {\bf (.105)} & {\bf .676} & {\bf (.120)} & .597 & (.097) & .619  & (.101) & .623 & (.115) \\ \hline
IDA:thyroid   & 5  & .760 & (.148) & {\bf .782} & {\bf (.165)} & {\bf .804} & {\bf (.148)} & {\bf .796}  & {\bf (.178)} & .722 & (.153) \\ \hline
IDA:titanic   & 5  & {\bf .757} & {\bf (.205)} & {\bf .752} & {\bf (.191)} & {\bf .750} & {\bf (.182)} & .701  & (.184) & .712 & (.185)\\ \hline
IDA:diabetes  & 8  & {\bf .636} & {\bf (.099)} & .610 & (.090) & .594 & (.105) & .575  & (.105) & {\bf .663} & {\bf (.112)} \\ \hline
IDA:b-cancer  & 9  & {\bf .741} & {\bf (.160)} & .691 & (.147) & {\bf .707} & {\bf (.148)} & {\bf .737}  & {\bf (.159)} & {\bf .733} & {\bf (.160)} \\ \hline
IDA:f-solar  & 9  & .594 & (.087) & .590 & (.083) & {\bf .626} & {\bf (.102)} & {\bf .612}  & {\bf (.100)} & .584 & (.114) \\ \hline
IDA:heart   & 13 & .714 & (.140) & .694 & (.148) & {\bf .748} & {\bf (.149)} & {\bf .769}  & {\bf (.134)} & .726 & (.127)\\ \hline
IDA:german   & 20 & {\bf .612} & {\bf (.069)} & {\bf .604} & {\bf (.084)} & {\bf .605} & {\bf (.092)} & {\bf .597}  & {\bf (.101)} & {\bf .605} & {\bf (.095)} \\ \hline
IDA:ringnorm  & 20 & {\bf .991} & {\bf (.012)} & {\bf .993} & {\bf (.007)} & .944 & (.091) & .971  & (.062) & {\bf .992} & {\bf (.010)} \\ \hline
IDA:waveform  & 21 & .812 & (.107) & .843 & (.123) & {\bf .879} & {\bf (.122)} & {\bf .875}  & {\bf (.117)} & {\bf .885} & {\bf (.102)} \\ \hline
Speech                & 50     & .788 & (.068) & {\bf .830}  & {\bf (.060)} & .804   & (.101) & {\bf .821}  & {\bf (.076)} & {\bf .836} & {\bf (.083)}\\ \hline
20News (`rec')   & 100   & .598 & (.063) & .593  & (.061)& .628  & (.105) & .614  & (.093) & {\bf .767} & {\bf (.100)} \\ \hline
20News (`sci')   & 100   & .592 & (.069) & .589  & (.071)& .620  & (.094) & .609  & (.087) & {\bf .704} & {\bf (.093)} \\ \hline
20News (`talk')   & 100   & .661 & (.084) & .658  & (.084)& .672  & (.117) & .670  & (.102) & {\bf .823} & {\bf (.078)} \\ \hline
USPS (1 vs.~2)  & 256 &  .889       & (.052)       & {\bf .926}  & {\bf (.037)} & .848 & (.081) & .878 & (.088) & .898  & (.051)\\ \hline
USPS (2 vs.~3)  & 256 & .823        & (.053)       & .835  & (.050)   & .803 & (.093) & .818 & (.085) & {\bf .879}  & {\bf (.074)}\\ \hline
USPS (3 vs.~4)  & 256 & .901        & (.044)       & .939  & (.031)   & .950 & (.056) & .961 & (.041) & {\bf .984}  & {\bf (.016)}\\ \hline
USPS (4 vs.~5)  & 256 & .871        & (.041)       & .890  & (.036)   & .857 & (.099) & .874 & (.082) & {\bf .941}  & {\bf (.031)}\\ \hline
USPS (5 vs.~6)  & 256 & .825        & (.058)       & .859  & (.052)   & .863 & (.078) & .867 & (.068) & {\bf .901}  & {\bf (.049)}\\ \hline
USPS (6 vs.~7)  & 256 & .910        & (.034)       & .950  & (.025)   & .972 & (.038) & .984 & (.018) & {\bf .994}  & {\bf (.010)}\\ \hline
USPS (7 vs.~8)  & 256 & .938        & (.030)       & .967  & (.021)   & .941 & (.053) & .951 & (.039) & {\bf .980}  & {\bf (.015)}\\ \hline
USPS (8 vs.~9)  & 256 & .721        & (.072)       & .728  & (.073)   & .721 & (.084) & .728 & (.083) & {\bf .761}  & {\bf (.096)}\\ \hline
USPS (9 vs.~0)  & 256 & .920        & (.037)       & .966  & (.023)   & .982 & (.048) & .989 & (.022) & {\bf .994}  & {\bf (.011)}\\
\hline
\end{tabular}
\end{table}

\subsection{Transfer Learning}
Finally, we apply the proposed method to outlier detection.

\subsubsection{Transductive Transfer Learning by Importance Sampling}
Let us consider a problem of \emph{semi-supervised learning}
\citep{book:Chapelle+etal:2006} from labeled training samples
$\{(\boldx^{\mathrm{tr}}_j,y^{\mathrm{tr}}_j)\}_{j=1}^{n_{\mathrm{tr}}}$
and unlabeled test samples $\{\boldx^{\mathrm{te}}_i\}_{i=1}^{n_{\mathrm{te}}}$.
The goal is to predict a test output value $y^{\mathrm{te}}$
for a test input point $\boldx^{\mathrm{te}}$.
Here, we consider the setup where
the labeled training samples
$\{(\boldx^{\mathrm{tr}}_j,y^{\mathrm{tr}}_j)\}_{j=1}^{n_{\mathrm{tr}}}$
are drawn i.i.d.~from $p(y|\boldx)p_{\mathrm{tr}}(\boldx)$,
while the unlabeled test samples $\{\boldx^{\mathrm{te}}_i\}_{i=1}^{n_{\mathrm{te}}}$
are drawn i.i.d.~from $p_{\mathrm{te}}(\boldx)$, which is generally different
from $p_{\mathrm{tr}}(\boldx)$;
the (unknown) test sample $(\boldx^{\mathrm{te}},y^{\mathrm{te}})$ follows
$p(y|\boldx)p_{\mathrm{te}}(\boldx)$.
This setup means that the conditional probability $p(y|\boldx)$ is common
to training and test samples,
but the marginal densities $p_{\mathrm{tr}}(\boldx)$ and $p_{\mathrm{te}}(\boldx)$
are generally different for training and test input points.
Such a problem is called
\emph{transductive transfer learning} \citep{IEEE-KDE:Pan+Yang:2010},
\emph{domain adaptation} \citep{ACL:Jiang+Zhai:2007},
or \emph{covariate shift} \citep{JSPI:Shimodaira:2000,book:Sugiyama+Kawanabe:2011}.

Let $\mathrm{loss}(y,\widehat{y})$ be a point-wise loss function 
that measures a discrepancy between $y$ and $\widehat{y}$ (at input $\boldx$).
Then the \emph{generalization error} which we would like to ultimately minimize is defined as
\begin{align*}
  \mathbbE_{p(y|\boldx)p_{\mathrm{te}}(\boldx)}\left[\mathrm{loss}(y,f(\boldx))\right],
\end{align*}
where $f(\boldx)$ is a function model.
Since the generalization error is inaccessible because
the true probability $p(y|\boldx)p_{\mathrm{te}}(\boldx)$ is unknown,
empirical-error minimization is often used in practice \citep{book:Vapnik:1998}:
\begin{align*}
  \min_{f\in\calF}
  \left[
  \frac{1}{n_{\mathrm{tr}}} \sum_{j=1}^{n_{\mathrm{tr}}}
  \mathrm{loss}(y^{\mathrm{tr}}_j,f(\boldx^{\mathrm{tr}}_j))
  \right].
\end{align*}
However, under the covariate shift setup, plain empirical-error minimization is
not \emph{consistent} (i.e., it does not converge to
the optimal function) if the model $\calF$ is \emph{misspecified}
\citep[i.e., the true function is not included in the model; see][]{JSPI:Shimodaira:2000}.
Instead,
the following \emph{importance-weighted} empirical-error minimization is consistent
under covariate shift:
\begin{align*}
  \min_{f\in\calF}
  \left[
  \frac{1}{n_{\mathrm{tr}}} \sum_{j=1}^{n_{\mathrm{tr}}}
  \ratio(\boldx^{\mathrm{tr}}_j)\mathrm{loss}(y^{\mathrm{tr}}_j,f(\boldx^{\mathrm{tr}}_j))
  \right],
\end{align*}
where $\ratio(\boldx)$ is called the \emph{importance} \citep{book:Fishman:1996}
in the context of covariate shift adaptation:
\begin{align*}
  \ratio(\boldx):=\frac{p_{\mathrm{te}}(\boldx)}{p_{\mathrm{tr}}(\boldx)}.    
\end{align*}

However, since importance-weighted learning is not \emph{statistically efficient}
(i.e., it tends to have larger variance),
slightly \emph{flattening} the importance weights is practically useful
for stabilizing the estimator.
\citet{JSPI:Shimodaira:2000} proposed to use the \emph{exponentially-flattened importance weights}
as
\begin{align*}
  \min_{f\in\calF}
  \left[
  \frac{1}{n_{\mathrm{tr}}} \sum_{j=1}^{n_{\mathrm{tr}}}
  \ratio(\boldx^{\mathrm{tr}}_j)^\tau\mathrm{loss}(y^{\mathrm{tr}}_j,f(\boldx^{\mathrm{tr}}_j))
  \right],
\end{align*}
where $0\le\tau\le 1$ is called the \emph{exponential flattening parameter}.
$\tau=0$ corresponds to plain empirical-error minimization,
while $\tau=1$ corresponds to importance-weighted empirical-error minimization;
$0<\tau<1$ will give an intermediate estimator that balances the trade-off
between statistical efficiency and consistency.
The exponential flattening parameter $\tau$ can be optimized by
model selection criteria
such as 
the \emph{importance-weighted Akaike information criterion} 
for regular models \citep{JSPI:Shimodaira:2000},
the \emph{importance-weighted subspace information criterion}
for linear models \citep{StatDeci:Sugiyama+Mueller:2005},
and
\emph{importance-weighted cross-validation}
for arbitrary models \citep{JMLR:Sugiyama+etal:2007}.

One of the potential drawbacks of the above exponential flattering approach
is that estimation of $\ratio(\boldx)$ (i.e., $\tau=1$) is rather hard, as shown in this paper.
Thus, when $\ratio(\boldx)$ is estimated poorly, 
all flattened weights $\ratio(\boldx)^\tau$ are also unreliable
and then covariate shift adaptation does not work well in practice.
To cope with this problem, we propose to use \emph{relative importance weights}
alternatively:
\begin{align*}
  \min_{f\in\calF}
  \left[
  \frac{1}{n_{\mathrm{tr}}} \sum_{j=1}^{n_{\mathrm{tr}}}
  \relratio(\boldx^{\mathrm{tr}}_j)\mathrm{loss}(y^{\mathrm{tr}}_j,f(\boldx^{\mathrm{tr}}_j))
  \right],
\end{align*}
where $\relratio(\boldx)$ ($0\le\alpha\le1$) is the $\alpha$-relative importance weight defined by
\begin{align*}
  \relratio(\boldx):=
  \frac{p_{\mathrm{te}}(\boldx)}{(1-\alpha) p_{\mathrm{te}}(\boldx)+\alpha p_{\mathrm{tr}}(\boldx)}.
\end{align*}
Note that,
compared with the definition of
the $\alpha$-relative density-ratio \eqref{alpha-ratio},
$\alpha$ and $(1-\alpha)$ are swapped
in order to be consistent with exponential flattening.
Indeed, the relative importance weights play a similar role to
exponentially-flattened importance weights;
$\alpha=0$ corresponds to plain empirical-error minimization,
while $\alpha=1$ corresponds to importance-weighted empirical-error minimization;
$0<\alpha<1$ will give an intermediate estimator that balances the trade-off
between efficiency and consistency.
We note that the relative importance weights and exponentially flattened importance weights
agree only when $\alpha=\tau=0$ and $\alpha=\tau=1$;
for $0<\alpha=\tau<1$, they are generally different.

A possible advantage of the above relative importance weights is that
its estimation for $0<\alpha<1$ does not depend on that for $\alpha=1$,
unlike exponentially-flattened importance weights.
Since $\alpha$-relative importance weights for $0<\alpha<1$
can be reliably estimated by RuLSIF proposed in this paper,
the performance of covariate shift adaptation is expected to be improved.
Below, we experimentally investigate this effect.

\subsubsection{Artificial Datasets}
First, we illustrate how the proposed method behaves
in covariate shift adaptation using one-dimensional artificial datasets.

In this experiment, we employ the following kernel regression model:
\[
 f(x; \boldbeta) = \sum_{i = 1}^{n_{\mathrm{te}}}
 \beta_i \exp \left( -\frac{(x - x^{\mathrm{te}}_i )^2}{2\rho^2} \right),
\]
where $\boldbeta = (\beta_1, \ldots , \beta_{n_{\mathrm{te}}})^\top$ is the parameter to be learned
and $\rho$ is the Gaussian width.
The parameter $\boldbeta$ is learned by \emph{relative importance-weighted least-squares} (RIW-LS):
\begin{align*}
\widehat{\boldbeta}_{\mathrm{RIW-LS}} =
\argmin_{\boldbeta} \left[\frac{1}{n_{\mathrm tr}} \sum_{j = 1}^{n_{\mathrm tr}}
  \widehat{r}_\alpha(x_j^{\mathrm{tr}}) \left(f(x_j^{\mathrm{tr}}; \boldbeta) - y_j^{\mathrm{tr}}\right)^2 \right],
\end{align*}
or \emph{exponentially-flattened importance-weighted least-squares} (EIW-LS):
\begin{align*}
\widehat{\boldbeta}_{\mathrm{EIW-LS}}
 =
\argmin_{\boldbeta} \left[\frac{1}{n_{\mathrm tr}} \sum_{j = 1}^{n_{\mathrm tr}}
  \widehat{r}(x_j^{\mathrm{tr}})^\tau \left(f(x_j^{\mathrm{tr}}; \boldbeta) - y_j^{\mathrm{tr}}\right)^2 \right].
\end{align*}
The relative importance weight $\widehat{r}_\alpha(x_j^{\mathrm{tr}})$ is estimated
by RuLSIF, and the exponentially-flattened importance weight $\widehat{r}(x_j^{\mathrm{tr}})^\tau$
is estimated by uLSIF (i.e., RuLSIF with $\alpha=1$).
The Gaussian width $\rho$ is chosen by 5-fold \emph{importance-weighted cross-validation}
\citep{JMLR:Sugiyama+etal:2007}.

\begin{figure}[t]
  \centering
  \subfigure[Densities and ratios]{
    \includegraphics[width=.31\textwidth]{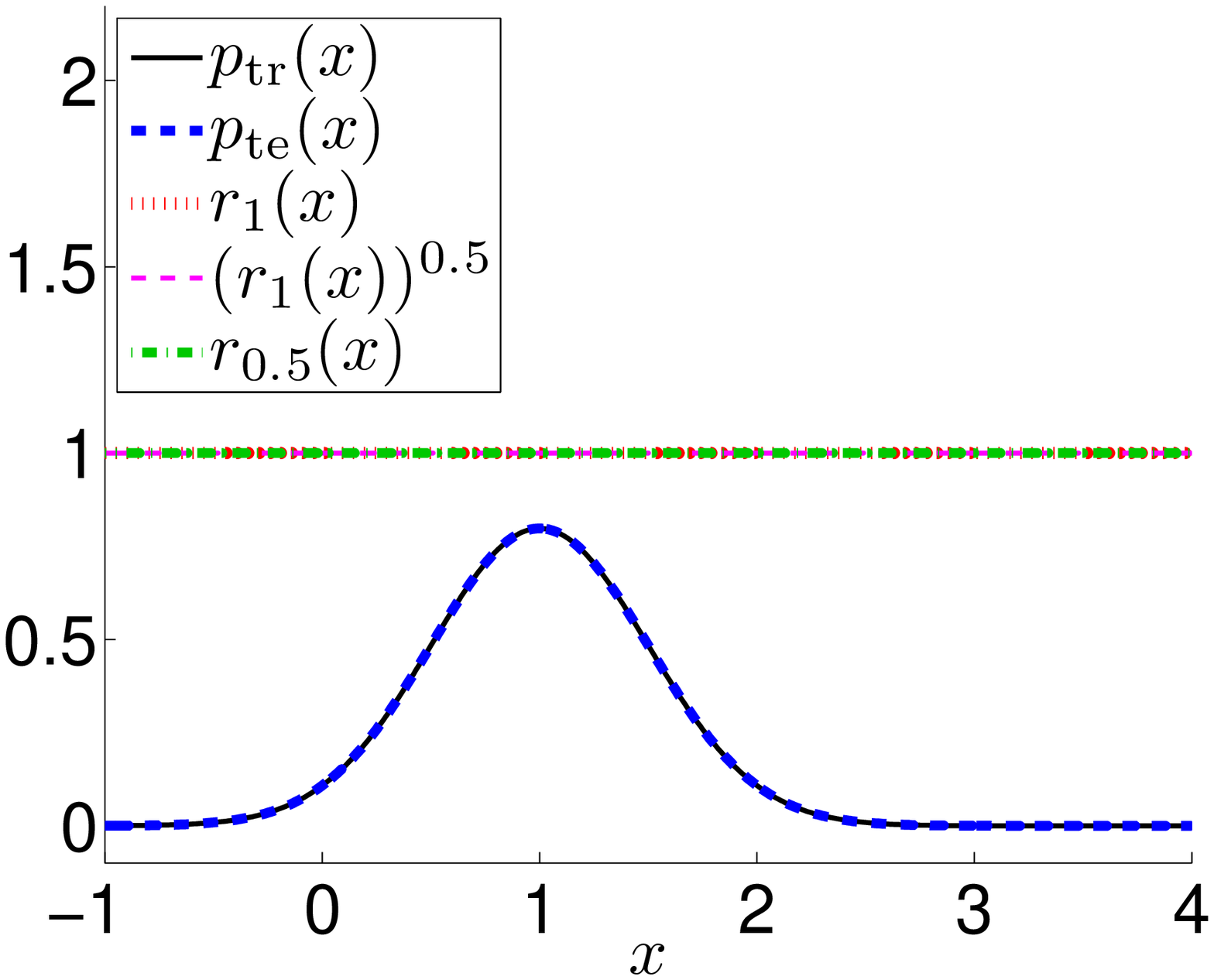}
    \label{fig:illust_transfer_regression_same-density}
  }
  \subfigure[Learned functions]{
    \includegraphics[width=.31\textwidth]{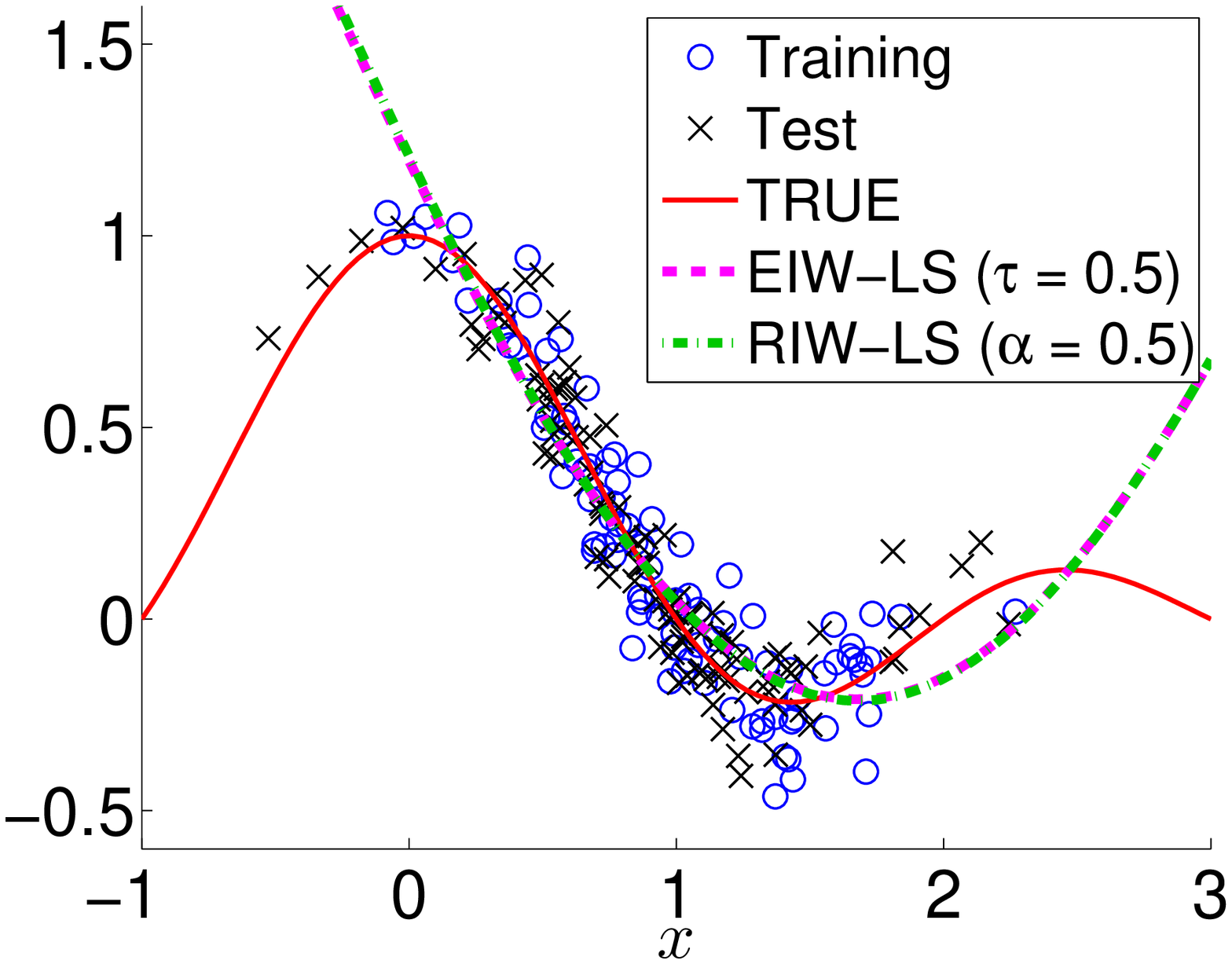}
    \label{fig:illust_transfer_regression_same-function}
  }
  \subfigure[Test error]{
    \includegraphics[width=.31\textwidth]{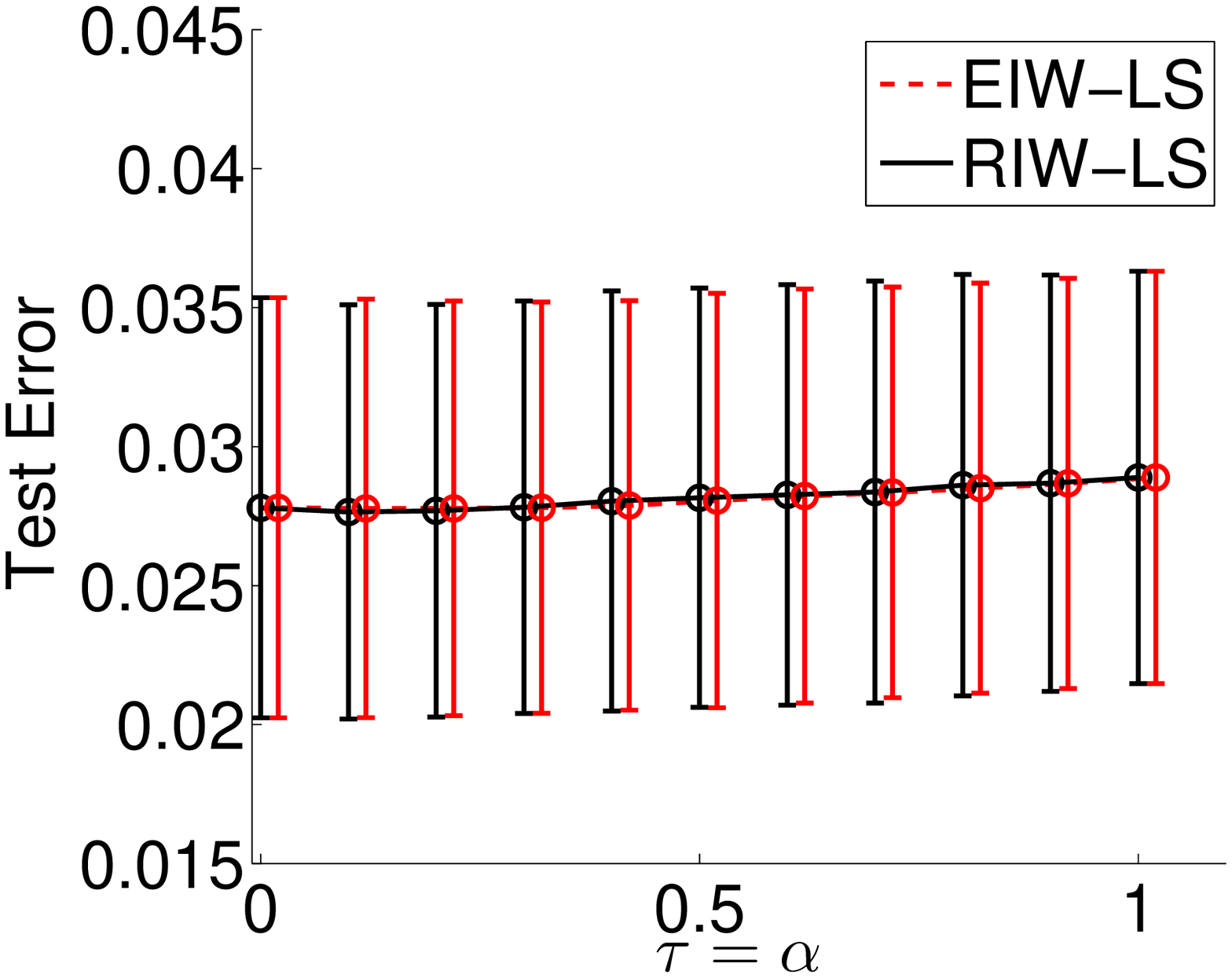}
    \label{fig:illust_transfer_regression_same-MSE}
  }
  \caption{Illustrative example of transfer learning under no distribution change.}
  \label{fig:illust_transfer_regression_same}
  \vspace*{5mm}
  \subfigure[Densities and ratios]{
    \includegraphics[width=.31\textwidth]{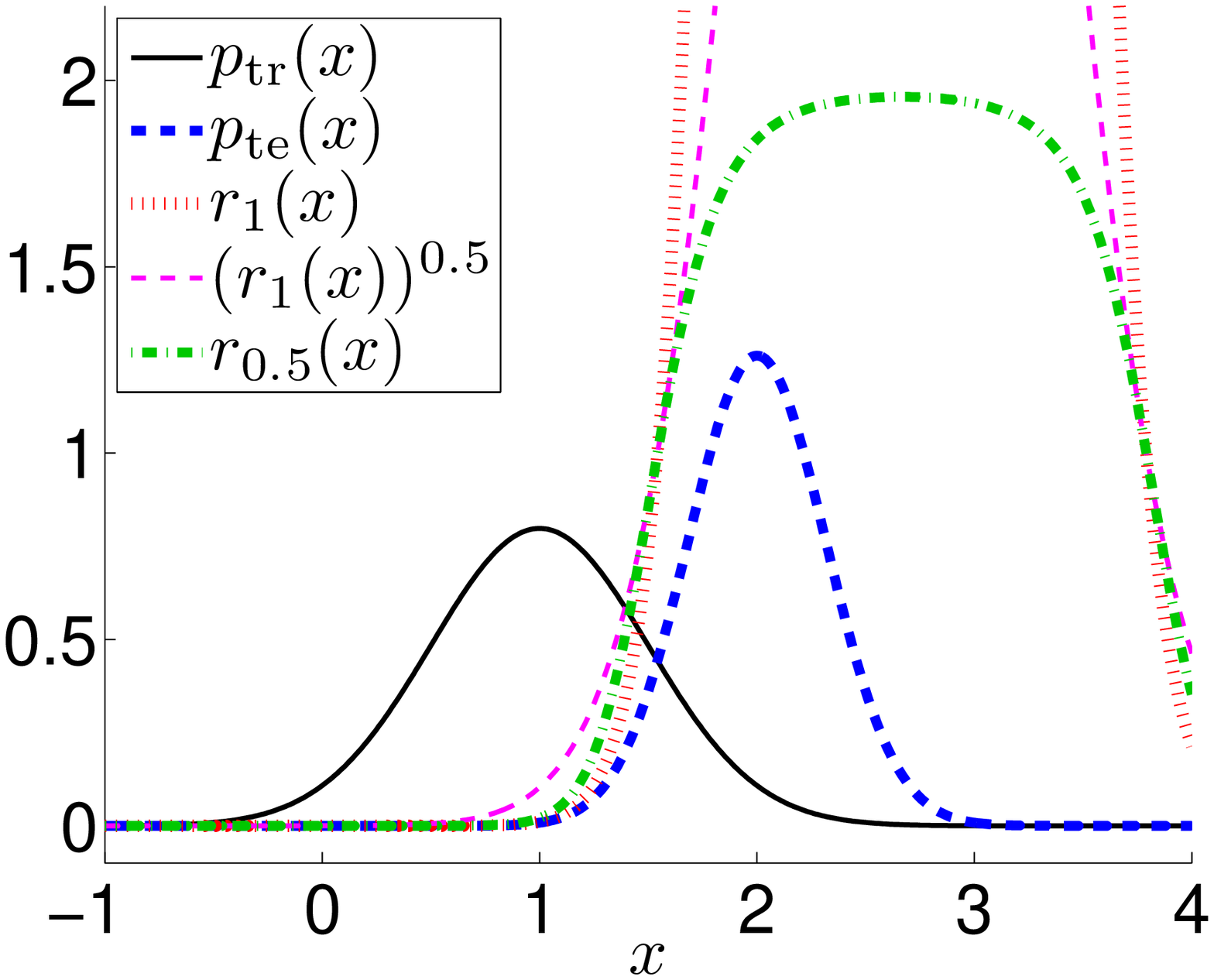}
    \label{fig:illust_transfer_regression_diff-density}
  }
  \subfigure[Learned functions]{
    \includegraphics[width=.31\textwidth]{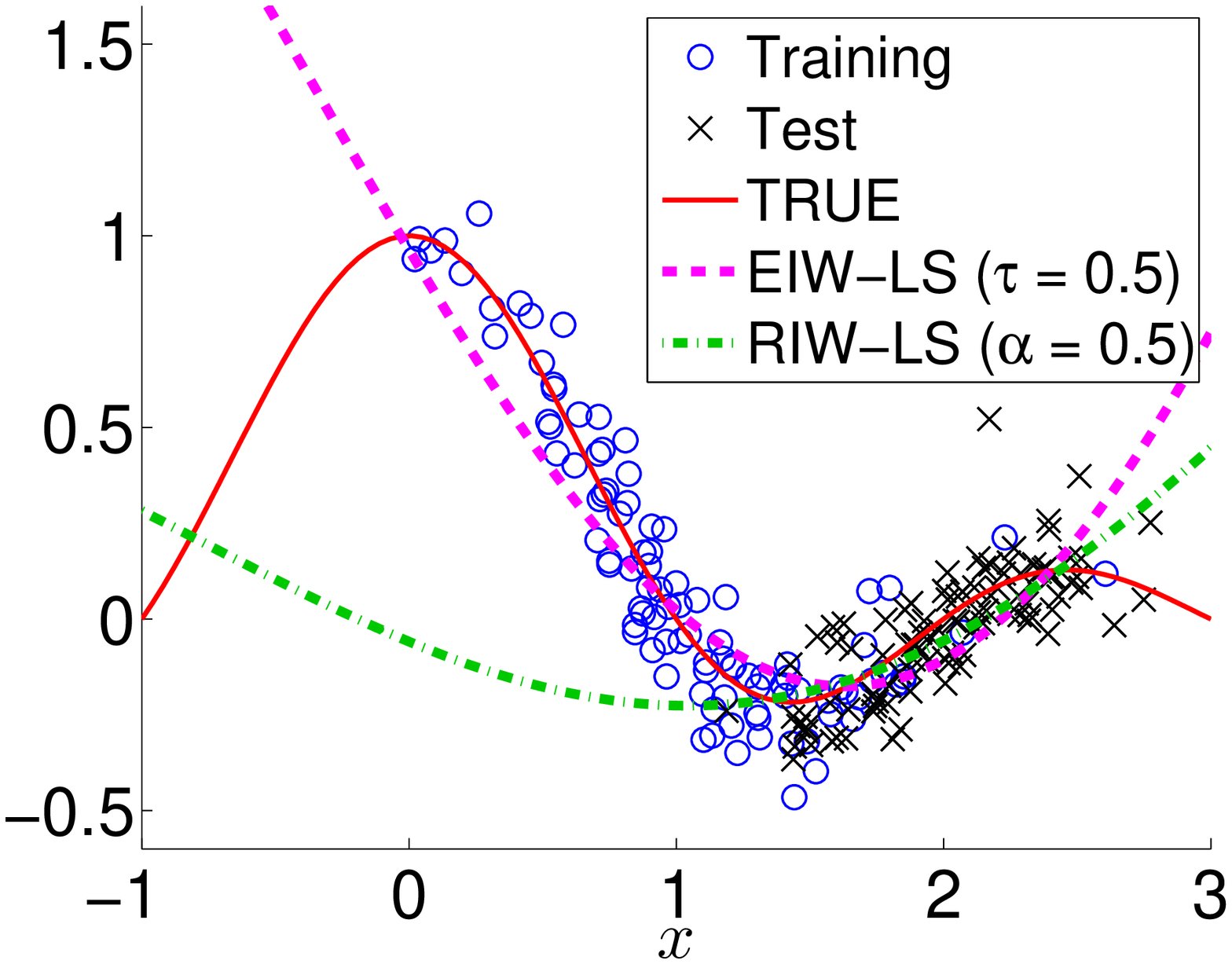}
    \label{fig:illust_transfer_regression_diff-function}
  }
  \subfigure[Test error]{
    \includegraphics[width=.31\textwidth]{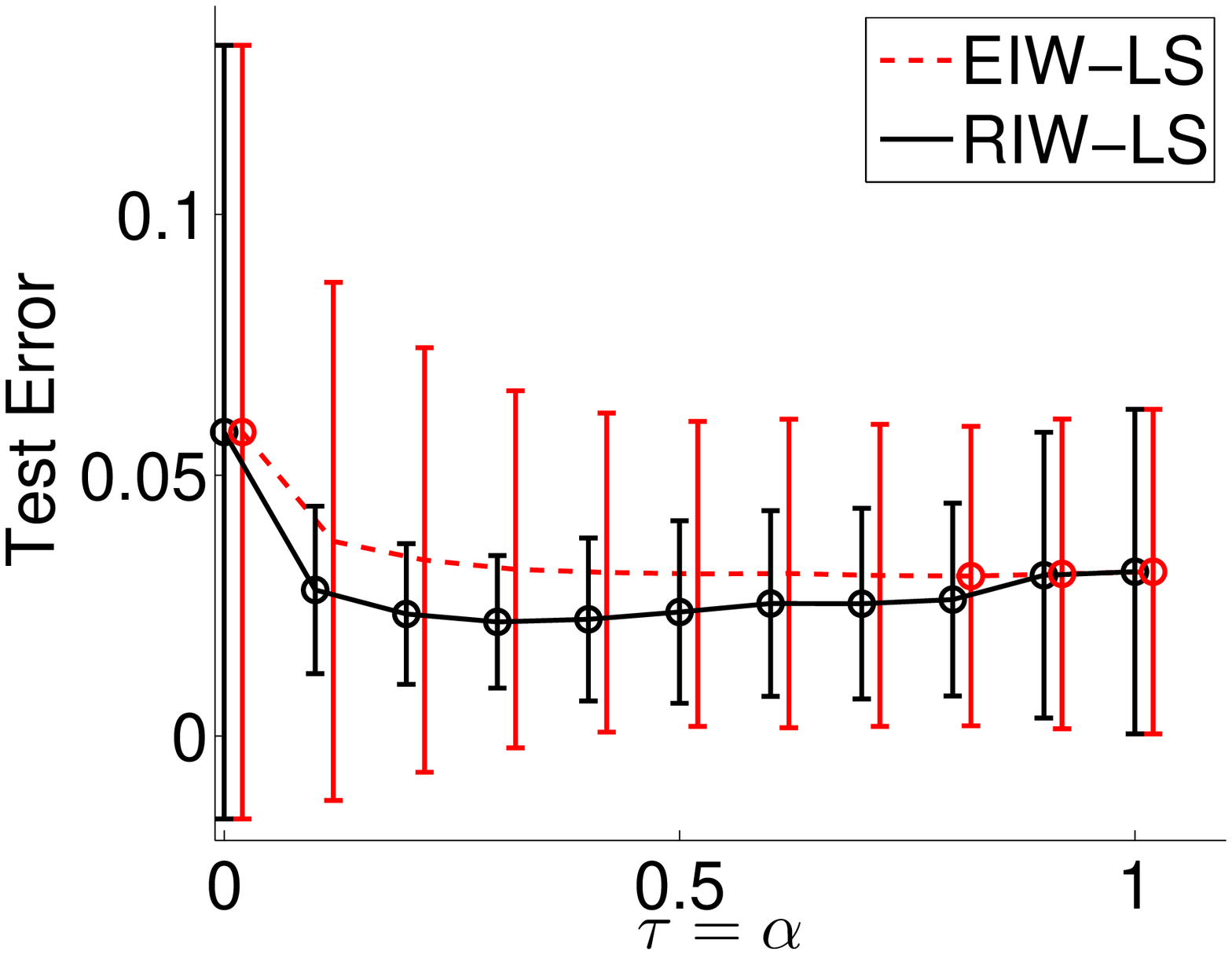}
    \label{fig:illust_transfer_regression_diff-MSE}
  }
  \caption{Illustrative example of transfer learning under covariate shift.}
  \label{fig:illust_transfer_regression_diff}
\end{figure}

First, we consider the case where input distributions do not change:
\begin{align*}
P_{\mathrm{tr}} =P_{\mathrm{te}}= N(1,0.25).
\end{align*}
The densities and their ratios are plotted
in Figure~\ref{fig:illust_transfer_regression_same-density}.
The training output samples $\{y_j^{\mathrm{tr}}\}_{j = 1}^{n_{\mathrm{tr}}}$ are generated as
\begin{align*}
y_j^{\mathrm{tr}} = \textnormal{sinc}(x_j^{\mathrm{tr}}) + \epsilon_j^{\mathrm{tr}},  
\end{align*}
where $\{\epsilon_j^{\mathrm{tr}}\}_{j = 1}^{n_{\mathrm{tr}}}$ is additive noise following $N(0,0.01)$.
We set $n_{\mathrm{tr}} = 100$ and $n_{\mathrm{te}} = 200$.
Figure~\ref{fig:illust_transfer_regression_same-function}
shows a realization of training and test samples 
as well as learned functions obtained by RIW-LS with $\alpha=0.5$ and EIW-LS with $\tau=0.5$.
This shows that RIW-LS with $\alpha=0.5$ and EIW-LS with $\tau=0.5$
give almost the same functions,
and both functions fit the true function well in the test region.
Figure~\ref{fig:illust_transfer_regression_same-MSE} shows the
mean and standard deviation of the test error under the squared loss over $200$ runs,
as functions of the relative flattening parameter $\alpha$ in RIW-LS
and the exponential flattening parameter $\tau$ in EIW-LS.
The method having a lower mean test error
and another method that is comparable
according to the \emph{t-test} at the significance level $5\%$ are
specified by `$\circ$'.  As can be observed, the proposed RIW-LS
compares favorably with EIW-LS.

Next, we consider the situation where input distribution changes
(Figure~\ref{fig:illust_transfer_regression_diff-density}):
\begin{align*}
P_{\mathrm{tr}} &= N(1,0.25),\\
P_{\mathrm{te}} &= N(2,0.1).
\end{align*}
The output values are created in the same way as the previous case.
Figure~\ref{fig:illust_transfer_regression_diff-function}
shows a realization of training and test samples 
as well as learned functions obtained by RIW-LS with $\alpha=0.5$ and EIW-LS with $\tau=0.5$.
This shows that RIW-LS with $\alpha=0.5$ fits the true function
slightly better than EIW-LS with $\tau=0.5$ in the test region.
Figure~\ref{fig:illust_transfer_regression_diff-MSE}
shows that the proposed RIW-LS tends to outperform EIW-LS,
and the standard deviation of the test error
for RIW-LS is much smaller than EIW-LS.
This is because EIW-LS with $0<\tau<1$ is based on
an importance estimate with $\tau=1$, which tends to have high fluctuation.
Overall, the stabilization effect of relative importance estimation
was shown to improve the test accuracy.

\subsubsection{Real-World Datasets}
Finally, we evaluate the proposed transfer learning method
on a real-world transfer learning task.

\begin{figure}[t]
 \centering
 \includegraphics[width=0.6\textwidth]{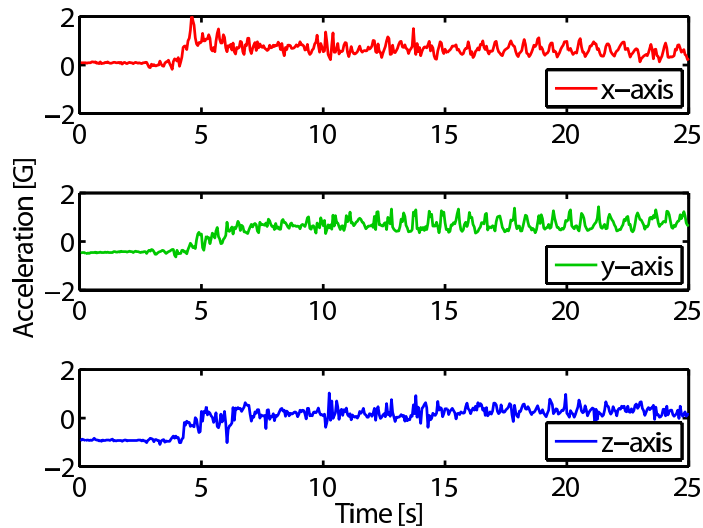}
 \caption{An example of three-axis accelerometer data for ``walking''
   collected by \emph{iPod touch}.}
 \label{fig:example-walk}
\end{figure}

We consider the problem of human activity recognition from accelerometer data
collected by \emph{iPod touch}\footnote{
\url{http://alkan.mns.kyutech.ac.jp/web/data.html}}.
In the data collection procedure,
subjects were asked to perform a specific action
such as walking, running, and bicycle riding.
The duration of each task was arbitrary and the sampling rate
was $20$Hz with small variations.
An example of three-axis accelerometer data for ``walking''
is plotted in Figure~\ref{fig:example-walk}.

To extract features from the accelerometer data,
each data stream was segmented in a sliding window manner
with window width $5$ seconds and sliding step $1$ second.
Depending on subjects, the position and orientation of \emph{iPod touch} was arbitrary---held
by hand or kept in a pocket or a bag.
For this reason, we decided to 
take the $\ell_2$-norm of the $3$-dimensional acceleration vector at each time step,
and computed the following $5$ 
orientation-invariant features from each window:
\emph{mean}, \emph{standard deviation},
\emph{fluctuation of amplitude}, \emph{average energy}, and \emph{frequency-domain entropy}
\citep{PerCom:Bao+Intille:2004,IFAWC:Bharatula+etal:2005}.

Let us consider a situation where a new user
wants to use the activity recognition system.
However, since the new user is not willing to label his/her accelerometer data
due to troublesomeness,
no labeled sample is available for the new user.
On the other hand, unlabeled samples for the new user
and labeled data obtained from existing users are available.
Let labeled training data $\{(\boldx^{\mathrm{tr}}_j,y^{\mathrm{tr}}_j)\}_{j=1}^{n_{\mathrm{tr}}}$ 
be the set of labeled accelerometer data for $20$ existing users.
Each user has at most $100$ labeled samples for each action.
Let unlabeled test data $\{\boldx^{\mathrm{te}}_i\}_{i=1}^{n_{\mathrm{te}}}$
be unlabeled accelerometer data obtained from the new user.

\begin{table}[t]
\centering
\caption{Experimental results of transfer learning in real-world human activity recognition.
  Mean classification accuracy (and the standard deviation in the bracket)
  over $100$ runs for activity recognition of a new user is reported.
  The method having the lowest mean classification accuracy and
  comparable methods according to the \emph{t-test} at the
  significance level $5\%$  are specified by bold face.}
\label{tb:ALKAN_result}
\begin{tabular}{|l||l@{}r|l@{}r|l@{}r|l@{}r|l@{}r|l@{}r|}
\hline
Task
& \multicolumn{2}{c|}{KLR} &  \multicolumn{2}{c|}{RIW-KLR}
& \multicolumn{2}{c|}{EIW-KLR} & \multicolumn{2}{c|}{IW-KLR}\\
& \multicolumn{2}{c|}{($\alpha=0$, $\tau=0$)}
& \multicolumn{2}{c|}{($\alpha = 0.5$)}
& \multicolumn{2}{c|}{($\tau = 0.5$)}
& \multicolumn{2}{c|}{($\alpha =1$, $\tau=1$)}\\
\hline \hline%
Walks vs.~run      & 0.803    &(0.082)   & {\bf 0.889}    & {\bf (0.035)}  & {\bf 0.882} &{\bf (0.039)} & {\bf 0.882} &{\bf (0.035)}\\ \hline
Walks vs.~bicycle  & 0.880    &(0.025)   & {\bf 0.892}    & {\bf (0.035)}  &      0.867  & (0.054)      &      0.854  & (0.070)\\ \hline
Walks vs.~train    & 0.985    &(0.017)   & {\bf 0.992}    & {\bf (0.008)}  &      0.989  & (0.011)      &      0.983  & (0.021) \\
\hline
\end{tabular}
\end{table}

We use \emph{kernel logistic regression} (KLR) for activity recognition.
We compare the following four methods:
\begin{itemize}
\item Plain KLR without importance weights (i.e., $\alpha=0$ or $\tau=0$).
\item KLR with relative importance weights for $\alpha=0.5$.
\item KLR with exponentially-flattened importance weights for $\tau=0.5$.
\item KLR with plain importance weights (i.e., $\alpha=1$ or $\tau=1$).
\end{itemize}
The experiments are repeated $100$ times with different sample choice
for $n_{\mathrm{tr}}=500$ and $n_{\mathrm{te}}=200$.
Table~\ref{tb:ALKAN_result} depicts the classification accuracy
for three binary-classification tasks:
\emph{walk vs.~run}, \emph{walk vs.~riding a bicycle}, 
and \emph{walk vs.~taking a train}.
The classification accuracy is evaluated
for $800$ samples from the new user that are not used 
for classifier training (i.e., the $800$ test samples are different
from $200$ unlabeled samples).
The table shows that
KLR with relative importance weights for $\alpha=0.5$
compares favorably with other methods
in terms of the classification accuracy.
KLR with plain importance weights
and
KLR with exponentially-flattened importance weights for $\tau=0.5$
are outperformed by
KLR without importance weights in the \emph{walk vs.~riding a bicycle} task
due to the instability of importance weight estimation
for $\alpha=1$ or $\tau=1$.

Overall, the proposed relative density-ratio estimation method
was shown to be useful also in transfer learning under covariate shift.

\section{Conclusion}
\label{sec:conclusion}

In this paper, we proposed to use a relative divergence for robust distribution comparison.
We gave a computationally efficient method
for estimating the relative Pearson divergence
based on direct relative density-ratio approximation.
We theoretically elucidated the convergence rate
of the proposed divergence estimator under non-parametric setup,
which showed that the proposed approach of estimating the relative Pearson divergence
is more preferable than the existing approach of estimating the plain Pearson divergence.
Furthermore, we proved that the asymptotic variance
of the proposed divergence estimator is independent of the model complexity
under a correctly-specified parametric setup. 
Thus, the proposed divergence estimator hardly overfits even with complex models.
Experimentally, we demonstrated the practical usefulness
of the proposed divergence estimator in two-sample homogeneity test,
inlier-based outlier detection, and transductive transfer learning
under covariate shift.

In addition to two-sample homogeneity test, outlier detection, and transfer learning,
density ratios were shown to be useful for tackling various machine learning problems,
including
multi-task learning \citep{ICML:Bickel+etal:2008,IPSJ:Simm+etal:2011},
independence test \citep{IEICE:Sugiyama+Suzuki:2011},
feature selection \citep{BMCBio:Suzuki+etal:2009},
causal inference \citep{AAAI:Yamada+Sugiyama:2010},
independent component analysis \citep{NC:Suzuki+Sugiyama:2011},
dimensionality reduction \citep{AISTATS:Suzuki+Sugiyama:2010},
unpaired data matching \citep{AISTATS:Yamada+Sugiyama:2011},
clustering \citep{JACIII:Kimura+Sugiyama:2011},
conditional density estimation \citep{IEICE:Sugiyama+etal:2010a},
and 
probabilistic classification \citep{IEICE:Sugiyama:2010}.
Thus, it would be promising to explore more applications
of the proposed relative density-ratio approximator
beyond two-sample homogeneity test, outlier detection, and transfer learning tasks.

\acks{%
MY was supported by the JST PRESTO program,
TS was partially supported by MEXT KAKENHI 22700289 and
Aihara Project, the FIRST program from JSPS, initiated by CSTP,
TK was partially supported by Grant-in-Aid for Young Scientists (20700251),
HH was supported by the FIRST program,
and
MS was partially supported by SCAT, AOARD, and the FIRST program.
}

\appendix
\section{Technical Details of Non-Parametric Convergence Analysis}
\label{appendix:nonparametric}
Here, we give the technical details of the non-parametric convergence analysis
described in Section~\ref{subsec:nonpara-analysis}.

\subsection{Results}
For notational simplicity, we define linear operators $\PP,\Pn,\QQ,\Qn$ as   
\begin{align*}
&\PP f := \EE_{p} f,~~~\Pn f := \frac{\sum_{j=1}^{\np} f(\boldxp_j)}{\np},\\  
&\QQ f := \EE_{q} f,~~~\Qn f := \frac{\sum_{i=1}^{\nq} f(\boldxq_i)}{\nq}.
\end{align*}
For $\alpha \in [0,1]$,
we define $S_{n,n'}$ and $S$ as 
$$
S_{n,n'} = \alpha \Pn + (1-\alpha) \Qn,~~~S = \alpha \PP + (1-\alpha) \QQ.
$$
We estimate the Pearson divergence between $p$ and $\alpha p +(1-\alpha)q$ through estimating the density ratio
$$
\tw := \frac{\pp}{\alpha \pp+(1-\alpha)\qq}.
$$ 
Let us consider the following density ratio estimator:
\begin{align*}
  \hw:= 
&
\mathop{\mathrm{argmin}}_{g\in \calG} 
\left[
\frac{1}{2} \left(\alpha \Pn + (1-\alpha)\Qn\right)g^2  
 - \Pn g
 +  \frac{\lambda_{\bar{n}}}{2} R(g)^2\right] \\
 =
 &
 \mathop{\mathrm{argmin}}_{g\in \calG} 
\left(
\frac{1}{2} S_{n,n'} g^2  
 - \Pn g
 +  \frac{\lambda_{\bar{n}}}{2} R(g)^2\right).
\end{align*}
where $\bar{n}=\min(\np,\nq)$ and $R(g)$ 
is a non-negative regularization functional such that
\begin{align}
\sup_{\boldx}[|g(\boldx)|] \leq R(g).
\label{eq:RegularizationBound}
\end{align}
A possible estimator of the Pearson (PE) divergence $\widehat{\mathrm{PE}}_\alpha$ is  
\begin{align*}
\widehat{\mathrm{PE}}_\alpha := 
\Pn \hw - \frac{1}{2} S_{n,n'} \hw^2 
-\frac{1}{2}.
\end{align*}
Another possibility is
\begin{align*}
\widetilde{\mathrm{PE}}_\alpha := 
\frac{1}{2} \Pn \hw -\frac{1}{2}.
\end{align*}

A useful example is
to use a \emph{reproducing kernel Hilbert space} \citep[RKHS;][]{AMS:Aronszajn:1950}
as $\calG$ and the RKHS norm as $R(g)$.
Suppose $\calG$ is an RKHS associated with bounded kernel $k(\cdot,\cdot)$:
\begin{align*}
\sup_{\boldx} [k(\boldx,\boldx)] \leq C.
\end{align*}
Let $\|\cdot\|_{\calG}$ denote the norm in the RKHS $\calG$.
Then $R(g)= \sqrt{C} \|g\|_{\calG}$ satisfies Eq.\eqref{eq:RegularizationBound}:
\begin{align*}
g(\boldx) &
=  \langle k(\boldx, \cdot), g(\cdot) \rangle  \leq 
 \sqrt{k(\boldx, \boldx )}  \|g\|_{\calG} \leq \sqrt{C} \|g\|_{\calG},
\end{align*}
where we used the reproducing property of the kernel
and Schwartz's inequality.
Note that the Gaussian kernel satisfies this with $C=1$.
It is known that the Gaussian kernel RKHS spans a dense subset in the set of continuous functions.
Another example of RKHSs is Sobolev space. 
The canonical norm for this space is the integral of the squared derivatives of functions.
Thus the regularization term $R(g) = \|g\|_{\calG}$ imposes the solution to be smooth.
The RKHS technique in Sobolev space has been
well exploited in the context of spline models \citep{Book:Wahba:1990}.
We intend that the regularization term $R(g)$ is a generalization of the RKHS norm. 
Roughly speaking, $R(g)$ is like a ``norm'' of the function space $\calG$.


We assume that the true density-ratio function $g^*(\boldx)$ 
is contained in the model $\calG$ and    
is bounded from above:
\[
g^*(\boldx) \leq M_0
\;\;\mbox{ for all }\;\;
\boldx\in \calDX.
\]
Let $\calG_M$ be a {\it ball} of $\calG$ with radius $M > 0$:
\begin{align}
&\calG_M:=\{g\in \calG \mid R(g)\leq M \}. \notag
\end{align}
To derive the convergence rate of our estimator, we utilize the {\it bracketing entropy} that is a complexity measure of a function class \citep[see p.~83 of][]{Book:VanDerVaart:WeakConvergence}.
\begin{definition}
Given two functions $l$ and $u$,
the bracket $[l,u]$ is the set of all functions $f$ with 
$l(\boldx) \leq f(\boldx) \leq u(\boldx)$
for all $\boldx$.
An $\epsilon$-bracket with respect to $L_2(\tilde{p})$ is a bracket $[l,u]$ with $\|l-u\|_{L_2(\tilde{p})} < \epsilon$. 
The bracketing entropy $\calH_{[]}(\calF,\epsilon,L_2(\tilde{p}))$ is the logarithm of the minimum number of $\epsilon$-brackets with respect to $L_2(\tilde{p})$ needed to cover a function set $\calF$.
\end{definition}
We assume that
there exists $\gamma$  $(0< \gamma < 2)$ such that, for all $M>0$,
\begin{align}
&\calH_{[]}(\calG_M,\epsilon,L_2(\pp))=O\left(\left(\frac{M}{\epsilon}\right)^{\gamma}\right),~~
\calH_{[]}(\calG_M,\epsilon,L_2(\qq))=O\left(\left(\frac{M}{\epsilon}\right)^{\gamma}\right)\label{eq:bracketGM}.
\end{align}
This quantity represents a complexity of function class $\calG$---the larger
$\gamma$ is, the more complex the function class $\calG$ is
because, for larger $\gamma$, more brackets are needed to cover the function class.
The Gaussian RKHS 
satisfies this condition 
for arbitrarily small $\gamma$ \citep{AS:STEINWART+SCOVEL:2007}. 
Note that when $R(g)$ is the RKHS norm, 
the condition \eqref{eq:bracketGM} holds for all $M>0$ if that holds for $M=1$.

Then we have the following theorem.

\begin{theorem}
\label{th:nonpara_convrate}
Let $\bar{n}=\min(\np,\nq)$, $M_0=\|\tw\|_{\infty}$, and $c = (1+\alpha) \sqrt{\PP(\tw - \PP\tw)^2} + (1-\alpha)\sqrt{\QQ(\tw - \QQ\tw)^2}$. Under the above setting, 
if $\lambda_{\bar{n}} \to 0$ and $\lambda_{\bar{n}}^{-1} = o(\bar{n}^{2/(2+\gamma)})$,
then we have
\begin{align*}
\widehat{\mathrm{PE}}_\alpha-\mathrm{PE}_\alpha
= \calO_p(\lambda_{\bar{n}}\max(1,R(\tw)^2) + \bar{n}^{-1/2}cM_0), 
\end{align*}
and 
\begin{align*}
\widetilde{\mathrm{PE}}_\alpha-\mathrm{PE}_\alpha
=&
\calO_p( 
\lambda_{\bar{n}}\max\{1,M_0^{\frac{1}{2}(1-\frac{\gamma}{2})},R(\tw)M_0^{\frac{1}{2}(1-\frac{\gamma}{2})} ,R(\tw)\}
+
\lambda_{\bar{n}}^{\frac{1}{2}}\max\{M_0^{\frac{1}{2}},M_0^{\frac{1}{2}} R(\tw)\}
),
\end{align*}
where $\calO_p$ denotes the asymptotic order in probability.
\end{theorem}


In the proof of Theorem~\ref{th:nonpara_convrate}, we use the following auxiliary lemma.

\begin{lemma}
\label{th:convergence_rate}
Under the setting of Theorem~\ref{th:nonpara_convrate}, 
if $\lambda_{\bar{n}} \to 0$ and $\lambda_{\bar{n}}^{-1} = o(\bar{n}^{2/(2+\gamma)})$, then we have
\begin{align*}
&\|\hw - g^* \|_{L_2(S)} = \calO_p(\lambda_{\bar{n}}^{1/2}\max\{1,R(\tw)\}),~~~R(\hw) = \calO_p(\max\{1,R(\tw)\}),
\end{align*}
where $\|\cdot \|_{L_2(S)}$ denotes the $L_2(\alpha p+(1-\alpha)q)$-norm.
\end{lemma}

\subsection{Proof of Lemma~\ref{th:convergence_rate}}
\label{sec:proofnonpara}

First, we prove Lemma~\ref{th:convergence_rate}.

From the definition, we obtain
\begin{align*}
&\frac{1}{2}S_{n,n'} \hw^2 - \Pn \hw + \lambda_{\bar{n}} R(\hw)^2 \leq 
\frac{1}{2}S_{n,n'} \tw^2 - \Pn \tw + \lambda_{\bar{n}} R(\tw)^2  \\
\Rightarrow
~~~&\frac{1}{2}S_{n,n'}(\hw - \tw)^2 - S_{n,n'}(\tw(\tw-\hw)) - \Pn (\hw -\tw) + \lambda_{\bar{n}} (R(\hw)^2 - R(\tw)^2) 
\leq 0.
\end{align*}
On the other hand, 
$
S(\tw(\tw-\hw)) = \PP(\tw-\hw)
$
indicates
\begin{align*}
&\frac{1}{2}(S-S_{n,n'})(\hw - \tw)^2 - (S-S_{n,n'})(\tw(\tw-\hw)) - (\PP-\Pn) (\hw -\tw) - \lambda_{\bar{n}} (R(\hw)^2 - R(\tw)^2) \notag \\
&\geq 
 \frac{1}{2} S(\hw - \tw)^2.
\end{align*}
Therefore, to bound $\|\hw - \tw \|_{L_2(S)}$, it suffices to bound the left-hand side of the above inequality.

Define $\calF_M$ and $\calF_M^2$ as 
$$
\calF_M := \{g-\tw \mid g \in \calG_M \}~~~\mbox{and}~~~\calF_M^2 := \{f^2 \mid f \in \calF_{M} \}.
$$
To bound $|(S-S_{n,n'})(\hw - \tw)^2|$, we need to bound the bracketing entropies of $\calF_M^2$.
We show that  
\begin{align*}
\calH_{[]}(\calF_M^2,\delta,L_2(p)) &= O\left(\left(\frac{(M+M_0)^2}{\delta}\right)^{\gamma}\right),\\
\calH_{[]}(\calF_M^2,\delta,L_2(q)) &= O\left(\left(\frac{(M+M_0)^2}{\delta}\right)^{\gamma}\right).
\end{align*}
This can be shown as follows.
Let $f_L$ and $f_U$ be a $\delta$-bracket for $\calG_M$ with respect to $L_2(p)$; $f_L(x) \leq f_U(x)$ and $\|f_L - f_U\|_{L_2(p)} \leq \delta$.
Without loss of generality, we can assume that $\|f_L\|_{L_{\infty}},\|f_U\|_{L_{\infty}}\leq M + M_0$ .
Then $f'_U$ and $f'_L$ defined as
\begin{align*}
&f'_U(x) := \max\{f_L^2(x),f_U^2(x)\}, \\ 
&f'_L(x) := \begin{cases}\min\{f_L^2(x),f_U^2(x)\} &(\mathrm{sign}(f_L(x))=\mathrm{sign}(f_U(x))), \\ 0&(\text{otherwise})  \end{cases},
\end{align*}
are also a bracket such that $f'_L \leq g^2 \leq f'_U$ for all $g\in \calG_M$ s.t. $f_L \leq g \leq f_U$ and $\|f'_L - f'_U\|_{L_2(p)} \leq 2 \delta (M+M_0)$
because $\|f_L - f_U\|_{L_2(p)}\leq \delta$ and the following relation is met: 
\begin{align*}
(f'_L(x) - f'_U(x))^2 &\leq  \begin{cases}(f_L^2(x) -f_U^2(x))^2 &(\mathrm{sign}(f_L(x))=\mathrm{sign}(f_U(x))), \\ \max\{f_L^4(x),f_U^4(x)\}&(\text{otherwise})  \end{cases} \\
&\leq  \begin{cases}(f_L(x) -f_U(x))^2(f_L(x) + f_U(x))^2 &(\mathrm{sign}(f_L(x))=\mathrm{sign}(f_U(x))), \\ \max\{f_L^4(x),f_U^4(x)\}&(\text{otherwise})  \end{cases} \\
&\leq  \begin{cases}(f_L(x) -f_U(x))^2(f_L(x) + f_U(x))^2 &(\mathrm{sign}(f_L(x))=\mathrm{sign}(f_U(x))), \\ 
                    (f_L(x) - f_U(x))^2 (|f_L(x)| + |f_U(x)|)^2 &(\text{otherwise})  \end{cases} \\
                    &\leq  4 (f_L(x) -f_U(x))^2(M+M_0)^2.
\end{align*}
Therefore the condition for the bracketing entropies \eqref{eq:bracketGM} gives 
$\calH_{[]}(\calF_M^2,\delta,L_2(p)) = O\left(\left(\frac{(M+M_0)^2}{\delta}\right)^{\gamma}\right)$.
We can also show that $\calH_{[]}(\calF_M^2,\delta,L_2(q)) = O\left(\left(\frac{(M+M_0)^2}{\delta}\right)^{\gamma}\right)$ in the same fashion.

Let $f := \hw - \tw$. 
Then, as in Lemma 5.14 and Theorem 10.6 in \citet{Book:VanDeGeer:EmpiricalProcess}, 
we obtain
\begin{align}
&\!\!\!\!\!\!\!\!\!\!\!\!\!
|(S_{n,n'} - S)(f^2)| \leq \alpha |(\Pn - \PP)(f^2)| + (1-\alpha)|(\Qn - \QQ)(f^2)| \notag \\
=&
\alpha  \calO_p\left(\frac{1}{\sqrt{\bar{n}}} \|f^2\|_{L_2(\PP)}^{1-\frac{\gamma}{2}}(1+R(\hw)^2 + M_0^2)^{\frac{\gamma}{2}} \vee 
\bar{n}^{-\frac{2}{2+\gamma}}(1+R(\hw)^2 +M_0^2)\right) \notag \\
&+ (1-\alpha)  \calO_p\left(\frac{1}{\sqrt{\bar{n}}} \|f^2\|_{L_2(\QQ)}^{1-\frac{\gamma}{2}}(1+R(\hw)^2 + M_0^2)^{\frac{\gamma}{2}} \vee 
\bar{n}^{-\frac{2}{2+\gamma}}(1+R(\hw)^2 +M_0^2)\right) \notag \\
\leq
&
  \calO_p\left(\frac{1}{\sqrt{\bar{n}}} \|f^2\|_{L_2(S)}^{1-\frac{\gamma}{2}}(1+R(\hw)^2 + M_0^2)^{\frac{\gamma}{2}} \vee 
\bar{n}^{-\frac{2}{2+\gamma}}(1+R(\hw)^2 +M_0^2)\right),
\label{(Qn-Q)f^2}
\end{align}
where $a \vee b=\max(a, b)$ and we used
\begin{align*}
  \alpha \|f^2\|_{L_2(\PP)}^{1-\frac{\gamma}{2}} + (1-\alpha)\|f^2\|_{L_2(\QQ)}^{1-\frac{\gamma}{2}}
\leq \left( \int f^4 \dd (  \alpha  \PP + (1-\alpha) \QQ) \right)^{\frac{1}{2}(1-\frac{\gamma}{2})} = \|f^2\|_{L_2(S)}^{1-\frac{\gamma}{2}} 
\end{align*}
by Jensen's inequality for a concave function. 
Since 
\[
\|f^2\|_{L_2(S)} \leq  \|f\|_{L_2(S)} \sqrt{2(1+R(\hw)^2 + M_0^2)},
\]
the right-hand side of Eq.\eqref{(Qn-Q)f^2} is further bounded by
\begin{align}
&\!\!\!\!\!\!\!\!\!\!\!\!\!
|(S_{n,n'} - S)(f^2)| \notag \\
= & \calO_p\left(\frac{1}{\sqrt{\bar{n}}} \|f\|_{L_2(S)}^{1-\frac{\gamma}{2}}(1+ R(\hw)^2 + M_0^2)^{\frac{1}{2}+\frac{\gamma}{4}} \vee \bar{n}^{-\frac{2}{2+\gamma}}(1+R(\hw)^2 + 
M_0^2)\right).
\label{eq:firstbound}
\end{align}
Similarly, we can show that 
\begin{align}
&\!\!\!\!\!\!\!\!\!\!\!\!\!
|(S_{n,n'} - S)(\tw(\tw-\hw))| \notag \\
= & \calO_p \left(\frac{1}{\sqrt{\bar{n}}} \|f\|_{L_2(S)}^{1-\frac{\gamma}{2}}(1+R(\hw)M_0 + M_0^2)^{\frac{\gamma}{2}} \vee \bar{n}^{-\frac{2}{2+\gamma}}(1+R(\hw)M_0 + M_0^2)\right),
\label{eq:secondbound}
\end{align}
and
\begin{align}
&|(\Pn - \PP)(\tw -\hw)| 
= \calO_p\left(\frac{1}{\sqrt{\bar{n}}} \|f\|_{L_2(\PP)}^{1-\frac{\gamma}{2}}(1+R(\hw) + M_0)^{\frac{\gamma}{2}} \vee \bar{n}^{-\frac{2}{2+\gamma}}(1+R(\hw) +M_0)\right) \notag \\
&\leq
\calO_p\left(\frac{1}{\sqrt{\bar{n}}} \|f\|_{L_2(S)}^{1-\frac{\gamma}{2}}(1+R(\hw) + M_0)^{\frac{\gamma}{2}}M_0^{\frac{1}{2}(1-\frac{\gamma}{2})} \vee \bar{n}^{-\frac{2}{2+\gamma}}(1+R(\hw) +M_0)\right),
\label{eq:thirdbound}
\end{align}
where we used 
\[
\|f\|_{L_2(\PP)} = \sqrt{\int f^2 \dd \PP} 
= \sqrt{\int f^2 \tw \dd S} \leq M_0^{\frac{1}{2}} \sqrt{\int f^2 \dd S}
\]
in the last inequality.
Combining Eqs.\eqref{eq:firstbound}, \eqref{eq:secondbound}, and \eqref{eq:thirdbound}, 
we can bound the $L_2(S)$-norm of $f$ as
\begin{align}
&\frac{1}{2}\|f\|^2_{L_2(S)} + \lambda_{\bar{n}} R(\hw)^2  \nonumber\\
&\leq  
 \lambda_{\bar{n}} R(\tw)^2 + \calO_p\left(\frac{1}{\sqrt{\bar{n}}} \|f\|_{L_2(S)}^{1-\frac{\gamma}{2}}(1+R(\hw)^2 + M_0^2)^{\frac{1}{2}+\frac{\gamma}{4}} \vee \bar{n}^{-\frac{2}{2+\gamma}}(1+R(\hw)^2 +M_0^2)\right).
\label{eq:basicBoundNonpara}
\end{align}

The following is similar to the argument in Theorem 10.6 in \citet{Book:VanDeGeer:EmpiricalProcess},
but we give a simpler proof.

By Young's inequality, we have $a^{\frac{1}{2}-\frac{\gamma}{4}} b^{\frac{1}{2}+\frac{\gamma}{4}} \leq (\frac{1}{2}-\frac{\gamma}{4})a 
+ (\frac{1}{2}+\frac{\gamma}{4}) b \leq a + b$ for all $a,b>0$.
Applying this relation to Eq.\eqref{eq:basicBoundNonpara}, we obtain 
\begin{align*}
&\frac{1}{2}\|f\|^2_{L_2(S)} + \lambda_{\bar{n}} R(\hw)^2  \notag \\
&\leq  
 \lambda_{\bar{n}} R(\tw)^2 + \calO_p\left( \|f\|_{L_2(S)}^{2(\frac{1}{2}-\frac{\gamma}{4})}
\left\{\bar{n}^{-\frac{2}{2+\gamma}} (1+R(\hw)^2 + M_0^2)\right\}^{\frac{1}{2}+\frac{\gamma}{4}} \vee \bar{n}^{-\frac{2}{2+\gamma}}(1+R(\hw)^2 +M_0^2)\right) \\
&\mathop{\leq}^{\text{Young}}  
 \lambda_{\bar{n}} R(\tw)^2 + \frac{1}{4}\|f\|_{L_2(S)}^{2}+  \calO_p\left(\bar{n}^{-\frac{2}{2+\gamma}} (1+R(\hw)^2 + M_0^2) + \bar{n}^{-\frac{2}{2+\gamma}}(1+R(\hw)^2 +M_0^2)\right)\\
&= 
 \lambda_{\bar{n}} R(\tw)^2 + \frac{1}{4}\|f\|_{L_2(S)}^{2}+  \calO_p\left(\bar{n}^{-\frac{2}{2+\gamma}}(1+R(\hw)^2 +M_0^2)\right),
\end{align*}
which indicates
\begin{align*}
&\frac{1}{4}\|f\|^2_{L_2(S)} + \lambda_{\bar{n}} R(\hw)^2  
\leq 
 \lambda_{\bar{n}} R(\tw)^2 +  o_p\left(\lambda_{\bar{n}}(1+R(\hw)^2 +M_0^2)\right).
\end{align*}
Therefore, by moving $o_p(\lambda_{\bar{n}} R(\hw)^2)$ to the left hind side, we obtain 
\begin{align*}
\frac{1}{4}\|f\|^2_{L_2(S)} + \lambda_{\bar{n}}(1 - o_p(1)) R(\hw)^2  
&\leq 
\calO_p\left( \lambda_{\bar{n}} (1 + R(\tw)^2 + M_0^2)\right) \\
&\leq 
\calO_p\left( \lambda_{\bar{n}} (1 + R(\tw)^2)\right).
\end{align*}
This gives 
\begin{align*}
&\|f\|_{L_2(S)} = \calO_p(\lambda_{\bar{n}}^{\frac{1}{2}} \max\{1,R(\tw)\}), \\
&R(\hw) = \calO_p(\sqrt{1+R(\tw)^2}) = \calO_p(\max\{1,R(\tw)\}).
\end{align*}
Consequently, the proof of Lemma~\ref{th:convergence_rate} was completed.

\subsection{Proof of Theorem~\ref{th:nonpara_convrate}}
\label{sec:proofnonpara2}
Based on Lemma~\ref{th:convergence_rate},
we prove Theorem~\ref{th:nonpara_convrate}.

As in the proof of Lemma~\ref{th:convergence_rate}, let $f:= \hw - \tw$.
Since $(\alpha \PP+ (1-\alpha)\QQ)(f\tw) =S(f\tw) = Pf$, we have
\begin{align}
\widehat{\mathrm{PE}}_\alpha - \mathrm{PE}_\alpha
&=\frac{1}{2}S_{n,n'} \hw^2 - \Pn \hw - (\frac{1}{2}S\tw^2 - \PP\tw) \notag \\
&=\frac{1}{2}S_{n,n'} (f+\tw)^2 - \Pn(f+\tw) - \left(\frac{1}{2}S \tw^2 - \PP\tw\right) \notag \\
&=
\frac{1}{2}S f^2 + \frac{1}{2}(S_{n,n'} - S)f^2 + (S_{n,n'} - S)(\tw f) - (\Pn - \PP)f\notag\\
&\phantom{=}
 + \frac{1}{2}(S_{n,n'} - S)\tw^2  - (\Pn \tw - \PP\tw). 
\label{eq:empIsboundpreliminary}
\end{align}
Below, we show that each term of the right-hand side of the above equation is $\calO_p(\lambda_{\bar{n}})$.
By the central limit theorem, we have
\begin{align*}
&\frac{1}{2}(S_{n,n'} - S)\tw^2  - (\Pn \tw - \PP\tw)\\
&~~~ = \calO_p\left(\bar{n}^{-1/2}M_0\left((1+\alpha)\sqrt{\PP(\tw - \PP\tw)^2} + (1-\alpha)\sqrt{\QQ(\tw - \QQ\tw)^2}\right)\right).
\end{align*}
Since Lemma~\ref{th:convergence_rate} gives $\|f\|_2 = \calO_p(\lambda_{\bar{n}}^{\frac{1}{2}}\max(1,R(\tw)))$ and $R(\hw) = \calO_p(\max(1,R(\tw)))$,  
Eqs.\eqref{eq:firstbound}, \eqref{eq:secondbound}, and \eqref{eq:thirdbound} in the proof of Lemma~\ref{th:convergence_rate}
imply  
\begin{align}
|(S_{n,n'} - S)f^2| &= \calO_p \left(\frac{1}{\sqrt{\bar{n}}} \|f\|_{L_2(S)}^{1-\frac{\gamma}{2}}(1+ R(\tw))^{1+\frac{\gamma}{2}}\vee \bar{n}^{-\frac{2}{2+\gamma}} R(\tw)^2\right) \notag\\
&\leq
\calO_p(\lambda_{\bar{n}} \max(1,R(\tw)^2)), \notag\\
|(S_{n,n'} - S)(\tw f)| &= 
  \calO_p \left(\frac{1}{\sqrt{\bar{n}}} \|f\|_{L_2(S)}^{1-\frac{\gamma}{2}}(1+R(\hw)M_0 + M_0^2)^{\frac{\gamma}{2}} \vee \bar{n}^{-\frac{2}{2+\gamma}}(1+R(\hw)M_0 + M_0^2)\right) \notag \\
&\leq \calO_p(\lambda_{\bar{n}}\max(1, R(\tw)M_0^{\frac{\gamma}{2}}, M_0^{\gamma}R(\tw)^{1-\frac{\gamma}{2}},M_0R(\tw),M_0^2)) \notag \\
&\leq \calO_p(\lambda_{\bar{n}}\max(1, R(\tw)M_0^{\frac{\gamma}{2}}, M_0R(\tw))), \notag \\
&\leq
\calO_p(\lambda_{\bar{n}} \max(1,R(\tw)^2)), \notag\\
|(\Pn - \PP)f | 
&\leq
\calO_p\left(\frac{1}{\sqrt{\bar{n}}} \|f\|_{L_2(S)}^{1-\frac{\gamma}{2}}(1+R(\hw) + M_0)^{\frac{\gamma}{2}}M_0^{\frac{1}{2}(1-\frac{\gamma}{2})} \vee \bar{n}^{-\frac{2}{2+\gamma}}(1+R(\hw) +M_0)\right) \notag \\
&=  \calO_p(\lambda_{\bar{n}}\max(1,M_0^{\frac{1}{2}(1-\frac{\gamma}{2})},R(\tw)M_0^{\frac{1}{2}(1-\frac{\gamma}{2})} ,R(\tw))) \label{eq:PnPfBound} \\
&\leq
\calO_p(\lambda_{\bar{n}} \max(1,R(\tw)^2)), \notag 
\end{align}
where we used $\lambda_{\bar{n}}^{-1}=o(\bar{n}^{2/(2+\gamma)})$ and $M_0 \leq R(\tw)$.
Lemma~\ref{th:convergence_rate} also implies
$$
S f^2 = \|f\|_2^2 = \calO_p(\lambda_{\bar{n}} \max(1,R(\tw)^2)).
$$ 
Combining these inequalities with Eq.\eqref{eq:empIsboundpreliminary} implies 
\begin{align*}
\widehat{\mathrm{PE}}_\alpha - \mathrm{PE}_\alpha 
&= \calO_p(\lambda_{\bar{n}}\max(1,R(\tw)^2) + n^{-1/2}cM_0), 
\end{align*}
where we again used $M_0 \leq R(\tw)$.

On the other hand, we have 
\begin{align}
\widetilde{\mathrm{PE}}_\alpha - \mathrm{PE}_\alpha
&=\frac{1}{2}\Pn \hw -  \frac{1}{2}\PP \tw \notag \\
&=
\frac{1}{2}\left[(\Pn -\PP)(\hw - \tw) + \PP(\hw - \tw)  + (\Pn - \PP) \tw \right].
\label{eq:PE2boundBasic}
\end{align}
Eq.\eqref{eq:PnPfBound} gives
\begin{align*}
(\Pn -\PP)(\hw - \tw) 
&=  \calO_p(\lambda_{\bar{n}}\max(1,M_0^{\frac{1}{2}(1-\frac{\gamma}{2})},R(\tw)M_0^{\frac{1}{2}(1-\frac{\gamma}{2})} ,R(\tw))).
\end{align*}
We also have 
\begin{align*}
\PP(\hw - \tw) & \leq \|\hw - \tw \|_{L_2(\PP)} \leq \|\hw - \tw \|_{L_2(S)} M_0^{\frac{1}{2}} = \calO_p(\lambda_{\bar{n}}^{\frac{1}{2}}\max(M_0^{\frac{1}{2}},M_0^{\frac{1}{2}} R(\tw))), 
\end{align*}
and
$$
(\Pn - \PP) \tw = O_p(\bar{n}^{-\frac{1}{2}}\sqrt{\PP(\tw - \PP\tw)^2}) \leq O_p(\bar{n}^{-\frac{1}{2}}M_0) \leq \calO_p(\lambda_{\bar{n}}^{\frac{1}{2}}\max(M_0^{\frac{1}{2}},M_0^{\frac{1}{2}} R(\tw))),
$$
Therefore by substituting these bounds into the relation \eqref{eq:PE2boundBasic}, one observes that 
\begin{align}
&\widetilde{\mathrm{PE}}_\alpha - \mathrm{PE}_\alpha \notag \\
=&
\calO_p(
\lambda_{\bar{n}}^{\frac{1}{2}}\max(M_0^{\frac{1}{2}},M_0^{\frac{1}{2}} R(\tw))
+ 
\lambda_{\bar{n}}\max(1,M_0^{\frac{1}{2}(1-\frac{\gamma}{2})},R(\tw)M_0^{\frac{1}{2}(1-\frac{\gamma}{2})} ,R(\tw))).
\end{align}
This completes the proof.
\hfill\QED

\section{Technical Details of Parametric Variance Analysis}
\label{appendix:parametric}
Here, we give the technical details of the parametric variance analysis
described in Section~\ref{subsec:para-analysis}.

\subsection{Results}

For the estimation of the $\alpha$-relative density-ratio
\eqref{alpha-ratio}, the statistical model
\begin{align*}
 \relratioModel=\{g(\boldx;\boldtheta)~|~\boldtheta\in\Theta\subset\Rbb^b\}
\end{align*}
is used where $b$ is a finite number.
Let us consider the following estimator of $\alpha$-relative density-ratio, 
\begin{align*}
 \widehat{g}=\argmin_{g\in\relratioModel}\,\frac{1}{2}
 \bigg\{
 \frac{\alpha}{\mnu}\sum_{i=1}^{\mnu}(g(\boldxnu_i))^2
+\frac{1-\alpha}{\mde}\sum_{j=1}^{\mde}(g(\boldxde_j))^2
 \bigg\}
 -\frac{1}{\mnu}\sum_{i=1}^{\mnu}g(\boldxnu_i). 
\end{align*}
Suppose that the model is correctly specified, i.e.,
there exists $\boldtheta^*$ such that
\begin{align*}
  \ratiomodel(\boldx;\boldtheta^*)=\relratio(\boldx).
\end{align*}
Then, under a mild assumption \cite[see Theorem 5.23 of][]{Book:VanDerVaart:AsymptoticStat}, 
the estimator $\widehat{g}$ is consistent and the estimated
parameter $\widehat{\boldtheta}$ satisfies the asymptotic normality in the large sample
limit. 
Then, a possible estimator of the $\alpha$-relative Pearson divergence $\relPE$ is 
\begin{align*}
 \hatPEest=
 \frac{1}{\mnu}\sum_{i=1}^{\mnu}\widehat{g}(\boldxnu_i)
 -
 \frac{1}{2}\bigg\{
 \frac{\alpha}{\mnu}\sum_{i=1}^{\mnu}(\widehat{g}(\boldxnu_i))^2
 +\frac{1-\alpha}{\mde}\sum_{j=1}^{\mde}(\widehat{g}(\boldxde_j))^2
 \bigg\}
 -\frac{1}{2}. 
\end{align*}
Note that there are other possible estimators for $\relPE$ such as
\begin{align*}
 \tildePEest=\frac{1}{2\mnu} \sum_{i=1}^{\mnu}\widehat{g}(\boldxnu_i) -\frac{1}{2}. 
\end{align*}

We study the asymptotic properties of $\hatPEest$. 
The expectation under the probability $\pnu$ ($\pde$) 
is denoted as $\Enu[\cdot]$ ($\Ede[\cdot]$). 
Likewise, the variance is denoted as $\Vnu[\cdot]$ ($\Vde[\cdot]$). 
Then, we have the following theorem. 
\begin{theorem}
 \label{theorem:var_upperbound}
 Let $\|\ratio\|_\infty$ be the sup-norm of the standard density ratio $\ratio(\boldx)$, 
 and $\|\relratio\|_\infty$ be the 
 sup-norm of the $\alpha$-relative density ratio, i.e., 
 \begin{align*}
  \|\relratio\|_\infty=\frac{\|\ratio\|_\infty}{\alpha\|\ratio\|_\infty+1-\alpha}. 
 \end{align*}
 The variance of $\hatPEest$ is denoted as $\Vbb[\hatPEest]$. 
 Then, under the regularity condition for the asymptotic normality, 
 we have the following upper bound of $\Vbb[\hatPEest]$:
 \begin{align*}
 \Vbb[\hatPEest]
  &=
 \frac{1}{\mnu}\Vnu\bigg[\relratio-\frac{\alpha{}\relratio^2}{2}\bigg]
 +\frac{1}{\mde}\Vde\bigg[\frac{(1-\alpha)\relratio^2}{2}\bigg]
 +o\bigg(\frac{1}{\mnu},\,\frac{1}{\mde}\bigg) 
\\
 &\leq\frac{\|\relratio\|_\infty^2}{\mnu}
 +\frac{\alpha^2\|\relratio\|_\infty^4}{4\mnu}
 +\frac{(1-\alpha)^2\|\relratio\|_\infty^4}{4\mde}
 +o\!\left(\frac{1}{\mnu},\frac{1}{\mde}\right). 
\end{align*}
\end{theorem}

\begin{theorem}
 \label{theorem:PE2_var_upperbound}
 The variance of $\tildePEest$ is denoted as $\Vbb[\tildePEest]$. 
 Let $\nabla{g}$ be the gradient vector of $g$ with respect to $\param$ at
 $\param=\param^*$, i.e., 
 $(\nabla{g}(\x;\param^*))_j=\frac{\partial{g}(\x;\param^*)}{\partial\theta_j}$. 
The matrix $\boldU_\alpha$ is defined by 
 \begin{align*}
  \boldU_\alpha=\alpha\Enu[\nabla{g}\nabla{g}^\top]+(1-\alpha)\Ede[\nabla{g}\nabla{g}^\top]. 
 \end{align*}
 Then, under the regularity condition, the variance of $\tildePEest$ is asymptotically
 given as 
 \begin{align*}
 \Vbb[\tildePEest]
  &=
  \frac{1}{\mnu}\Vnu\bigg[\frac{\relratio+(1-\alpha\relratio)
  \Enu[\nabla{g}]^\top\boldU_\alpha^{-1}\nabla{g}}{2}\bigg]\nonumber\\
&\phantom{=}
 +\frac{1}{\mde}\Vde\bigg[\frac{(1-\alpha)\relratio\Enu[\nabla{g}]^\top\boldU_\alpha^{-1}\nabla{g}}{2}\bigg]
 +o\bigg(\frac{1}{\mnu},\frac{1}{\mde}\bigg).
\end{align*} 
\end{theorem}

\subsection{Proof of Theorem~\ref{theorem:var_upperbound}}
 \label{appendix:Proof_of_Theorem_var_upper_bound}

 Let $\widehat{\boldtheta}$ be the estimated parameter, i.e., 
 $\widehat{g}(\boldx)=g(\boldx;\widehat{\boldtheta})$. 
 Suppose that $\relratio(\boldx)=g(\boldx;\boldtheta^*)\in\relratioModel$ holds. 
 Let $\delta\boldtheta=\widehat{\boldtheta}-\boldtheta^*$, then the asymptotic expansion of
 $\hatPEest$ is given as 
\begin{align*}
 \hatPEest
 &=
 \frac{1}{\mnu}\sum_{i=1}^{\mnu}g(\boldxnu_i;\widehat{\boldtheta})
 -\frac{1}{2}
 \bigg\{
 \frac{\alpha}{\mnu}\sum_{i=1}^{\mnu}g(\boldxnu_i;\widehat{\boldtheta})^2
 +\frac{1-\alpha}{\mde}\sum_{j=1}^{\mde}g(\boldxde_j;\widehat{\boldtheta})^2
 \bigg\}-\frac{1}{2}\\
 &=
 \relPE 
 +
 \frac{1}{\mnu}\sum_{i=1}^{\mnu}(\relratio(\boldxnu_i)-\Enu[\relratio])
 +\frac{1}{\mnu}\sum_{i=1}^{\mnu}\nabla g(\boldxnu_i;\boldtheta^*)^\top\delta\boldtheta\\
 &\phantom{=}
 -\frac{1}{2}
 \bigg\{
 \frac{\alpha}{\mnu}\sum_{i=1}^{\mnu} (\relratio(\boldxnu_i)^2-\Enu[\relratio^2])
 + \frac{1-\alpha}{\mde}\sum_{j=1}^{\mde} (\relratio(\boldxde_j)^2-\Ede[\relratio^2]) 
 \bigg\}\\
 &\phantom{=}
 -\bigg\{
  \frac{\alpha}{\mnu}\sum_{i=1}^{\mnu}\relratio(\boldxnu_i) \nabla g(\boldxnu_i;\boldtheta^*)
 +\frac{1-\alpha}{\mde}\sum_{j=1}^{\mde} \relratio(\boldxde_j) \nabla g(\boldxde_j;\boldtheta^*)
 \bigg\}^\top
 \delta\boldtheta
 +o_p\bigg(\frac{1}{\sqrt{\mnu}},\,\frac{1}{\sqrt{\mde}}\bigg).
\end{align*}
 Let us define the linear operator $\Gnu$ as
 \[
 \Gnu{f}=\frac{1}{\sqrt{\mnu}}\sum_{i=1}^{\mnu}(f(\boldxnu_i)-\Enu[f]). 
 \]
Likewise, the operator $\Gde$ is defined for the samples from $\pde$. 
Then, we have 
\begin{align*}
 &\phantom{=}\hatPEest-\relPE\\ 
 &=
 \frac{1}{\sqrt{\mnu}}\Gnu\big(\relratio-\frac{\alpha}{2}\relratio^2\big)
 -\frac{1}{\sqrt{\mde}}\Gde\big(\frac{1-\alpha}{2}\relratio^2\big)\\
 &\phantom{=}
 +\left\{
 \Enu[\nabla{g}] -\alpha\Enu[\relratio\nabla{g}]-(1-\alpha)\Ede[\relratio\nabla{g}] \right\}^\top
 \delta\boldtheta
 +o_p\bigg(\frac{1}{\sqrt{\mnu}},\,\frac{1}{\sqrt{\mde}}\bigg)\\
 &=
 \frac{1}{\sqrt{\mnu}}\Gnu\big(\relratio-\frac{\alpha}{2}\relratio^2\big)
 -\frac{1}{\sqrt{\mde}}\Gde\big(\frac{1-\alpha}{2}\relratio^2\big)
 +o_p\bigg(\frac{1}{\sqrt{\mnu}},\,\frac{1}{\sqrt{\mde}}\bigg).
\end{align*}
The second equality follows from 
\[
\Enu[\nabla{g}] -\alpha\Enu[\relratio\nabla{g}]-(1-\alpha)\Ede[\relratio\nabla{g}]=0. 
\]
Then, the asymptotic variance is given as
\begin{align}
 \Vbb[\hatPEest]
 &=
  \frac{1}{\mnu}\Vnu\bigg[\relratio-\frac{\alpha}{2}\relratio^2\bigg]
 +\frac{1}{\mde}\Vde\bigg[\frac{1-\alpha}{2}\relratio^2\bigg]
 +o\bigg(\frac{1}{\mnu},\,\frac{1}{\mde}\bigg). 
 \label{eqn:proof-PE1val}
\end{align}
 We confirm that both $\relratio-\frac{\alpha}{2}\relratio^2$ and
 $\frac{1-\alpha}{2}\relratio^2$ are non-negative and 
 increasing functions with respect to $\ratio$ for any $\alpha\in[0,1]$. 
 Since the result is trivial for $\alpha=1$, we suppose $0\leq \alpha<1$. 
 The function $\relratio-\frac{\alpha}{2}\relratio^2$ is represented as 
 \begin{align*}
  \relratio-\frac{\alpha}{2}\relratio^2
  =
  \frac{\ratio(\alpha \ratio+2-2\alpha)}{2(\alpha \ratio+1-\alpha)^2}, 
 \end{align*}
 and thus, we have $\relratio-\frac{\alpha}{2}\relratio^2=0$ for $\ratio=0$. 
 In addition, the derivative is equal to 
 \begin{align*}
  \frac{\partial}{\partial{\ratio}}
  \frac{\ratio(\alpha \ratio+2-2\alpha)}{2(\alpha \ratio+1-\alpha)^2}
  =  
  \frac{(1-\alpha)^2}{(\alpha \ratio+1-\alpha)^3}, 
 \end{align*}
 which is positive for $\ratio\geq 0$ and $\alpha\in[0,1)$. 
 Hence, 
 the function $\relratio-\frac{\alpha}{2}\relratio^2$ is non-negative and 
 increasing with respect to $\ratio$. 
 Following the same line, we see that $\frac{1-\alpha}{2}\relratio^2$ is non-negative and
 increasing with respect to $\ratio$. Thus, we have the following inequalities, 
\begin{align*}
 0&\leq \relratio(\boldx)-\frac{\alpha}{2}\relratio(\boldx)^2 \leq 
 \|\relratio\|_\infty -\frac{\alpha}{2}\|\relratio\|_\infty^2,\\
 0&\leq\frac{1-\alpha}{2}\relratio(\boldx)^2\leq 
 \frac{1-\alpha}{2}\|\relratio\|_\infty^2. 
\end{align*}
As a result, upper bounds of the variances in Eq.\eqref{eqn:proof-PE1val} are given as 
\begin{align*}
\Vnu\bigg[\relratio-\frac{\alpha}{2}\relratio^2\bigg]
 &\leq 
 \bigg(\|\relratio\|_\infty -\frac{\alpha}{2}\|\relratio\|_\infty^2\bigg)^2, \\
\Vde\bigg[\frac{1-\alpha}{2}\relratio^2\bigg]
 &\leq 
 \frac{(1-\alpha)^2}{4}\|\relratio\|_\infty^4. 
\end{align*}
Therefore, the following inequality holds, 
\begin{align*}
 \Vbb[\hatPEest]&\leq 
 \frac{1}{\mnu}
 \bigg(\|\relratio\|_\infty-\frac{\alpha\|\relratio\|_\infty^2}{2}\bigg)^2
 +\frac{1}{\mde}\cdot\frac{(1-\alpha)^2\|\relratio\|_\infty^4}{4}
 +o\!\left(\frac{1}{\mnu},\,\frac{1}{\mde}\right)\\
 &\leq\frac{\|\relratio\|_\infty^2}{\mnu}
 +\frac{\alpha^2\|\relratio\|_\infty^4}{4\mnu}
 +\frac{(1-\alpha)^2\|\relratio\|_\infty^4}{4\mde}
 +o\!\left(\frac{1}{\mnu},\frac{1}{\mde}\right), 
\end{align*}
which completes the proof.

\subsection{Proof of Theorem~\ref{theorem:PE2_var_upperbound}}
The estimator $\widehat{\param}$ is the optimal solution of the following problem: 
\begin{align*}
 \min_{\mathbf{\theta}\in\Param}
 \left[
   \frac{1}{2\mnu}\sum_{i=1}^{\mnu}\alpha  g(\xnu_i;\param)^2
   +\frac{1}{2\mde}\sum_{j=1}^{\mde}(1-\alpha)g(\xde_j;\param)^2
   -\frac{1}{\mnu}\sum_{i=1}^{\mnu}g(\xnu_i;\param)
 \right].
\end{align*}
Then, the extremal condition yields the equation, 
\begin{align*}
 \frac{\alpha}{\mnu}\sum_{i=1}^{\mnu}
 {g}(\xnu_i;\widehat{\param})\nabla{g}(\xnu_i;\widehat{\param})
 +
  \frac{1-\alpha}{\mde}\sum_{j=1}^{\mde}
 g(\xde_j;\widehat{\param})
 \nabla{g}(\xde_j;\widehat{\param})
 -\frac{1}{\mnu}\sum_{i=1}^{\mnu}\nabla g(\xnu_i;\widehat{\param}) = 0. 
\end{align*}
Let $\delta\param=\widehat{\param}-\param^*$. 
The asymptotic expansion of the above equation around $\param=\param^*$ 
leads to 
\begin{align*}
& \frac{1}{\mnu}\sum_{i=1}^{\mnu}(\alpha\relratio(\xnu_i)-1)\nabla{g}(\xnu_i;\param^*)
 +
 \frac{1-\alpha}{\mde}\sum_{j=1}^{\mde}\relratio(\xde_j)\nabla{g}(\xde_j;\param^*)
+\boldU_\alpha\delta\param
+o_p\bigg(\frac{1}{\sqrt{\mnu}},\frac{1}{\sqrt{\mde}}\bigg)
 =\mathbf{0}. 
\end{align*}
Therefore, we obtain 
\begin{align*}
 &
 \delta\param 
 =
 \frac{1}{\sqrt{\mnu}}\Gnu((1-\alpha\relratio)\boldU_\alpha^{-1}\nabla{g})
 -\frac{1}{\sqrt{\mde}}\Gde((1-\alpha)\relratio{}\boldU_\alpha^{-1}\nabla{g})
 +o_p\bigg(\frac{1}{\sqrt{\mnu}},\frac{1}{\sqrt{\mde}}\bigg). 
\end{align*}
Next, we compute the asymptotic expansion of $\tildePEest$: 
\begin{align*}
\tildePEest
 &= 
 \frac{1}{2}\Enu[\relratio]+
 \frac{1}{2\mnu}\sum_{i=1}^{\mnu}(\relratio(\xnu_i)-\Enu[\relratio])\\
&\phantom{=}
 +\frac{1}{2\mnu}\sum_{i=1}^{\mnu}\nabla g(\xnu_i;\param^*)^\top\delta\param 
 -\frac{1}{2}
 +o_p\bigg(\frac{1}{\sqrt{\mnu}},\frac{1}{\sqrt{\mde}}\bigg)\\
 &= 
\relPE 
 +\frac{1}{2\sqrt{\mnu}}\Gnu(\relratio) +\frac{1}{2}\Enu[\nabla{g}]^\top\delta\param
 +o_p\bigg(\frac{1}{\sqrt{\mnu}},\frac{1}{\sqrt{\mde}}\bigg). 
\end{align*}
Substituting $\delta\param$ into the above expansion, we have 
\begin{align*}
 \tildePEest-\relPE
 &= 
   \frac{1}{2\sqrt{\mnu}}\Gnu(\relratio+(1-\alpha\relratio)\Enu[\nabla{g}]^\top\boldU_\alpha^{-1}\nabla{g})\\
&\phantom{=}
 - \frac{1}{2\sqrt{\mde}}\Gde((1-\alpha)\relratio\Enu[\nabla{g}]^\top\boldU_\alpha^{-1}\nabla{g})
 +o_p\bigg(\frac{1}{\sqrt{\mnu}},\frac{1}{\sqrt{\mde}}\bigg). 
\end{align*}
As a result, we have
\begin{align*}
 \Vbb[\tildePEest]
 &=
  \frac{1}{\mnu}\Vnu\bigg[\frac{\relratio
 +(1-\alpha\relratio)\Enu[\nabla{g}]^\top\boldU_\alpha^{-1}\nabla{g}}{2}\bigg]\\
&\phantom{=}
 +\frac{1}{\mde}\Vde
 \bigg[\frac{(1-\alpha)\relratio\Enu[\nabla{g}]^\top\boldU_\alpha^{-1}\nabla{g}}{2}\bigg]
 +o\bigg(\frac{1}{\mnu},\frac{1}{\mde}\bigg),
\end{align*}
which completes the proof.

\bibliography{bib_yamada,bib_suzuki,bib_sugiyama}

\begin{thebibliography}{49}
\bibitem[Pearson(1900)]{PhMag:Pearson:1900}
K.~Pearson.
\newblock On the criterion that a given system of deviations from the probable
  in the case of a correlated system of variables is such that it can be
  reasonably supposed to have arisen from random sampling.
\newblock \emph{Philosophical Magazine}, 50:\penalty0 157--175, 1900.
\end{thebibliography}

\end{document}